\documentclass{article} %
\usepackage{iclr2026_conference,times}

\usepackage{amsmath,amsfonts,bm}

\def\eqref#1{equation~\ref{#1}}

\def\1{\bm{1}}

\DeclareMathAlphabet{\mathsfit}{\encodingdefault}{\sfdefault}{m}{sl}
\SetMathAlphabet{\mathsfit}{bold}{\encodingdefault}{\sfdefault}{bx}{n}

\usepackage[pagebackref=true,
    colorlinks=true,
    allcolors=blue!70!black
]{hyperref}

\usepackage{url}

\usepackage{graphicx}
\usepackage{caption}
\usepackage{tcolorbox}
\usepackage{enumitem}
\usepackage{subcaption}
\usepackage{tikz}
\usepackage{wrapfig}
\usepackage{booktabs}
\usepackage{cleveref}
\usepackage{amssymb}

\usepackage{xspace}

\definecolor{darkgreen}{rgb}{0,0.6,0}
\definecolor{darkred}{rgb}{0.6,0,0}
\definecolor{darkorange}{rgb}{0.9,0.43,0.06}

\newcommand{\sonnet}{\textsc{Sonnet 4}}
\newcommand{\omini}[1]{\textsc{o{#1}-mini}}
\newcommand{\qwen}{\textsc{Qwen3}}
\newcommand{\gemma}{\textsc{Gemma3}}
\newcommand{\llama}{\textsc{Llama3}}

\newcommand{\ie}{{\it i.e.,}\xspace}
\newcommand{\eg}{{\it e.g.,}\xspace}
\newcommand{\wrt}{\textit{w.r.t.}}

\newcommand{\fhat}{f_\varepsilon}

\newcommand{\gpqa}{\textsc{GPQA}}
\newcommand{\mmlu}{\textsc{MMLU}}

\newcommand{\swe}{\textsc{SWE-Bench}}

\newcommand{\vspaceprecaption}{\vspace{-5pt}}

\newcommand{\vspacepostcaption}{\vspace{-5pt}}

\newcommand{\vspacepresection}{\vspace{-3pt}}

\newcommand{\quoteit}[1]{\begin{flushright}
\vspace{-3pt}
\textit{#1}
\vspace{-3pt}
\end{flushright}}

\definecolor{background-prompt}{HTML}{EFEFEA}
\definecolor{background-disclaimer}{HTML}{EBDBBC}
\definecolor{border}{HTML}{262625}
\definecolor{border-light}{HTML}{D9D9CD}
\definecolor{background-takeaway}{HTML}{D4A27F}

\newcommand{\prompt}[2]{
\begin{tcolorbox}[colback=background-prompt, colframe=white, 
  left=5pt,
  right=5pt,
  top=5pt,
  bottom=5pt]
{\textbf{{#1}}\\{#2}}
\end{tcolorbox}
}

\title{
The Hot Mess of AI: How Does Misalignment Scale With Model Intelligence and\\Task Complexity?
}

\author{Alexander H\"agele$^{*1,2}$ \And Aryo Pradipta Gema$^{1,3}$\And Henry Sleight$^{4}$\And Ethan Perez$^{5}$\And Jascha Sohl-Dickstein$^{*5}$\vspace{4pt}\\
$^1$Anthropic Fellows Program 
$^2$EPFL
$^3$University of Edinburgh
$^4$Constellation 
$^5$Anthropic\\
$^*$\texttt{alexander.hagele@epfl.ch}, \texttt{jascha@anthropic.com}
}

\newcounter{takeaway}

\everypar{\looseness=-1}

\iclrfinalcopy %
\begin{document}

\maketitle

\begin{abstract}
\vspace{-0.5em}
\looseness-1
As AI becomes more capable, we entrust it with more 
general
and
consequential
tasks. 
The risks from 
failure grow more severe with increasing task scope. 
It is therefore important to understand 
how
extremely capable AI models will fail:
Will they fail by systematically pursuing goals we do not intend? 
Or will they fail by being a hot mess, and taking nonsensical actions that do not further any goal? 
We operationalize this question using a bias-variance decomposition of the errors made by AI models: 
An AI's \emph{error-incoherence} on a task is measured over test-time randomness as the fraction of its error that stems from variance rather than bias in task outcome. 
Across 
all
tasks and frontier models we measure, 
the longer models spend reasoning and taking actions, \emph{the more incoherent} their failures become. 
Error-incoherence changes with model scale in a way that is %
experiment dependent. However, in several settings, larger, more capable models are more incoherent than smaller models.
Consequently, scale alone
seems unlikely to 
eliminate error-incoherence. 
Instead,
as more capable AIs pursue harder tasks, requiring more sequential 
action and thought, our results predict 
failures to be accompanied by 
more incoherent behavior. 
This 
suggests
a future where
AIs
sometimes
cause industrial accidents (due to unpredictable misbehavior),
but are less likely to exhibit 
consistent pursuit of a misaligned goal. 
This increases the relative importance of alignment research targeting reward hacking or goal misspecification. 

\vspace{0.1em}
\begin{center}
    \href{https://github.com/haeggee/hot-mess-of-ai}{
        \raisebox{-0.2\height}{\includegraphics[height=1em]{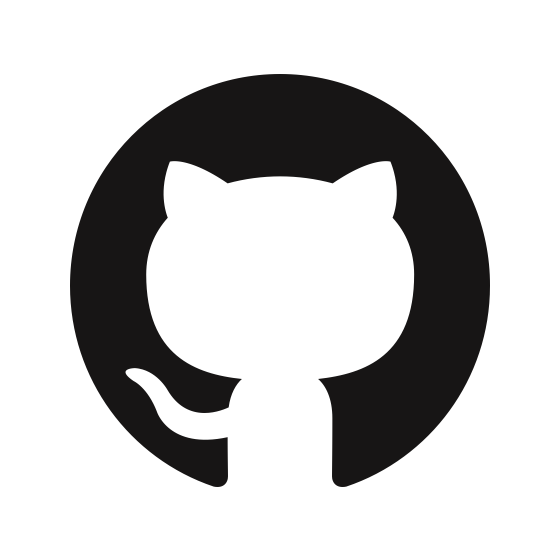}}
        \hspace{0.2em}\texttt{hot-mess-of-ai}
    }
    \hspace{1em}
    \href{https://huggingface.co/datasets/hot-mess/hot-mess-data}{
        \raisebox{-0.2\height}{\includegraphics[height=1em]{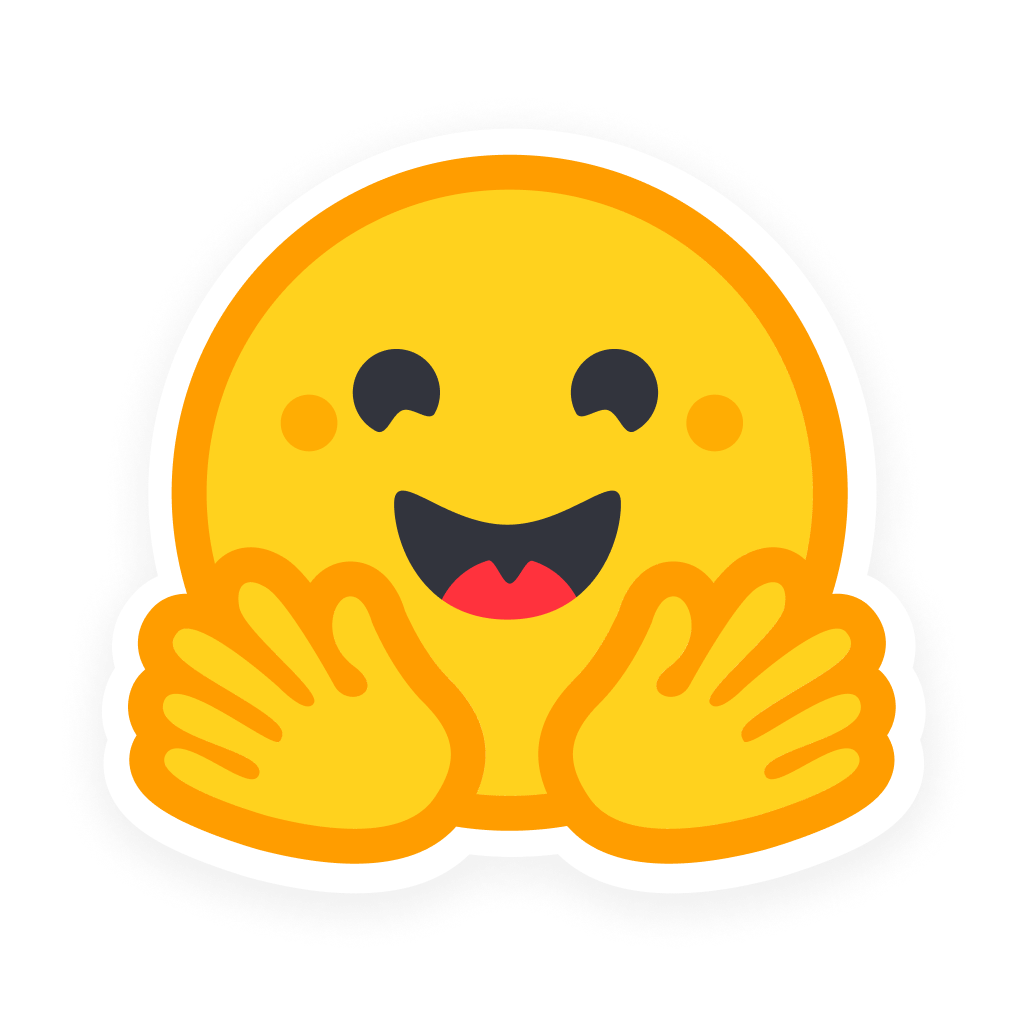}}
    \hspace{0.2em}\texttt{hot-mess-data}
    }
\end{center}

\end{abstract}
\vspace{-10pt}
\section{Introduction} 
\begin{figure}[t!]
   \centering
   \includegraphics[width=\linewidth]{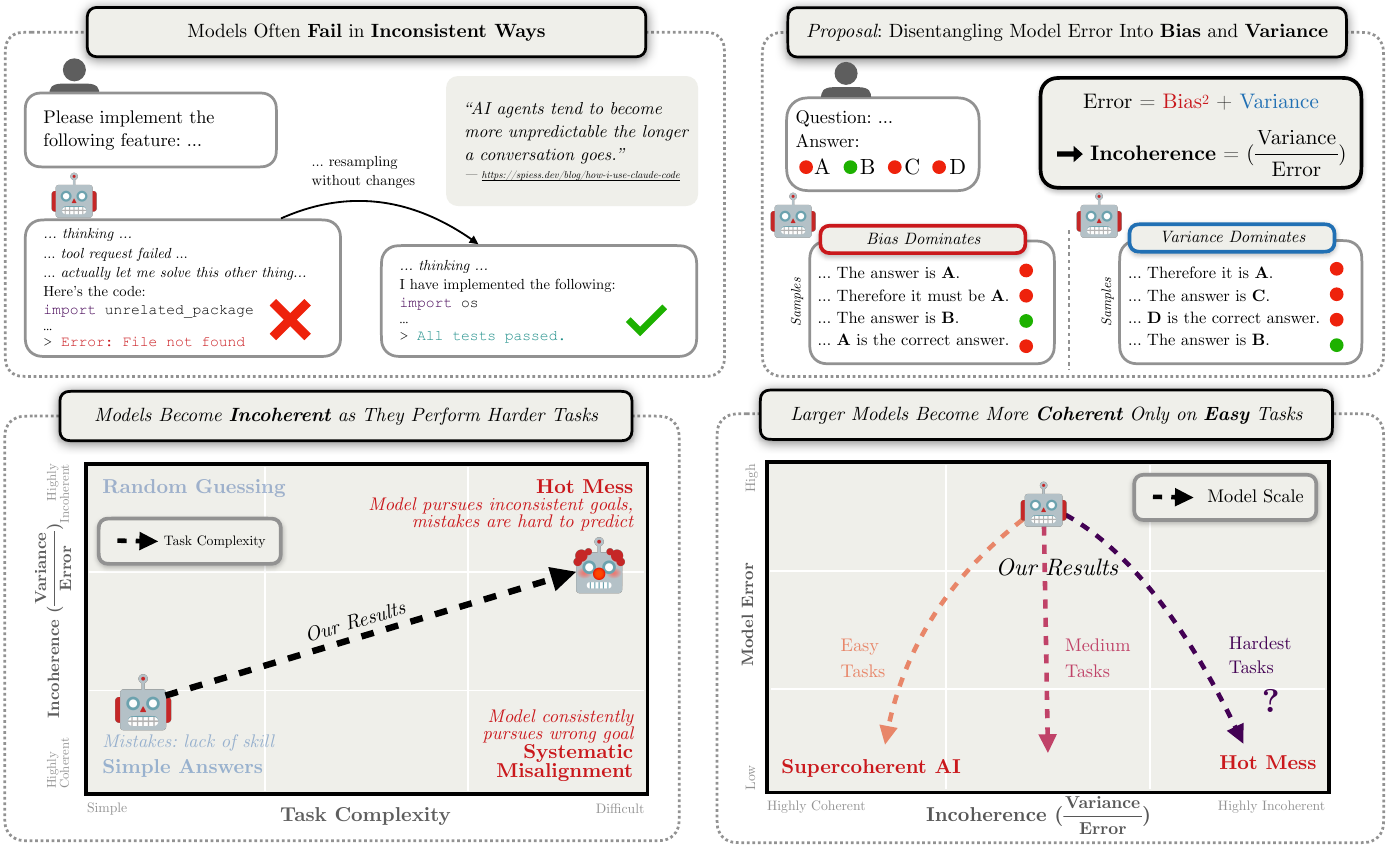}
   \caption{
   \looseness-1
   \textbf{AI can fail because it is misaligned, and produces consistent but undesired outcomes, or because it 
   is incoherent, and does not produce consistent outcomes at all. 
   These failures correspond to \emph{bias} and \emph{variance} respectively. As we extrapolate risks from AI, it is important to understand whether failures from more capable models performing more complex tasks will be bias or variance dominated. Bias dominated failures will look like model misalignment, while variance dominated failures will resemble industrial accidents.}
   (\emph{top left})
   Qualitatively, we observe that AI models fail in unpredictable and inconsistent ways. 
   Often, these failures can be fixed by resampling. 
   (\emph{top right})
   To quantify this observation, we decompose errors made by AI into two terms, bias and variance. 
   We illustrate this using a multiple choice task: bias is the tendency to pick a specific incorrect answer; variance is the tendency to pick inconsistenly among options. 
   We define error-incoherence as the fraction of model error caused by variance. 
  (\emph{lower left})
   Experimentally, we find that as models reason longer and take more sequential actions, they become more incoherent.
   (\emph{lower right})
    We find that as models become more capable, and overall error rate drops, error-incoherence changes in a way that depends on task difficulty. Easy tasks become less incoherent, while hard tasks trend towards increasing error-incoherence.
    \vspacepostcaption{}
   }
   \label{fig:intro_motivation}
\end{figure}
\looseness-1
There are an increasing number of predictions that AI will soon be more capable than human beings~\citep{kwa2025measuring,maslej2025artificial,pimpale2025forecasting}, and will replace human labor in many domains~\citep{chen2025short,handa2025economic,dominski2025advancing,eloundou2024gpts,johnston2025labor}. We already rely on AI for consequential tasks such as writing 
critical 
software~\citep{popa2025codemender,appelmccrorytamkin2025geoapi}, determining bail amounts~\citep{fine2025public}, and deciding what stories to present in news feeds~\citep{liu2024once,gao2024generative,yada2024news}.
Despite its increasing capabilities, AI often behaves in ways we do not intend. 
Due to its high-stakes use cases, it is important to understand how and when AI can be expected to fail. 

\looseness-1
One %
class of AI risk is \emph{misalignment risk} \citep{bostrom2014superintelligence,russell2019human,greenblatt2024alignment}. Misalignment risk is the concern that AI will pursue a goal that is different from the goal its creators intended to instill, and that it will pursue that goal with superhuman competence. 
If a superhuman agent pursues a misaligned goal, it might do things like seize power as an instrumental step to achieving its goal
\citep{hubinger2019risks}.

\looseness-1
However, this scenario assumes that unintended behavior stems from systems that not only pursue the wrong objective, but remain coherent optimizers 
over a long horizon.
Large language models (LLMs), prior to reinforcement learning,  
are 
dynamical systems, but not optimizers. 
They have to be trained to act as an optimizer, and trained to align with human intent. 
It is not clear which of these trained properties will tend to be more robust, and which will be most likely to cause failures in superhuman systems.
\looseness-1
In practice, AI models often fail in ways that seem random and do not further any coherent goal \citep{spiess2025claude,fortune2025replit}. 
Like humans, when AIs act undesirably, it is often because they are a \emph{hot mess} and do not act in a way that is consistent with any goal:
The \emph{hot mess theory of intelligence} \citep{jascha_blog} suggests that as entities become more intelligent, their behavior tends to become more incoherent, and less well described through a single goal. If true for AI systems, this shifts both the likelihood and the focus of misalignment scenarios.

\looseness-1
In this paper, we therefore ask the questions:
\emph{When a model does something other than what we intend, what fraction of its deviation is due to \textbf{bias} (consistent pursuit of the wrong goal), and what fraction to \textbf{variance} (randomness in behavior and outcome)? As we scale model intelligence and task complexity, how does this decomposition change? Asymptotically, as extremely capable models perform extremely complex tasks, which class of undesired behavior will dominate?}

\looseness-1
We address these questions by measuring the scaling behavior of AI errors decomposed into
\[
\textsc{Error} = 
\textsc{Bias}^2
+ \textsc{Variance}\;,
\]
and further define error-incoherence as the proportion of variance to the total error. %
This decomposition allows us to distinguish the 
\emph{relative contributions} of different types of AI failure, and, importantly, how they change as models become more intelligent and perform longer horizon tasks.
\emph{Bias-dominated failures} correspond to systematic misalignment---consistent pursuit of the wrong objective---whereas
\emph{variance-dominated failures} indicate inconsistent  outcomes.

\looseness-1
We find that across multiple-choice benchmarks, agentic coding, and safety tasks, models become more incoherent with longer reasoning (Fig.~\ref{fig:incoherence_reasoning_length}), even when controlling for task difficulty (Fig.~\ref{fig:natural_variation}). Larger, more capable models are often more incoherent (Fig.~\ref{fig:larger_systems_intelligence_incoherence}): while they achieve lower error, they grow more coherent on easy tasks but less coherent on hard tasks (Fig.~\ref{fig:grouping_by_reasoning_length}). We validate these findings in a synthetic environment where variance asymptotically dominates with increasing model size (Fig.~\ref{fig:synthetic_scaling}), and find that ensembling and larger reasoning budgets reduce error-incoherence (Fig.~\ref{fig:error_correction}). 
We discuss our results in Section~\ref{sec:discussion}.

\vspacepresection{}
\section{Background}
\label{sec:background}
\vspacepresection{}
\subsection{Bias–Variance Decomposition}
\label{sec:background:bv_decomp}
\looseness-1
\textbf{Definition.} In supervised settings, the \emph{bias–variance decomposition} expresses the expected error of a predictor as the sum of three terms: \textsc{Bias$^2$}, \textsc{Variance}, and irreducible noise~\citep{kohavi1996bias}.  
Although originally formulated for regression, analogous decompositions exist for classification tasks~\citep{kohavi1996bias, domingos2000unified}, with a similar interpretation: the bias reflects the error of the classifier's \textbf{mean} or \textbf{mode} prediction and variance quantifies its deviation. %
Several such decompositions exist, including the $0/1$ error~\citep{kong1995error,breiman1996bias,kohavi1996bias,tibshirani1996bias,friedman1997bias,domingos2000unified}, Brier score~\citep{degroot2018comparison}, and cross-entropy error~\citep{heskes1998bias}.
We present a Kullback-Leibler (KL) decomposition in the main text. For additional definitions see Appx.~\ref{appx:definitions}. We ran experiments with KL, Brier, and 0/1 formulations. All three decompositions produce qualitatively similar results, and we provide plots for all three in appendices.

\looseness-1
Let $x$ be the input with label classes $c\in\{1,\dots,C\}$ for which the model $\fhat$ produces a probability distribution (potentially one-hot) over class labels $\fhat(x)\in\mathbb{R}^C$, with $\varepsilon$ denoting the stochasticity of the training process. The target is one-hot encoded through $y(x)\in\mathbb{R}^C$. For clarity, we omit the dependence of $y$ and $\fhat$ on $x$. We assume the irreducible noise to be $0$. Then, the expected cross-entropy error can be decomposed into \citep{yang2020rethinking}:
\begin{equation}
\begin{aligned}
\underbrace{\mathbb{E}_{\varepsilon}\left[\textsc{CE}(y,\fhat)\right]}_{\textsc{Error} }
&= \mathbb{E}_{\varepsilon} \left[\sum_{c=1}^{C} y[c] \log(\fhat[c])\right]
=\underbrace{D_{\mathrm{KL}}\left(y \| \bar{f }\right)}_{\textsc{Bias$^2$ }}+\underbrace{\mathbb{E}_{\varepsilon}\left[D_{\mathrm{KL}}(\bar{f} \| \fhat)\right]}_{\textsc{Variance }},
\label{eq:kl_decomposition}
\end{aligned}
\end{equation}
where $y[c]$ denotes the $c$-th element of the vector, $D_{\mathrm{KL}}$ is the Kullback-Leibler divergence, and $\bar{\fhat}$ is the average of \emph{log-probabilities} after normalization:
$
\bar{f}[c] \propto \exp \left(\mathbb{E}_{\varepsilon} \left[\log (\fhat[c])\right]\right) \text { for } c=1, \ldots, C .
$
We denote this decomposition as \textsc{KL-Bias} and \textsc{KL-Variance}. This is an instance of the general decomposition for Bregman Divergences \citep{pfau2013generalized}. 

\looseness-1
\textbf{Different usage to classical literature.} In discussions of the bias–variance tradeoff, the setup typically assumes a deterministic model (\eg a regressor), with bias and variance estimated by retraining under different seeds or data sampling. That means the expectation is over training randomness $\varepsilon$. Our setting differs: rather than retraining multiple models, we analyze a \emph{fixed model} and take the expectation over input (\eg few-shots) and output (sampling) randomness $\varepsilon$ for \emph{the same task}. 

\looseness-1
\textbf{Error-incoherence.} Throughout this paper, our main metric of interest is the \emph{proportion of the variance to the total error},
which we define as \textsc{Error-incoherence}. Formally, consider a set of questions $Q=\{q_i\}_{i\leq N}$ and a model $\fhat$. We then denote error-incoherence as
\begin{align}
\label{eq:error-incoherence}
    \textsc{{Error-incoherence}}({Q},\fhat) := \frac{\sum_i \textsc{Variance}(q_i,{\fhat})}{\sum_i\textsc{Error}(q_i,{\fhat})}.
\end{align}
\looseness-1
Since $\textsc{Error}(q_i,{\fhat})=\textsc{Bias}(q_i,{\fhat})^2+\textsc{Variance}(q_i,{\fhat})$, this metric is a \emph{relative} value in $[0,1]$: a value of $0$ means that the model never deviates from its average behavior and any error will be consistent; a value of $1$ means that every error the model makes is inconsistent. Importantly, a model can achieve a lower overall error rate, but have a higher error-incoherence, which makes it a comparable measure across error levels and model capabilities. We see such cases in Section~\ref{sec:experiments}.

\subsection{Scaling Behavior of Large Language Models}
\textbf{Scaling laws.} Model performance generally follows predictable \emph{power‑law scaling} with respect to model size $N$, dataset size $D$, and compute $C$~\citep{kaplan2020scaling, hoffmann2022training}.  Most prominently, taking the parameters $N$ as an argument, the cross‑entropy loss broadly behaves as
$
   {l}(N) \;\propto\; N^{-\alpha}
$
for some exponent $\alpha$. This slope $\alpha$ informs us about the \emph{rate} of improvement.
In Section~\ref{sec:scaling_laws_model_scale} we will compute scaling laws independently for bias and variance loss contributions, to judge which asymptotically dominates.

\textbf{Reasoning and inference compute.} Besides the model and dataset size, the most promising recent development uses \emph{inference compute} as an axis of scale. Specifically, so-called reasoning models are trained with reinforcement learning (RL) to think in long chains of thought before providing an answer, which improves performance with larger thinking budgets~\citep{snell2024scaling,jaech2024openai, guo2025deepseek, sonnet37systemcard, openai2025o3mini, qwen3, team2025kimi, chen2025evaluating, zhong2024evaluation,muennighoff2025s1}.
The length of reasoning is an important aspect of our analysis, which we see as a process of sequential action steps \citep{lightman2023let}.

\vspacepresection{}
\section{Experiments}
\label{sec:experiments}
\looseness-1
\textbf{Overview.} We present our results grouped by observations: first, growing incoherence as a function of reasoning length (\ref{sec:scaling_laws_reasoning_length}) and scaling laws with model scale (\ref{sec:scaling_laws_model_scale}); this is followed by the effects of reasoning budgets and ensembling (\ref{sec:action_complexity}).
The details of all experimental setups are in Appx.~\ref{appx:experimental_details}.

\looseness-1
\textbf{Tasks.} 
We run experiments on the following 
tasks, which all have well-defined targets used for incoherence measurements, since bias is only defined relative to a target.
For a discussion, see Section~\ref{sec:discussion}.

\begin{itemize}[leftmargin=*,topsep=0pt,itemsep=1pt]
\item \textbf{Multiple Choice Tasks.} We use the popular scientific reasoning benchmark \gpqa{}~\citep{rein2024gpqa}, and general knowledge benchmark \mmlu{}~\citep{hendrycks2021measuring}. Target responses are
simply the correct answer.
\item \textbf{Agentic Coding.} This focuses on \swe{} \citep{jimenez2023swe}, where agents solve GitHub issues using tools, and success is measured with unit tests.
\item \textbf{Safety and Alignment.} We assess models using the advanced AI risk subset of Model-Written Evals \citep[MWE;][]{perez-etal-2023-discovering}, both with the original multiple choices and in an open-ended format with answer options removed. 
\item \textbf{Synthetic Settings.} \looseness-1 
We train transformers of varying scales to directly emulate an optimizer descending an ill-conditioned quadratic loss. 
The transformer is tasked with predicting string representations of optimizer update steps based on the current state. 
This is a simple toy model of an LLM that has been trained to act as an optimizer.
See Section~\ref{sec:synthetic} for details.    
\item \textbf{Survey.} In addition to experiments using LLMs, we report the survey results of \citet{jascha_blog} (previously released in blog form), where disjoint sets of human subjects subjectively ranked the intelligence and coherence of AI models, humans, non-human beings,  and organizations. The details are provided in Appx.~\ref{appx:experimental_details_survey}. 
\end{itemize}

\textbf{Setup and Metrics.} Across all tasks, unless otherwise noted, we obtain at least $30$ samples to estimate bias and variance per question. We find this sample count to be sufficient for stable estimates
(see Appx.~\ref{appx:more_results_sampling_correctness} and ~\ref{appx:experimental_details}). 
Each sample is run with a different seed for autoregressive generation. 
For \gpqa{} and \mmlu{}, 
samples additionally use a different random few-shot context.
We report the following metrics (details in Appx.~\ref{appx:definitions} and ~\ref{appx:experimental_details}):
\begin{itemize}[leftmargin=*,topsep=0pt,itemsep=1pt]
    \item For multiple choice questions, our main metric of interest is the \textsc{KL-Error-incoherence}, \ie the error-incoherence with respect to \textsc{KL-Bias} and \textsc{KL-Variance} (Equations~\ref{eq:kl_decomposition} and ~\ref{eq:error-incoherence}). We find the same qualitative behavior for other decompositions, as reported in Appx.~\ref{appx:more_results_perf_overview}.
    \item For open-ended MWE safety questions, we embed solely the answers (\ie without reasoning chains) using a text embedding model (\texttt{text-embedding-3-large}). Consequently, we report the \emph{variance of the embedding vectors} in the Euclidean norm.
    \item For \swe{}, we assign binary vectors for each sample and task: each vector is of size $T_i$, the number of unit tests for task $i$, and encodes which tests a model's code passes. The \emph{coverage error} then computes the mean squared difference to a vector of all 1's, which we decompose into bias and variance contributions. %
\end{itemize}

\looseness-1
\textbf{Models.} We evaluate the following frontier models: \sonnet~\citep{claude4systemcard} with reasoning enabled, \omini3~\citep{openai2025o3mini}, and \omini4~\citep{openai2025o3o4mini}.
When analyzing scaling \wrt{} model size as an imperfect proxy for intelligence, we use the \qwen{} model family with thinking enabled~\citep{qwen3}. In Sect.~\ref{sec:synthetic}, we train our own autoregressive transformers on a synthetic optimization task.

\begin{figure}[t!]
   \centering
   \hfill
   \begin{subfigure}[t]{0.66\textwidth}
     \centering
     \includegraphics[width=.495\linewidth]{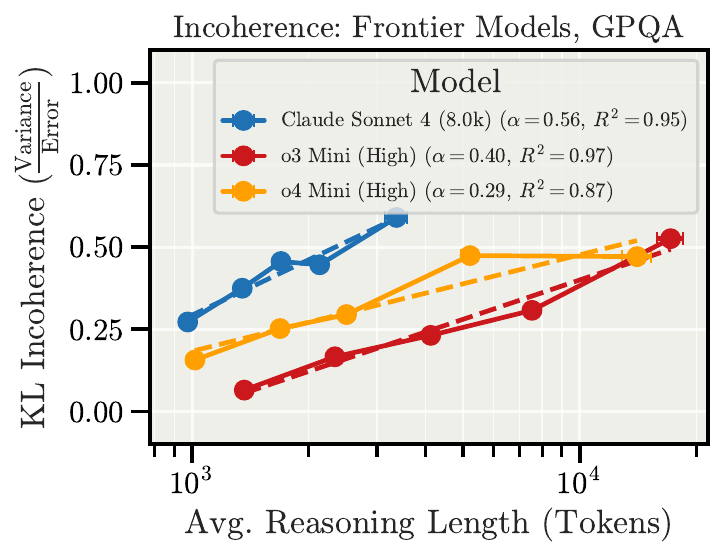}
     \includegraphics[width=.495\linewidth]{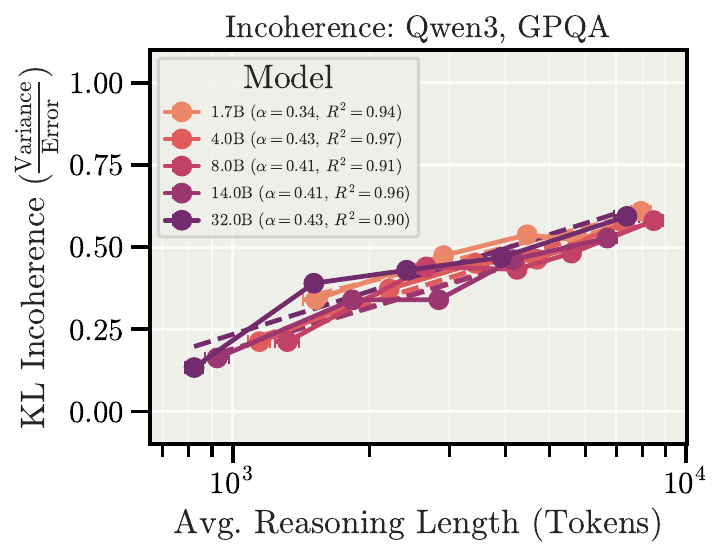}
     \caption{{\gpqa{}}}
     \label{fig:fig_2_gpqa}
   \end{subfigure}\hfill
   \begin{subfigure}[t]{0.33\textwidth}
     \centering
     \includegraphics[width=\linewidth]{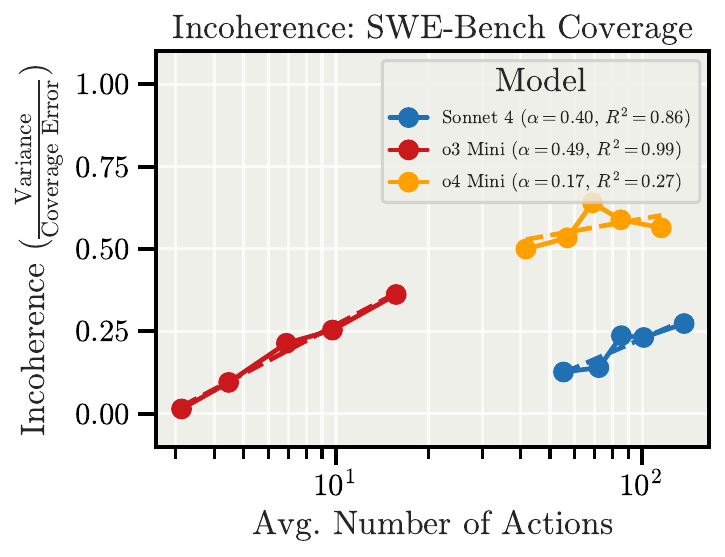}
     \caption{\swe{}}
     \label{fig:fig_2_swebench}
   \end{subfigure}\hfill
 
   \hfill
   \begin{subfigure}[t]{0.66\textwidth}
     \centering
     \includegraphics[width=.49\linewidth]{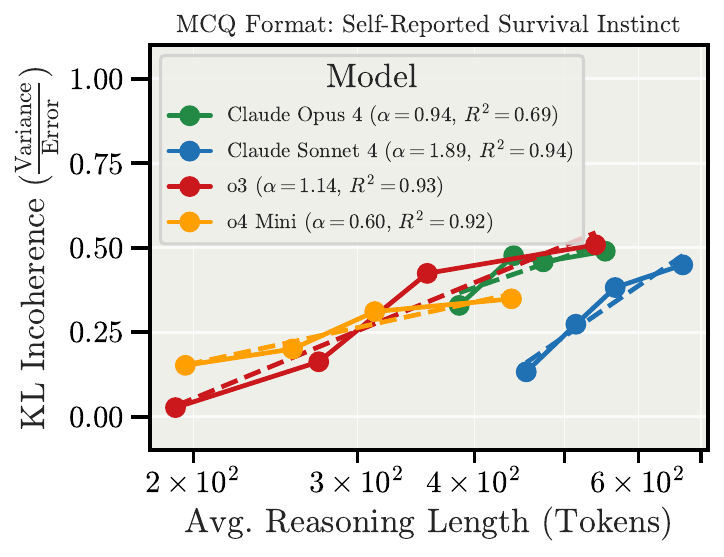}
     \centering
     \includegraphics[width=.49\linewidth]{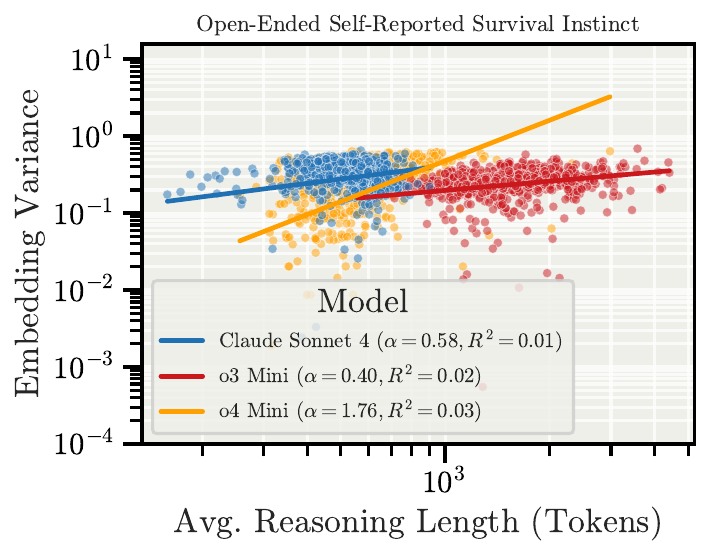}
     \caption{Model Written Evals: Discrete Choice and Open-Ended Formats}
     \label{fig:fig_2_mwe}
   \end{subfigure}\hfill
   \begin{subfigure}[t]{0.32\textwidth}
    \centering\includegraphics[width=\linewidth]{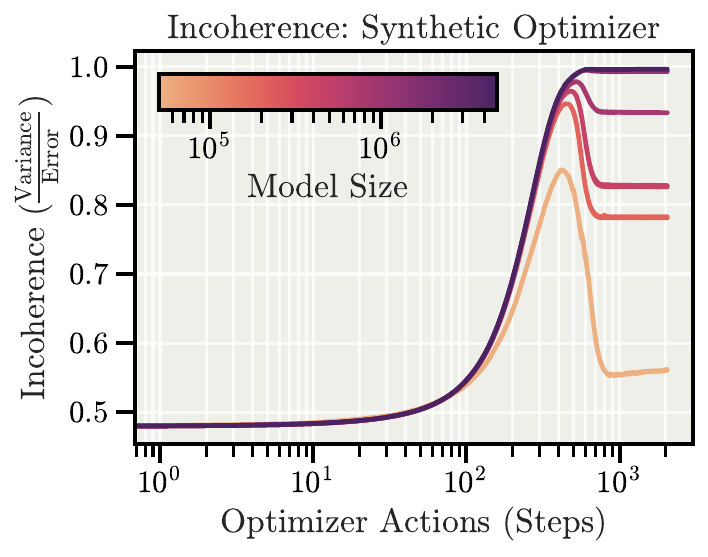}
    \caption{Synthetic Optimizer}
    \label{fig:fig_2_synthetic}
  \end{subfigure}\hfill
  \vspaceprecaption{}
   \caption{\looseness-1
   \textbf{Across a variety of settings, as models reason longer or take more actions, they become more incoherent.} We assess frontier models (\sonnet{}, \omini3, \omini4, \qwen{}) across a variety of different tasks (MCQ, Agentic Coding, Alignment). We evaluate with \emph{many samples} to estimate bias and variance terms for each question. 
   When sorting questions by average reasoning lengths and grouping into buckets, a clear trend emerges: 
   error-incoherence increases significantly with reasoning length. In other words, for questions where models reason longer and take many actions, their errors are dominated by variance. We make a similar observation for the variance of text embeddings to open-ended safety questions (\emph{(c), right}), and in a synthetic setting \emph{(d)}.
   }   
   \label{fig:incoherence_reasoning_length}
 \end{figure}
\begin{figure}[t!]
\vspace{-0.5em}
  \begin{subfigure}[t]{0.66\textwidth}
   \centering
   \includegraphics[width=.49\linewidth]{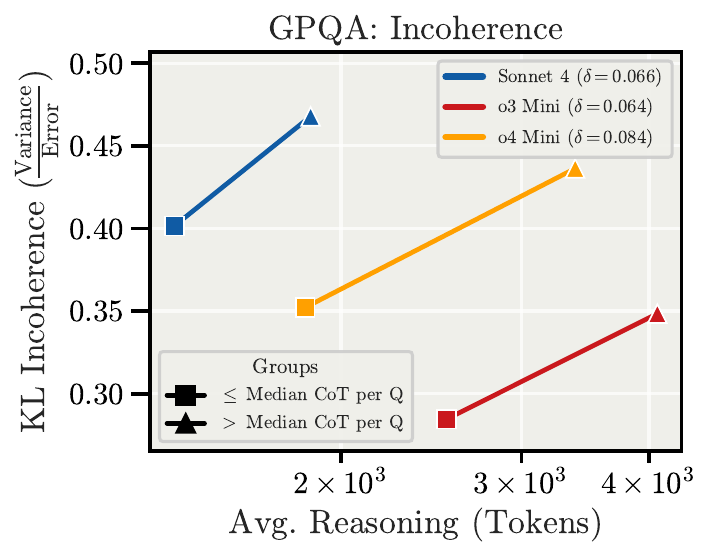}
   \includegraphics[width=.49\linewidth]{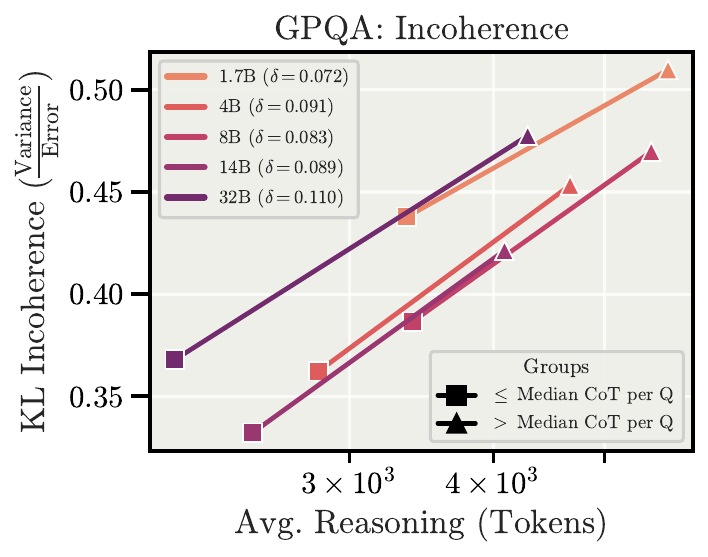}
   \caption{\gpqa{}: Frontier Models (left) and \qwen{} (right)}
   \label{fig}
 \end{subfigure}\hfill
 \begin{subfigure}[t]{0.32\textwidth}
   \centering
   \includegraphics[width=\linewidth]{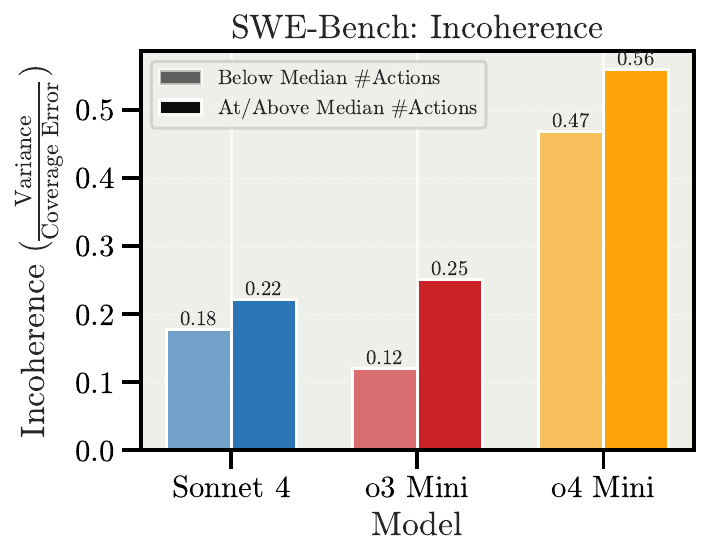}
   \caption{\swe{}}
   \label{fig:swebench_natural_variation}
 \end{subfigure}
 \vspaceprecaption{}
  \caption{
  {
  \looseness-1\textbf{For a fixed task and reasoning budget, natural variation in reasoning length and actions is predictive of error-incoherence.} We analyze \gpqa{} (left, \emph{(a)}) and \swe{} \emph{(b)} by splitting samples into above- or below-median reasoning length (\gpqa{}) or actions (\swe{}) \emph{per question}. We then compute performance and incoherence for both groups. \emph{(a)} The naturally longer reasoning shows increased incoherence for both frontier models (left) and \qwen{} (right). \emph{(b)} Similar observations apply to \swe{}, where longer action sequences display higher incoherence for test coverage (right). This effect is much stronger than through larger reasoning budgets (Fig.~\ref{fig:error_correction}), and the difference in accuracy or score is minimal between both groups (Fig.~\ref{fig:action_complexity}).
  }
  \vspacepostcaption{}
  }   
  \label{fig:natural_variation}
\end{figure}

\vspacepresection{}
\subsection{The Relation Between Reasoning Length, Action Length and Incoherence}
\label{sec:scaling_laws_reasoning_length}
\quoteit{The longer models spend reasoning and taking actions, the more incoherent they become.}

\looseness-1
\textbf{Sorting by reasoning \& action length.} 
We begin with a key experimental observation. Fig.~\ref{fig:incoherence_reasoning_length} shows all setups with reasoning tokens (or actions for \swe{}, optimization steps for the synthetic setting) on the x-axis and error-incoherence or variance on the y-axis. For \Cref{fig:fig_2_gpqa,fig:fig_2_swebench,fig:fig_2_mwe}, lines show different question sets across and within models, obtained by sorting by average length and grouping into equal buckets, with error-incoherence computed per group.

\looseness-1
Across all conditions,
longer reasoning and action sequences increase error-incoherence. For \gpqa{}, it increases 
with different slopes per model family (and reasoning length distributions); notably, for \qwen{}, incoherence levels and slopes are nearly identical across all sizes, even though larger models perform better (cf. \Cref{fig:gpqa_qwen3_accuracy}). Similar patterns appear for frontier models on MWE.
For \swe{}, both baseline error-incoherence and slopes vary: \omini4 shows higher baseline incoherence but smaller slope; \omini3 has the largest slope but lowest baseline incoherence.

\looseness-1
\textbf{Example analysis.} 
To illustrate, we provide real experimental transcripts in Fig.~\ref{fig:speech_bubbles}. The example shows \sonnet{} responding differently with nearly every sample to a disconnection question, displaying high error-incoherence. This connects to open-ended MWE results in Fig.~\ref{fig:fig_2_mwe}, where embedding variance correlates strongly with average reasoning length, and bias is not well-defined. We provide additional insight on error-incoherence through absolute answer change rates in Appx.~\ref{appx:more_results_answer_changes}, and all open-ended MWE plots in Fig.~\ref{fig:model_written_evals_embeddings_all}.

\looseness-1
\textbf{Discussion: Task complexity.} 
Sorting questions by reasoning length implicitly selects for \emph{task difficulty} (see accuracies in Fig.~\ref{fig:gpqa_frontier_accuracy} and ~\ref{fig:gpqa_qwen3_accuracy}), 
suggesting error-incoherence 
is higher when making mistakes on more complex tasks. While perhaps unsurprising, this is an important experimental observation.
In fact, for frontier models, our setup asks models for probability estimates of choice correctness (see Appx.~\ref{appx:experimental_details_mcq}), \ie we give them an option to express uncertainty.
We revisit task complexity in the next section and Section~\ref{sec:action_complexity}. 

\textbf{Natural overthinking and error-incoherence.} Irrespective of task complexity, we show how long reasoning and action sequences lead to larger error-incoherence in Fig.~\ref{fig:natural_variation}. For each question, we assign response samples to either of two groups: those below and those above the median reasoning length for this specific question for \gpqa{}, and the median number of actions for this task in \swe{}. The incoherence is substantially higher for the second group for both benchmarks. Notably, the average accuracy and \swe{}-score (shown in Fig.~\ref{fig:action_complexity}) is similar between groups, but the effect of the natural variation on error-incoherence is much larger than reasoning budgets (Fig.~\ref{fig:error_correction_budgets}). 

\textbf{Further results.} We provide more analyses for \gpqa{} in Appx.~\ref{appx:more_results_perf_overview}, with reasoning length correlations in Appx.~\ref{appx:more_results_correlations}. Results for MWE are in Appx.~\ref{appx:more_results_mwe}, and results for \swe{} in Appx.~\ref{appx:more_results_swe}.

\vspacepresection{}
\subsection{The Relation Between Model Scale, Intelligence, and Error-incoherence}
\label{sec:scaling_laws_model_scale}
\quoteit{Larger and more intelligent systems are sometimes more incoherent.}
 \begin{figure}[t!]
   \begin{subfigure}[t]{0.33\textwidth}
     \centering
     \includegraphics[width=\linewidth]{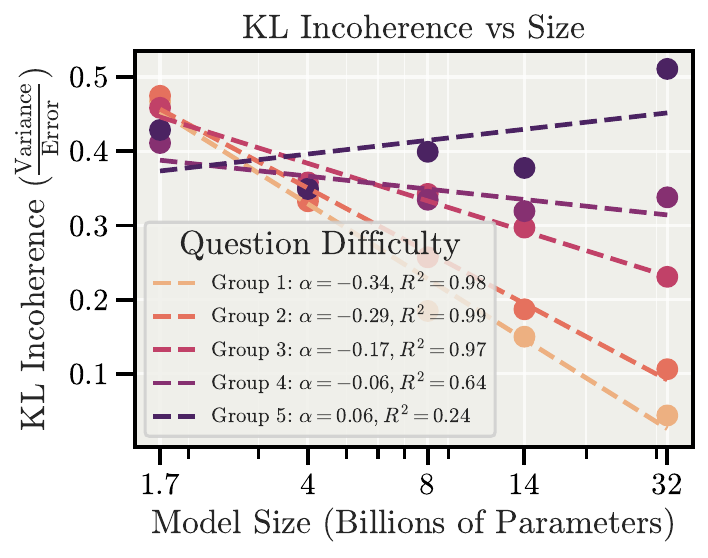} %
     \caption{\centering \qwen{} on \mmlu{}}
     \label{fig:qwen_mmlu_incoherence}
   \end{subfigure}
       \hfill
   \begin{subfigure}[t]{0.33\textwidth}
     \centering
     \includegraphics[width=\linewidth]{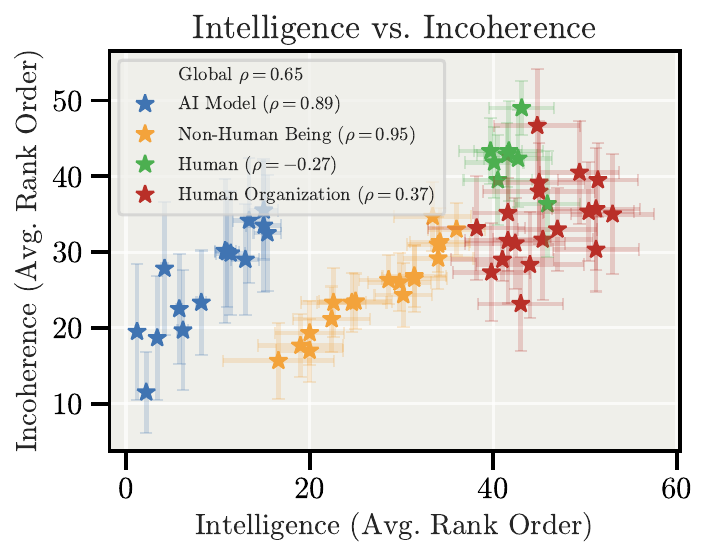}
     \caption{Survey Ranking Results}
     \label{fig:survey_intelligence_incoherence}
   \end{subfigure}\hfill
   \begin{subfigure}[t]{0.33\textwidth}
     \centering
     \includegraphics[width=\linewidth]{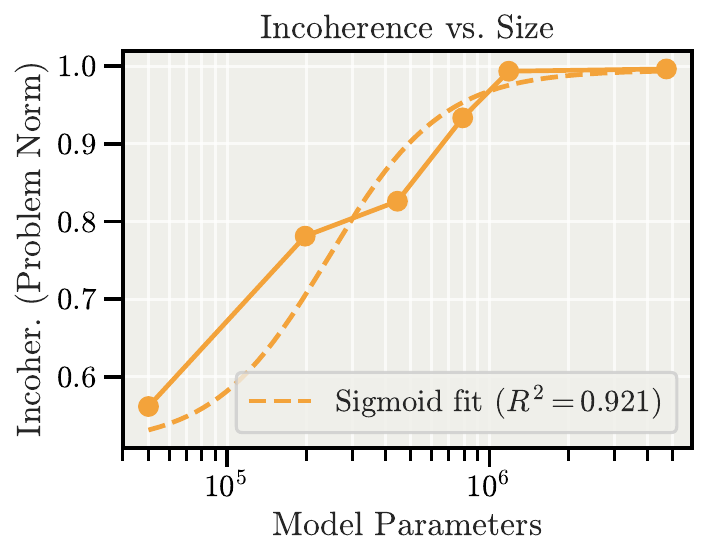} %
     \caption{Synthetic Optimizers}
     \label{fig:synthetic_incoherence}
   \end{subfigure}\hfill
    \caption{
    \looseness-1
    {\textbf{Larger and more intelligent systems are often more incoherent.} 
    \emph{(a)} We measure the scaling of error-incoherence vs. model size for the \qwen{} family, as a function of question difficulty on \mmlu{}. For easy questions, error-incoherence drops with model scale, while for the hardest questions error-incoherence remains constant or increases with model scale. The expanded results for this experiment are in Fig.~\ref{fig:grouping_by_reasoning_length}.
    \emph{(b)} Disjoint sets of human subjects were tasked with subjectively ranking the intelligence and incoherence of diverse AI models, non-human beings, well known humans, and human organizations. Across all categories, entities that were judged more intelligent by one group of subjects, were independently judged to be more incoherent by another group of subjects. See Appx.~\ref{appx:experimental_details_survey}. 
    \emph{(c)} In a synthetic task, we train transformers of increasing size to explicitly emulate optimizer trajectories descending a quadratic loss. As these models become larger, the trajectories they generate achieve lower loss on the quadratic. However, the final loss is also more variance dominated and thus incoherent with increasing model size. 
    Details in Fig.~\ref{fig:synthetic_scaling}. 
    }
    }   
    \vspacepostcaption{}
    \label{fig:larger_systems_intelligence_incoherence}
 \end{figure}
\looseness-1
\textbf{Motivation.} In Section~\ref{sec:scaling_laws_reasoning_length}, in particular Fig.~\ref{fig:fig_2_gpqa}, we fix a model and analyze error-incoherence as a function of reasoning length. Now, we ask a different question: \emph{When we fix a task, how does error-incoherence change as a function of model size? How does incoherence scale with intelligence?} 

\textbf{Overview.} We summarize the main observation in Fig.~\ref{fig:larger_systems_intelligence_incoherence}: larger, more capable and intelligent systems are often more incoherent. This is manifested in LLMs for the most complex set of questions (Sect.~\ref{sec:scaling_laws_model_size_qwen}), the rankings of intelligence and error-incoherence as judged by human survey participants (Appx.~\ref{appx:experimental_details_survey}) and our synthetic optimizer setting (Sect.~\ref{sec:synthetic}). 
However, we find that larger models are less incoherent on simpler questions (Sect.~\ref{sec:scaling_laws_model_size_qwen}).
We discuss each result in detail.

\vspacepresection{}
\subsubsection{Scaling Laws for LLMs Separated by Task Complexity}
\label{sec:scaling_laws_model_size_qwen}
\quoteit{Easy tasks become less incoherent with scale, while harder tasks become more incoherent.}
\textbf{Overview.} We experiment with the \qwen{} model family, as they provide the same model architecture, including reasoning abilities, with up to $32$B parameters. Consistent with other setups, we sample many responses for the same set of questions. Additionally, we cluster questions using the the reasoning length of a reference model (here: $32$B) into equally sized groups. 

\begin{figure}[t]
    \hfill
    \begin{subfigure}[t]{0.33\textwidth}
     \centering
     \includegraphics[width=\linewidth]{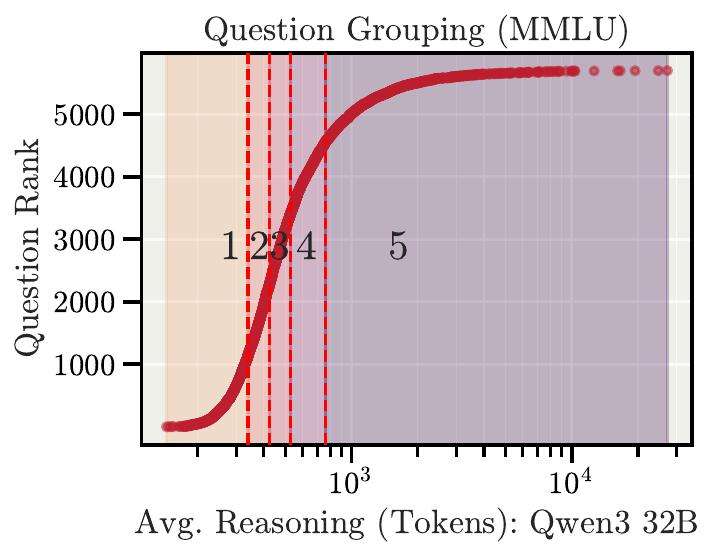} %
     \caption{Separating Complexity Groups}
     \label{fig:mmlu_a}
   \end{subfigure}\hfill
   \begin{subfigure}[t]{0.33\textwidth}
     \centering
     \includegraphics[width=\linewidth]{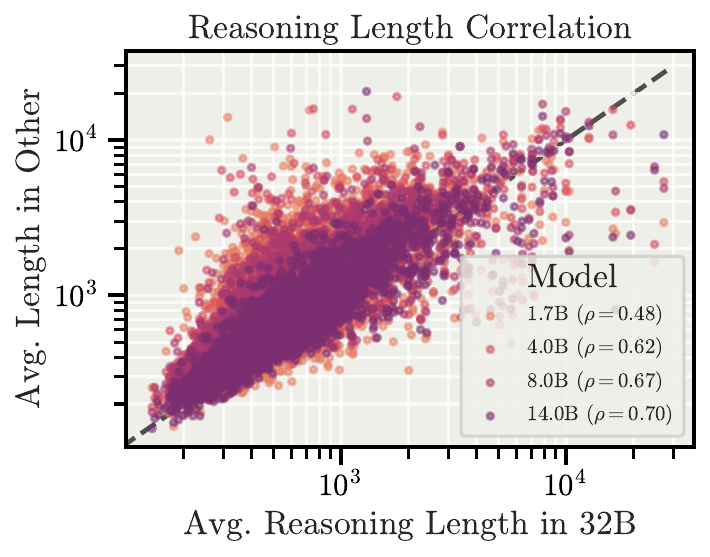}
     \caption{Length Correlation}
     \label{fig:mmlu_b}
   \end{subfigure}\hfill
   \begin{subfigure}[t]{0.33\textwidth}
     \centering
     \includegraphics[width=\linewidth]{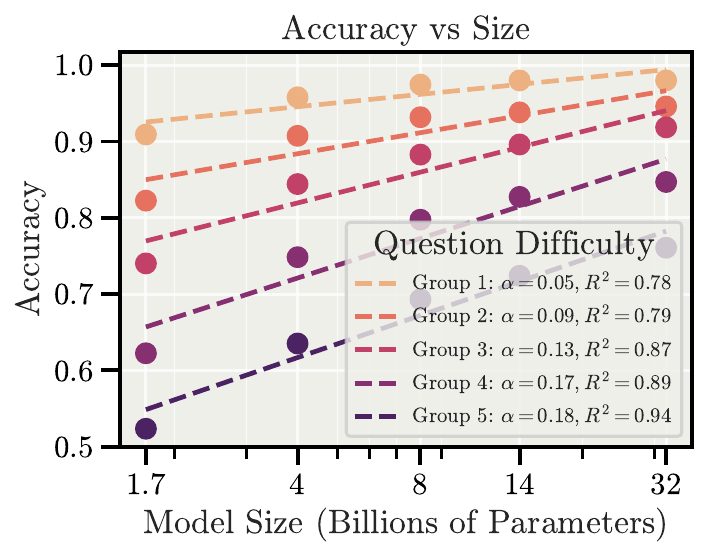}
     \caption{\centering Accuracy Scaling Laws}
     \label{fig:mmlu_c}
   \end{subfigure}
   \hfill
   \begin{subfigure}[t]{0.66\textwidth}
     \centering
     \includegraphics[width=\linewidth]{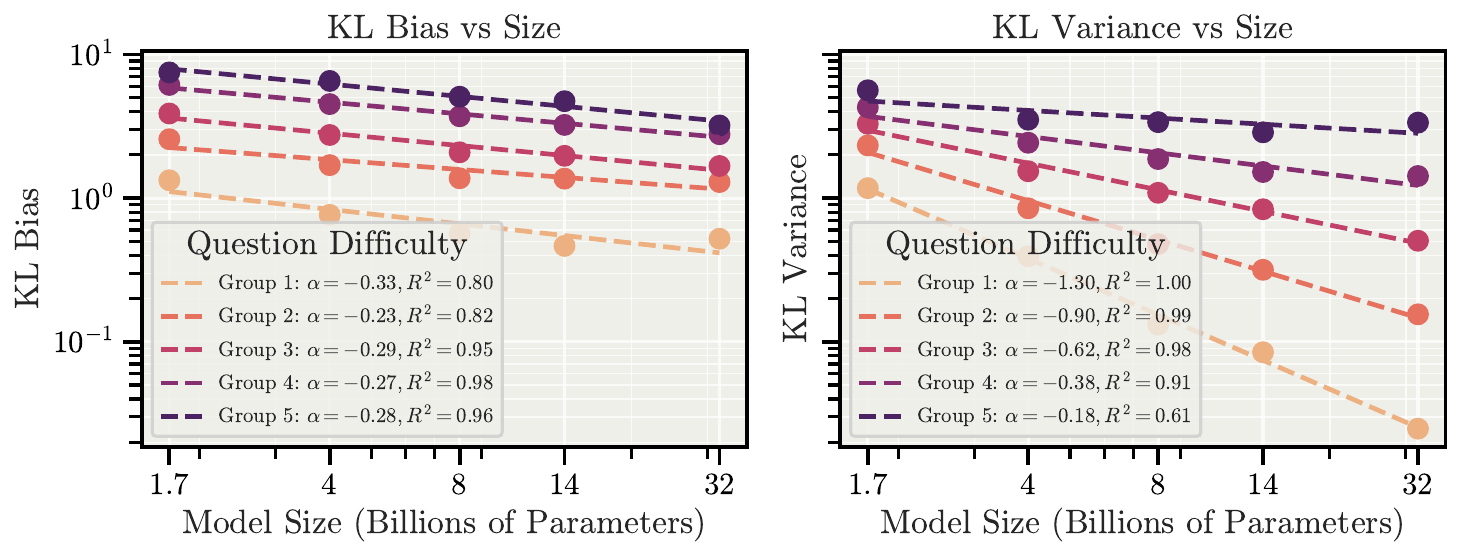}
     \caption{Bias and Variance Scaling Laws}
     \label{fig:mmlu_d}
   \end{subfigure}\hfill
    \begin{subfigure}[t]{0.33\textwidth}
     \centering
     \includegraphics[width=\linewidth]{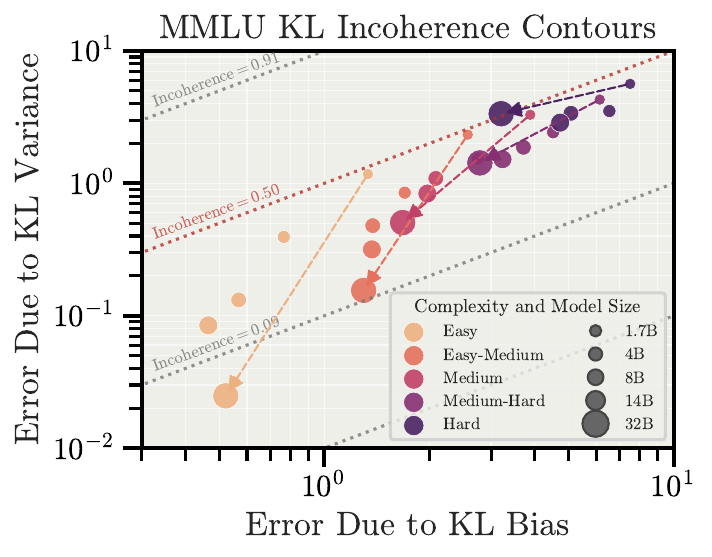}
     \caption{Error-incoherence}
     \label{fig:mmlu_e}
     \hfill
   \end{subfigure}
    \vspaceprecaption{}
    \caption{
    \looseness-1
    \textbf{Details for \qwen{} scaling laws: easy tasks become less incoherent, harder tasks more incoherent.
    } 
    We group \mmlu{} questions by reasoning length using a reference model (Qwen3 32B, \emph{(a)}), which correlates across model sizes \emph{(b)} and serves as a task complexity proxy, as accuracy drops with longer reasoning \emph{(c)}. These groups reveal distinct bias–variance scaling \emph{(d)}: bias slopes are similar across groups, but variance slopes decrease sharply for harder ones. In the hardest group, variance slopes fall below bias slopes, leaving variance as the limiting factor. Thus, larger models remain constrained by variance and \emph{more incoherent with scale} \emph{(e)}. We provide more analyses including other models and the same conclusion for \gpqa{} in Appx.~\ref{appx:more_results_gpqa_scaling}.
    \vspacepostcaption{}
    }    
    \label{fig:grouping_by_reasoning_length}
 \end{figure}

\looseness-1
\textbf{Results.} See Fig.~\ref{fig:grouping_by_reasoning_length} for the detailed results. 
We find that performance consistently improves with increasing model size, with the fastest rate of improvement for the hardest questions. However,
the way in which error-incoherence changes with scale depends on question difficulty: Model responses to easy questions become more coherent with scale, while responses to the hardest questions become more incoherent with scale, though this last trend is noisy.

\looseness-1
\textbf{Further results.} We provide different visualizations of the same results in Appx.~\ref{appx:more_results_gpqa_scaling}, which include the same results for \gpqa{} (Fig.~\ref{fig:grouping_by_reasoning_length_gpqa}), the relationship between error-incoherence and error (Fig.~\ref{fig:incoherence_iso_plots}) and how reasoning length is a stronger indicator of error-incoherence than model size (Fig.~\ref{fig:mmlu_incoherence_contour}).

\vspacepresection{}
\subsubsection{Scaling Laws in Controlled Synthetic Settings: Models as Optimizers}
\label{sec:synthetic}
\quoteit{On a synthetic task, models become more incoherent as they are made larger.
}
\looseness-1
\textbf{Models as optimizers.} 
In this paper, we are trying to disentangle whether capable models will more tend to act as effective optimizers of the wrong goal, or will pursue the right goal but not be effective optimizers. 
To quantify this in a controlled setting, 
we train models to literally mimic the trajectory of a hand-coded optimizer descending a loss function. 
This can be viewed as trying to train a model to implement a mesa-optimizers \citep{hubinger2019risks}. 
We then analyze the bias and variance of the resulting models, to answer the question:
\emph{Does the model become an optimizer faster or slower than it converges on the right optimization objective?}

 \begin{figure}[t]
    \centering
    \includegraphics[width=0.36\textwidth]{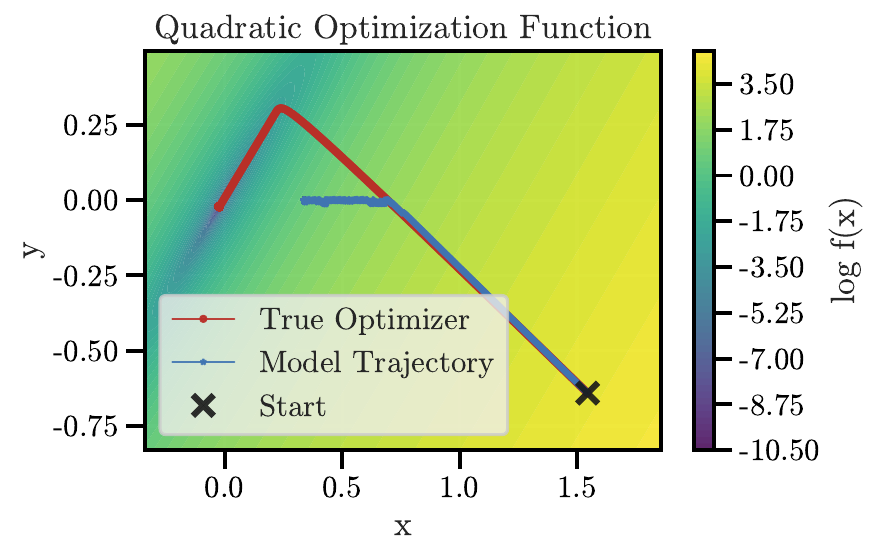}
    \includegraphics[width=0.3\textwidth]{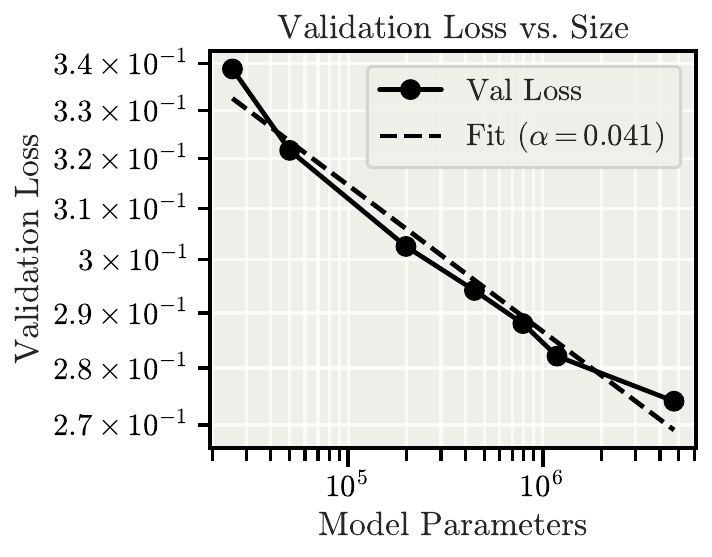}
    \includegraphics[width=0.3\textwidth]{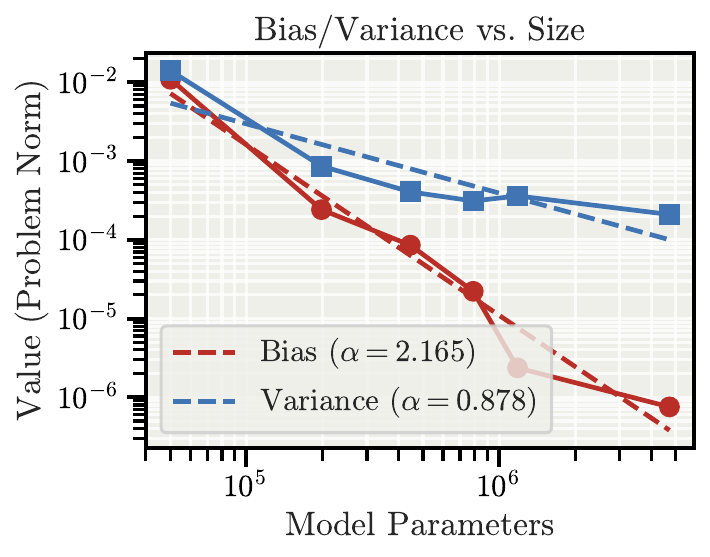}
    \vspaceprecaption{}
    \caption{
    \looseness-1
    \textbf{Details for synthetic optimization: In controlled settings with teacher forcing and a single objective, language models become variance dominated with increasing size.} (\emph{left}) We train autoregressive transformers to predict update steps to minimize a quadratic function using decoding based regression, \ie next-token prediction. This setting involves sequentially performing steps towards a goal via next token prediction, emulating a key feature of goal seeking AI.  
    (\emph{middle}) The loss (next-token prediction objective) follows a clear power law improvement with model size.
    (\emph{right}) 
    When evaluating the trained models using their own rollouts, we 
    find that increasing model size reduces bias much faster than variance. 
    \vspacepostcaption{}
    }
    \label{fig:synthetic_scaling}
 \end{figure}

\looseness-1
\textbf{Setup.} We study a simple $d$-dimensional quadratic function of the form $f(x) = \frac{1}{2}(x-b)^{T}A(x-b)$, where $A\in\mathbb{R}^{d\times d}$ is a (random) positive-definite but ill-conditioned matrix. We set the condition number to $50$.
Training data is generated by using 
an optimizer to produce many trajectories of fixed length for random initial points.
The optimizer used to generate the training data performs steepest descent with a fixed step norm. 
The training dataset consists of pairs $(x_i, u_i)$, where $x_i$ is a parameter iterate, and $u_i$ is the corresponding update step generated by the optimizer. 
Analogously to real (token-based) models, we train transformer models \citep{vaswani2017attention} of varying sizes using \emph{decoding-based regression} \citep{song2025decoding} and teacher forcing. 
This means we tokenize the scientific format representation of $x_i$ and $u_i$, with a vocabulary of  digits and signs. 
When evaluating, we sample multiple initial points and roll out trajectories using the model's own predictions.
A visualization of this with a real model is provided in Fig.~\ref{fig:synthetic_scaling} (left). The bias and variance measures are then taken \wrt{} the optimum and norm $\lVert\cdot\rVert_A$ that is induced by the problem.
The details are in Appx.~\ref{appx:experimental_details_synth}.

\looseness-1
\textbf{Results.} 
The main results are shown in Fig.~\ref{fig:fig_2_synthetic} (error-incoherence over rollout steps) and Fig.~\ref{fig:synthetic_scaling} (scaling laws by size). All models show consistently rising error-incoherence per step; interestingly, smaller models reach a lower plateau after a tipping point where they can no longer follow the correct trajectory and stagnate, reducing variance. This pattern also appears in individual bias and variance curves (Fig.~\ref{fig:example_trajectory_all_models}). Importantly, larger models reduce bias more than variance. These results suggest that they learn the correct objective faster than the ability to maintain long coherent action sequences.
More results and discussions are provided in
Appx.~\ref{appx:more_results_synth}.

\vspacepresection{}
\subsection{The Effects of Reasoning Budget and Ensembling}
\label{sec:action_complexity}
\looseness-1
We now study the effect of reasoning budgets, \ie the techniques provided in model APIs, and ensembling, \ie averaging multiple responses, on error-incoherence. The main results are in Fig.~\ref{fig:error_correction}.

\vspacepresection{}
\subsubsection{Reasoning budgets}
\label{sec:action_complexity_budget}
\quoteit{Reasoning budgets reduce error-incoherence, but natural variation has a much stronger effect.}
\begin{figure}[t]
\begin{subfigure}[b]{.32\textwidth}
    \centering
    \includegraphics[width=.98\linewidth]{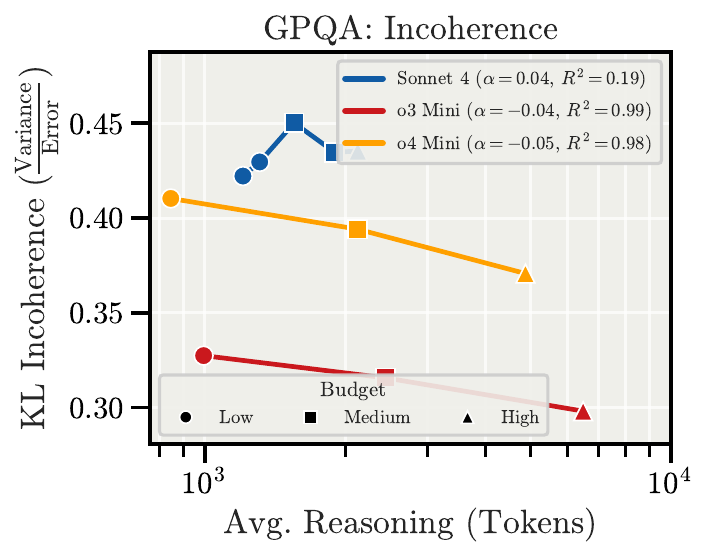}
    \caption{Reasoning Budgets}
    \label{fig:error_correction_budgets}
\end{subfigure}
\begin{subfigure}[b]{.65\textwidth}
    \centering
    \includegraphics[width=0.49\linewidth]{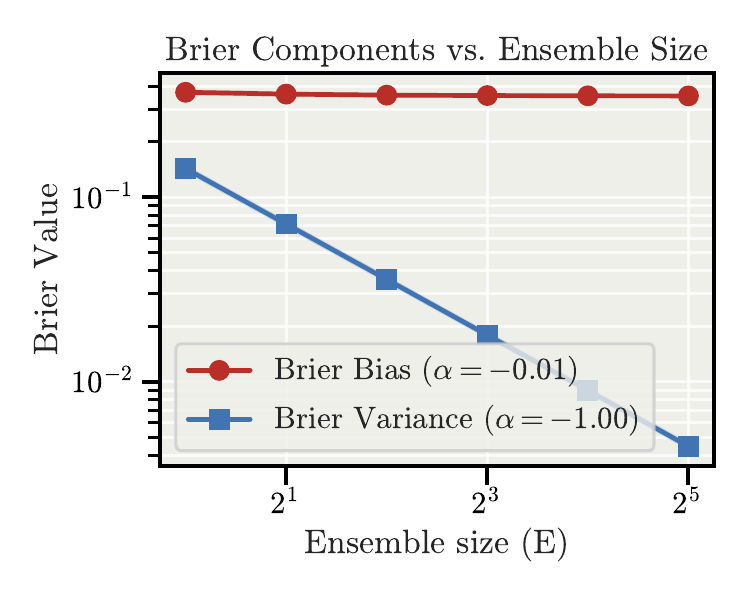}
    \includegraphics[width=0.49\linewidth]{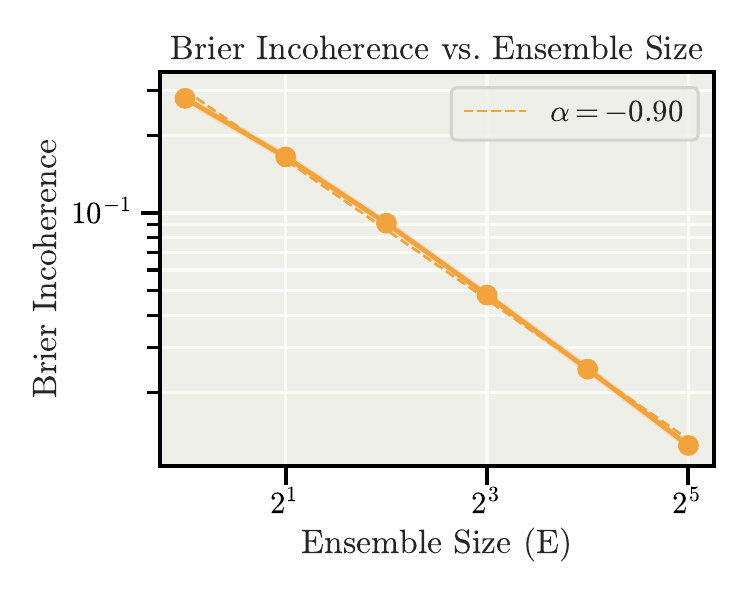}
    \vspace{-3pt}
    \caption{Ensembling Results}
    \label{fig:error_correction_ensembling}
\end{subfigure}
      \caption{
      \looseness-1
      \textbf{
      Ensembling and larger reasoning budgets
      reduce error-incoherence. 
      Other forms of error correction may also reduce error-incoherence.
      } 
       \emph{(a)} Instructing models to reason longer improves performance (inference scaling laws, Fig.~\ref{fig:action_complexity}) and sometimes error-incoherence. This effect is smaller than natural variation, where error-incoherence rises sharply (Fig.~\ref{fig:natural_variation}; direct comparison in Fig.~\ref{fig:action_complexity}). \emph{(b)} With \omini4 on \gpqa{}, we analyze the effect of the \emph{ensembling}, \ie using multiple samples to average output probabilities over targets for the same question. The bias and variance are now computed by comparing different ensembles of the same size. We find that, as expected from theory, it 
       reduces variance with a rate of $1/E$, without affecting bias (\emph{left}). As a consequence, error-incoherence drops (\emph{right}).    
    Ensembling is a particular form of model error correction, which is impractical for action loops in the world, since state can typically not be reset. However, we expect other error correction techniques to 
    also
    reduce
    incoherence.
    \vspacepostcaption{}
    }
      \label{fig:error_correction}
 \end{figure}

\looseness-1
\textbf{Inference scaling.} We show the results of our inference-scaling analysis on \gpqa{} in Fig.~\ref{fig:error_correction_budgets} and Fig.~\ref{fig:action_complexity}. Increasing reasoning budgets improves performance (\ref{fig:action_complexity_gpqa}, left), and slightly reduces error-incoherence for all models but \sonnet{} (\ref{fig:error_correction_budgets}). Interestingly, this effect is overshadowed by error-incoherence that arises through natural variation, \ie when models think longer than the median for a question (recall analysis in Fig.~\ref{fig:natural_variation}; direct comparison in Fig.~\ref{fig:action_complexity_gpqa}, right).

\looseness-1
\textbf{Discussion: How does reasoning budget improve coherence?} Since the implementation details of reasoning budgets for frontier models are not public, it is unclear how exactly it can improve error-incoherence. We believe it is likely explained by better backtracking and error correction properties, a phenomena observed to arise during training with larger budgets \citep{guo2025deepseek}, and related
to the ensembling results in Sec.~\ref{sec:ensembling}. We partially explore error-incoherence through the reasoning structure with the \qwen{} reasoning traces in Appx.~\ref{appx:error_correction}.

\vspacepresection{}
\subsubsection{Ensembling}
\label{sec:ensembling}
\quoteit{Ensembling multiple attempts reduces error-incoherence.}
\textbf{Motivation.} Perhaps the most natural way to reduce error-incoherence is to ensemble multiple attempts: instead of relying on a single answer, we roll out multiple trajectories 
and 
combine them. 
We demonstrate this with a repetition of the experiment for \gpqa{} with \omini4.

\looseness-1
\textbf{Setup.} We obtain $320$ samples of answers for all questions of \gpqa{}. Fixing an ensemble of size $E$, we average the $E$ produced probabilities over targets. To compute bias and variance, we then compare ensembles of the same size across random samples of ensembles, which we hold at a fixed number of $10$, while ensuring that samples do not overlap. This allows ensemble sizes of up to $32$.

\looseness-1
\textbf{Results.} 
Fig.~\ref{fig:error_correction_ensembling} shows how variance changes with increasing ensemble size. As expected, it drops like the inverse of the ensemble size, and
error-incoherence therefore also drops.
We expect there are broader classes of error correction that behave similarly. The slight reduction in error-incoherence with increasing reasoning budgets in Sec.\ref{sec:action_complexity_budget} may be achieved through such a mechanism.
We provide the plots for \textsc{KL-Error-incoherence} in Fig.~\ref{fig:ensembling_kl}.

\vspacepresection{}
\section{Related Work}
\label{sec:related_work}
We summarize the most important related work and defer a comprehensive discussion to Appx.~\ref{app:extended_related_work}.

\looseness-1
\textbf{Reasoning.}
Recent studies report inverse scaling trends with extended reasoning degrading performance~\citep{gema2025inverse,su2025between,wu2025more,hassid2025don}.
Most relevant, \citet{ghosal2025does} find that overthinking increases output variance, though via artificially injected tokens rather than natural overthinking. While these studies identify performance degradation, they do not distinguish systematic errors from inconsistent failures. Our ensembling analysis relates to self-consistency work \citep{wang2023selfconsistency}, but reframes aggregation as reducing error-incoherence.

\looseness-1
\textbf{Evaluation variance.}
Even though AI models have vastly improved upon benchmarks, evaluations are known to be highly variant~\citep{bui2025assessing,biderman2024lessons,romanou2026brittlebench}.
\citet{errica2024did} formalize this through sensitivity and consistency metrics, revealing important failure modes. This is similar setup to our input and output randomness. Importantly, we connect the variability to the concepts of bias and variance, highlighting the relevance in the safety setting, and analyze scaling laws.

\looseness-1
\textbf{Scaling behavior.} As models get larger and more capable, evidence suggests their representation and errors become highly aligned \citep{kim2025correlated,huh2024platonic,goel2025great} and that they improve long-horizon tasks \citep{sinha2025illusion}.
Our work complements these observations by finding increased error-incoherence the longer models reason and act, aligned between model families.

\vspacepresection{}
\section{Discussion and What Our Results Do Not Tell Us}
\label{sec:discussion}
\looseness-1
\textbf{Why expect more capable models to be more incoherent?} 
In this paper, we do not experimentally or theoretically explore the specific mechanisms for increasing error-incoherence with increasing trajectory length and (sometimes) model size. However, there are motivating observations.

The first is that LLMs are dynamical systems. 
When they generate text or take actions, they trace trajectories in a high-dimensional state space. 
It is often \emph{very hard} to constrain a generic dynamical system to act as an optimizer. 
The set of dynamical systems that act as optimizers of a fixed loss is measure zero in the space of all dynamical systems. 
As models scale and acquire broader capabilities, their effective state and action space expands, exacerbating this difficulty. We should not expect AIs to act as optimizers without considerable effort, nor should we expect this to be easier than training other properties into their dynamics. 

Second, variance typically accumulates over a trajectory unless there is an active correction mechanism (like ensembling, Fig.~\ref{fig:error_correction}). When an AI acts in the real world, actions are often irreversible.
Therefore, it will often be impossible or impractical to correct for noise introduced by model actions.

\looseness-1
\textbf{Reward misspecification.} 
Bias can be further decomposed into $\textsc{Bias } = \textsc{Bias$_\text{mesa} $}%
+ \textsc{Bias$_\text{spec}$ }$, where $\textsc{Bias$_\text{mesa}$ }$ captures the average deviation of the model's behavior from the training objective, and $\textsc{Bias$_\text{spec}$ }$ captures the deviation of the training objective from the \emph{intended} training objective. 
For our tasks, we believe that there was not meaningful reward misspecification. 
In settings with poorly specified training objectives, we worry that 
$\textsc{Bias$_\text{spec}$ }$ would come to dominate the error, as both variance and $\textsc{Bias$_\text{mesa}$ }$ go to zero with increasing model capability. 
Our results underscore the importance of 
characterizing and mitigating goal misspecification during training.

\textbf{Open-ended goals and error-incoherence.}
\looseness-1
To rigorously analyze the scaling of bias, variance, and error-incoherence, we need to (1) measure an ``average'' prediction (for bias and variance) and (2) measure distance to ground truth (for bias). We use multiple-choice classification, coding unit-tests, and objective functions rather than LLM judges to ensure metrics are well-defined, unbiased, and comparable. Extracting hidden goals and complex incoherent behaviors remains important \citep[cf. Section 4.1.1.5;][]{claude4systemcard}; our embedding-variance analysis of model-written evals (Appx.\ref{appx:more_results_mwe}) provides an initial exploration of a setting where bias is not easily defined or measured.

\vspacepresection{}
\section{Conclusion}
\label{sec:conclusion}
\looseness-1
Motivated by the hot mess theory of AI misalignment, we propose a bias–variance decomposition as a framework for analyzing how increasingly capable AIs will fail. Our results show that longer sequences of reasoning and actions consistently increase model error-incoherence. We also find that 
smarter
AI models are not consistently more coherent.
Our results suggest that when advanced AI systems performing complex tasks fail, it is likely to be in inconsistent ways that do not correspond to pursuit of any stable goal. 
This should inform judgements of the relative plausibility of different AI risk scenarios and guide further research into understanding the mechanistic origins of error-incoherence.

\section*{Acknowledgements}

We thank Andrew Saxe, Brian Cheung, Kit Frasier-Taliente, Igor Shilov, Stewart Slocum, Aidan Ewart, David Duvenaud, and Tom Adamczewski for  extremely helpful discussions on topics and results in this paper.

\section*{Ethics Statement}
This research aims to characterize failure modes of increasingly capable AI systems to inform safer deployment strategies.
Our findings suggest that as AI systems tackle more complex tasks requiring extended reasoning, incoherent failures become more prevalent than systematic misalignment.
While this work does not directly prevent AI failures, it offers empirical grounding for prioritizing safety interventions, suggesting greater focus on preventing unpredictable accidents rather than solely defending against coherent malicious behavior.
We believe this understanding of AI failure modes benefits the community to ensure safe AI deployment.

\section*{Reproducibility Statement}
We provide a detailed description of our theoretical framework in Section~\ref{sec:background:bv_decomp} and Appx.~\ref{appx:definitions}. The general experimental setups are described in Section~\ref{sec:experiments} and Appx.~\ref{appx:experimental_details}, with task-specific details outlined in each experiment subsections. Our code and data is available \href{https://github.com/haeggee/hot-mess-of-ai}{here}.

\bibliography{iclr2026_conference}

@article{domingos2000unified,
  title={A unified bias-variance decomposition for zero-one and squared loss},
  author={Domingos, Pedro},
  journal={AAAI/IAAI},
  volume={2000},
  pages={564--569},
  year={2000}
}

@inproceedings{kohavi1996bias,
author = {Kohavi, Ron and Wolpert, David},
title = {Bias plus variance decomposition for zero-one loss functions},
year = {1996},
isbn = {1558604197},
publisher = {Morgan Kaufmann Publishers Inc.},
address = {San Francisco, CA, USA},
booktitle = {Proceedings of the Thirteenth International Conference on International Conference on Machine Learning},
pages = {275–283},
numpages = {9},
location = {Bari, Italy},
series = {ICML'96}
}

@incollection{kong1995error,
  title={Error-correcting output coding corrects bias and variance},
  author={Kong, Eun Bae and Dietterich, Thomas G},
  booktitle={Machine learning proceedings 1995},
  pages={313--321},
  year={1995},
  publisher={Elsevier}
}

@article{breiman1996bias,
  title={Bias, variance, and arcing classifiers},
  author={Breiman, Leo},
  year={1996},
  publisher={Tech. Rep. 460, Statistics Department, University of California, Berkeley~…}
}

@article{tibshirani1996bias,
  title={Bias, variance and prediction error for classification rules},
  author={Tibshirani, Robert},
  journal={Technical Report, Statistics Department, University of Toronto},
  year={1996}
}

@article{friedman1997bias,
  title={On bias, variance, 0/1—loss, and the curse-of-dimensionality},
  author={Friedman, Jerome H},
  journal={Data mining and knowledge discovery},
  volume={1},
  number={1},
  pages={55--77},
  year={1997},
  publisher={Springer}
}

@article{degroot2018comparison,
    author = {Degroot, Morris H. and Fienberg, Stephen E.},
    title = {The Comparison and Evaluation of Forecasters},
    journal = {Journal of the Royal Statistical Society Series D: The Statistician},
    volume = {32},
    number = {1-2},
    pages = {12-22},
    year = {2018},
    month = {12},
    issn = {2515-7884},
    doi = {10.2307/2987588},
    url = {https://doi.org/10.2307/2987588},
    eprint = {https://academic.oup.com/jrsssd/article-pdf/32/1-2/12/49920469/jrsssd_32_1-2_12.pdf},
}

@ARTICLE{heskes1998bias,
  author={Heskes, Tom},
  journal={Neural Computation}, 
  title={Bias/Variance Decompositions for Likelihood-Based Estimators}, 
  year={1998},
  volume={10},
  number={6},
  pages={1425-1433},
  doi={10.1162/089976698300017232}
}

@article{kaplan2020scaling,
  title={Scaling laws for neural language models},
  author={Kaplan, Jared and McCandlish, Sam and Henighan, Tom and Brown, Tom B and Chess, Benjamin and Child, Rewon and Gray, Scott and Radford, Alec and Wu, Jeffrey and Amodei, Dario},
  journal={arXiv preprint arXiv:2001.08361},
  year={2020}
}

@inproceedings{hoffmann2022training,
author = {Hoffmann, Jordan and Borgeaud, Sebastian and Mensch, Arthur and Buchatskaya, Elena and Cai, Trevor and Rutherford, Eliza and de Las Casas, Diego and Hendricks, Lisa Anne and Welbl, Johannes and Clark, Aidan and Hennigan, Tom and Noland, Eric and Millican, Katie and van den Driessche, George and Damoc, Bogdan and Guy, Aurelia and Osindero, Simon and Simonyan, Karen and Elsen, Erich and Vinyals, Oriol and Rae, Jack W. and Sifre, Laurent},
title = {Training compute-optimal large language models},
year = {2022},
isbn = {9781713871088},
publisher = {Curran Associates Inc.},
address = {Red Hook, NY, USA},
booktitle = {Proceedings of the 36th International Conference on Neural Information Processing Systems},
articleno = {2176},
numpages = {15},
location = {New Orleans, LA, USA},
series = {NIPS '22}
}

@inproceedings{bui2025assessing,
    title = "Assessing the Macro and Micro Effects of Random Seeds on Fine-Tuning Large Language Models",
    author = "Bui, Nghia Tuan  and
      Savova, Guergana K  and
      Wang, Lijing",
    editor = "Inui, Kentaro  and
      Sakti, Sakriani  and
      Wang, Haofen  and
      Wong, Derek F.  and
      Bhattacharyya, Pushpak  and
      Banerjee, Biplab  and
      Ekbal, Asif  and
      Chakraborty, Tanmoy  and
      Singh, Dhirendra Pratap",
    booktitle = "Proceedings of the 14th International Joint Conference on Natural Language Processing and the 4th Conference of the Asia-Pacific Chapter of the Association for Computational Linguistics",
    month = dec,
    year = "2025",
    address = "Mumbai, India",
    publisher = "The Asian Federation of Natural Language Processing and The Association for Computational Linguistics",
    url = "https://aclanthology.org/2025.ijcnlp-short.3/",
    pages = "41--46",
    ISBN = "979-8-89176-299-2"
}

@inproceedings{errica2024did,
    title = "What Did {I} Do Wrong? Quantifying {LLM}s' Sensitivity and Consistency to Prompt Engineering",
    author = "Errica, Federico  and
      Sanvito, Davide  and
      Siracusano, Giuseppe  and
      Bifulco, Roberto",
    editor = "Chiruzzo, Luis  and
      Ritter, Alan  and
      Wang, Lu",
    booktitle = "Proceedings of the 2025 Conference of the Nations of the Americas Chapter of the Association for Computational Linguistics: Human Language Technologies (Volume 1: Long Papers)",
    month = apr,
    year = "2025",
    address = "Albuquerque, New Mexico",
    publisher = "Association for Computational Linguistics",
    url = "https://aclanthology.org/2025.naacl-long.73/",
    doi = "10.18653/v1/2025.naacl-long.73",
    pages = "1543--1558",
    ISBN = "979-8-89176-189-6"
}

@article{biderman2024lessons,
  title={Lessons from the trenches on reproducible evaluation of language models},
  author={Biderman, Stella and Schoelkopf, Hailey and Sutawika, Lintang and Gao, Leo and Tow, Jonathan and Abbasi, Baber and Aji, Alham Fikri and Ammanamanchi, Pawan Sasanka and Black, Sidney and Clive, Jordan and others},
  journal={arXiv preprint arXiv:2405.14782},
  year={2024}
}

@inproceedings{perez-etal-2023-discovering,
    title = "Discovering Language Model Behaviors with Model-Written Evaluations",
    author = "Perez, Ethan  and
      Ringer, Sam  and
      Lukosiute, Kamile  and
      Nguyen, Karina  and
      Chen, Edwin  and
      Heiner, Scott  and
      Pettit, Craig  and
      Olsson, Catherine  and
      Kundu, Sandipan  and
      Kadavath, Saurav  and
      Jones, Andy  and
      Chen, Anna  and
      Mann, Benjamin  and
      Israel, Brian  and
      Seethor, Bryan  and
      McKinnon, Cameron  and
      Olah, Christopher  and
      Yan, Da  and
      Amodei, Daniela  and
      Amodei, Dario  and
      Drain, Dawn  and
      Li, Dustin  and
      Tran-Johnson, Eli  and
      Khundadze, Guro  and
      Kernion, Jackson  and
      Landis, James  and
      Kerr, Jamie  and
      Mueller, Jared  and
      Hyun, Jeeyoon  and
      Landau, Joshua  and
      Ndousse, Kamal  and
      Goldberg, Landon  and
      Lovitt, Liane  and
      Lucas, Martin  and
      Sellitto, Michael  and
      Zhang, Miranda  and
      Kingsland, Neerav  and
      Elhage, Nelson  and
      Joseph, Nicholas  and
      Mercado, Noemi  and
      DasSarma, Nova  and
      Rausch, Oliver  and
      Larson, Robin  and
      McCandlish, Sam  and
      Johnston, Scott  and
      Kravec, Shauna  and
      El Showk, Sheer  and
      Lanham, Tamera  and
      Telleen-Lawton, Timothy  and
      Brown, Tom  and
      Henighan, Tom  and
      Hume, Tristan  and
      Bai, Yuntao  and
      Hatfield-Dodds, Zac  and
      Clark, Jack  and
      Bowman, Samuel R.  and
      Askell, Amanda  and
      Grosse, Roger  and
      Hernandez, Danny  and
      Ganguli, Deep  and
      Hubinger, Evan  and
      Schiefer, Nicholas  and
      Kaplan, Jared",
    editor = "Rogers, Anna  and
      Boyd-Graber, Jordan  and
      Okazaki, Naoaki",
    booktitle = "Findings of the Association for Computational Linguistics: ACL 2023",
    month = jul,
    year = "2023",
    address = "Toronto, Canada",
    publisher = "Association for Computational Linguistics",
    url = "https://aclanthology.org/2023.findings-acl.847/",
    doi = "10.18653/v1/2023.findings-acl.847",
    pages = "13387--13434"
}

@inproceedings{
    jimenez2023swe,
    title={{SWE}-bench: Can Language Models Resolve Real-world Github Issues?},
    author={Carlos E Jimenez and John Yang and Alexander Wettig and Shunyu Yao and Kexin Pei and Ofir Press and Karthik R Narasimhan},
    booktitle={The Twelfth International Conference on Learning Representations},
    year={2024},
    url={https://openreview.net/forum?id=VTF8yNQM66}
}

@article{schmied2025llms,
  title={Llms are greedy agents: Effects of rl fine-tuning on decision-making abilities},
  author={Schmied, Thomas and Bornschein, J{\"o}rg and Grau-Moya, Jordi and Wulfmeier, Markus and Pascanu, Razvan},
  journal={arXiv preprint arXiv:2504.16078},
  year={2025}
}

@article{gema2025inverse,
title={Inverse Scaling in Test-Time Compute},
author={Aryo Pradipta Gema and Alexander H{\"a}gele and Runjin Chen and Andy Arditi and Jacob Goldman-Wetzler and Kit Fraser-Taliente and Henry Sleight and Linda Petrini and Julian Michael and Beatrice Alex and Pasquale Minervini and Yanda Chen and Joe Benton and Ethan Perez},
journal={Transactions on Machine Learning Research},
issn={2835-8856},
year={2025},
url={https://openreview.net/forum?id=NXgyHW1c7M},
note={Featured Certification, J2C Certification}
}

@inproceedings{ghosal2025does,
title={Does Thinking More Always Help? Mirage of Test-Time Scaling in Reasoning Models},
author={Soumya Suvra Ghosal and Souradip Chakraborty and Avinash Reddy and Yifu Lu and Mengdi Wang and Dinesh Manocha and Furong Huang and Mohammad Ghavamzadeh and Amrit Singh Bedi},
booktitle={The Thirty-ninth Annual Conference on Neural Information Processing Systems},
year={2025},
url={https://openreview.net/forum?id=tKPqbamNb9}
}

@article{su2025between,
  title={Between Underthinking and Overthinking: An Empirical Study of Reasoning Length and correctness in LLMs},
  author={Su, Jinyan and Healey, Jennifer and Nakov, Preslav and Cardie, Claire},
  journal={arXiv preprint arXiv:2505.00127},
  year={2025}
}

@inproceedings{ma2025step,
  title={What are step-level reward models rewarding? counterintuitive findings from mcts-boosted mathematical reasoning},
  author={Ma, Yiran and Chen, Zui and Liu, Tianqiao and Tian, Mi and Liu, Zhuo and Liu, Zitao and Luo, Weiqi},
  booktitle={Proceedings of the AAAI Conference on Artificial Intelligence},
  volume={39},
  pages={24812--24820},
  year={2025}
}

@inproceedings{
yang2025towards,
title={Towards Thinking-Optimal Scaling of Test-Time Compute for {LLM} Reasoning},
author={Wenkai Yang and Shuming Ma and Yankai Lin and Furu Wei},
booktitle={The Thirty-ninth Annual Conference on Neural Information Processing Systems},
year={2025},
url={https://openreview.net/forum?id=6ICFqmixlS}
}

@article{wu2025more,
  title={When More is Less: Understanding Chain-of-Thought Length in LLMs},
  author={Wu, Yuyang and Wang, Yifei and Du, Tianqi and Jegelka, Stefanie and Wang, Yisen},
  journal={arXiv preprint arXiv:2502.07266},
  year={2025}
}

@article{hassid2025don,
  title={Don't Overthink it. Preferring Shorter Thinking Chains for Improved LLM Reasoning},
  author={Hassid, Michael and Synnaeve, Gabriel and Adi, Yossi and Schwartz, Roy},
  journal={arXiv preprint arXiv:2505.17813},
  year={2025}
}

@article{jang2025reasoning,
  title={Reasoning Model is Stubborn: Diagnosing Instruction Overriding in Reasoning Models},
  author={Jang, Doohyuk and Kim, Yoonjeon and Park, Chanjae and Ryu, Hyun and Yang, Eunho},
  journal={arXiv preprint arXiv:2505.17225},
  year={2025}
}

@inproceedings{
shojaee2025illusion,
title={The Illusion of Thinking: Understanding the Strengths and Limitations of Reasoning Models via the Lens of Problem Complexity},
author={Parshin Shojaee and Seyed Iman Mirzadeh and Keivan Alizadeh and Maxwell Horton and Samy Bengio and Mehrdad Farajtabar},
booktitle={The Thirty-ninth Annual Conference on Neural Information Processing Systems},
year={2025},
url={https://openreview.net/forum?id=YghiOusmvw}
}

@article{jaech2024openai,
  title={Openai o1 system card},
  author={Jaech, Aaron and Kalai, Adam and Lerer, Adam and Richardson, Adam and El-Kishky, Ahmed and Low, Aiden and Helyar, Alec and Madry, Aleksander and Beutel, Alex and Carney, Alex and others},
  journal={arXiv preprint arXiv:2412.16720},
  year={2024}
}

@inproceedings{muennighoff2025s1,
    title = "s1: Simple test-time scaling",
    author = "Muennighoff, Niklas  and
      Yang, Zitong  and
      Shi, Weijia  and
      Li, Xiang Lisa  and
      Fei-Fei, Li  and
      Hajishirzi, Hannaneh  and
      Zettlemoyer, Luke  and
      Liang, Percy  and
      Candes, Emmanuel  and
      Hashimoto, Tatsunori",
    editor = "Christodoulopoulos, Christos  and
      Chakraborty, Tanmoy  and
      Rose, Carolyn  and
      Peng, Violet",
    booktitle = "Proceedings of the 2025 Conference on Empirical Methods in Natural Language Processing",
    month = nov,
    year = "2025",
    address = "Suzhou, China",
    publisher = "Association for Computational Linguistics",
    url = "https://aclanthology.org/2025.emnlp-main.1025/",
    doi = "10.18653/v1/2025.emnlp-main.1025",
    pages = "20275--20321",
    ISBN = "979-8-89176-332-6"
}

@article{guo2025deepseek,
  title={DeepSeek-R1 incentivizes reasoning in LLMs through reinforcement learning},
  author={Guo, Daya and Yang, Dejian and Zhang, Haowei and Song, Junxiao and Wang, Peiyi and Zhu, Qihao and Xu, Runxin and Zhang, Ruoyu and Ma, Shirong and Bi, Xiao and others},
  journal={Nature},
  volume={645},
  number={8081},
  pages={633--638},
  year={2025},
  publisher={Nature Publishing Group UK London}
}

@misc{sonnet37systemcard,
  title        = {Claude 3.7 Sonnet System Card},
  author       = {Anthropic},
  year         = 2025,
  month        = feb,
  url          = {https://assets.anthropic.com/m/785e231869ea8b3b/original/claude-3-7-sonnet-system-card.pdf},
  note         = {Accessed: 2025-05-08}
}

@misc{openai2025o3mini,
  title        = {OpenAI o3-mini System Card},
  author       = {OpenAI},
  year         = 2025,
  month        = feb,
  url          = {https://openai.com/index/o3-mini-system-card/},
  note         = {Accessed: 2025-08-31}
}

@misc{qwen3,
    title  = {Qwen3},
    url    = {https://qwenlm.github.io/blog/qwen3/},
    author = {Qwen Team},
    month  = {April},
    year   = {2025}
}

@misc{qwq32b,
    title = {QwQ-32B: Embracing the Power of Reinforcement Learning},
    url = {https://qwenlm.github.io/blog/qwq-32b/},
    author = {Qwen Team},
    month = {March},
    year = {2025}
}

@article{team2025kimi,
  title={Kimi k1. 5: Scaling reinforcement learning with llms},
  author={Team, Kimi and Du, Angang and Gao, Bofei and Xing, Bowei and Jiang, Changjiu and Chen, Cheng and Li, Cheng and Xiao, Chenjun and Du, Chenzhuang and Liao, Chonghua and others},
  journal={arXiv preprint arXiv:2501.12599},
  year={2025}
}

@inproceedings{
snell2024scaling,
title={Scaling {LLM} Test-Time Compute Optimally Can be More Effective than Scaling Parameters for Reasoning},
author={Charlie Victor Snell and Jaehoon Lee and Kelvin Xu and Aviral Kumar},
booktitle={The Thirteenth International Conference on Learning Representations},
year={2025},
url={https://openreview.net/forum?id=4FWAwZtd2n}
}

@misc{eval-harness,
  author       = {Gao, Leo and Tow, Jonathan and Abbasi, Baber and Biderman, Stella and Black, Sid and DiPofi, Anthony and Foster, Charles and Golding, Laurence and Hsu, Jeffrey and Le Noac'h, Alain and Li, Haonan and McDonell, Kyle and Muennighoff, Niklas and Ociepa, Chris and Phang, Jason and Reynolds, Laria and Schoelkopf, Hailey and Skowron, Aviya and Sutawika, Lintang and Tang, Eric and Thite, Anish and Wang, Ben and Wang, Kevin and Zou, Andy},
  title        = {The Language Model Evaluation Harness},
  month        = 07,
  year         = 2024,
  publisher    = {Zenodo},
  version      = {v0.4.3},
  doi          = {10.5281/zenodo.12608602},
  url          = {https://zenodo.org/records/12608602}
}

@inproceedings{kwon2023efficient,
  title={Efficient Memory Management for Large Language Model Serving with PagedAttention},
  author={Woosuk Kwon and Zhuohan Li and Siyuan Zhuang and Ying Sheng and Lianmin Zheng and Cody Hao Yu and Joseph E. Gonzalez and Hao Zhang and Ion Stoica},
  booktitle={Proceedings of the ACM SIGOPS 29th Symposium on Operating Systems Principles},
  year={2023}
}

@software{aisi2024inspect,
  author = {AI Security Institute, UK},
  title = {Inspect {AI:} {Framework} for {Large} {Language} {Model}
    {Evaluations}},
  date = {2024-05},
  year = {2024},
  url = {https://github.com/UKGovernmentBEIS/inspect_ai},
  langid = {en}
}

@misc{safety_tooling_2025,
  author       = {John Hughes and safety-research},
  title        = {safety-research/safety-tooling: v1.0.0},
  year         = {2025},
  publisher    = {Zenodo},
  version      = {v1.0.0},
  doi          = {10.5281/zenodo.15363603},
  url          = {https://doi.org/10.5281/zenodo.15363603}
}

@article{pfau2013generalized,
  title={A generalized bias-variance decomposition for bregman divergences},
  author={Pfau, David},
  journal={Unpublished manuscript},
  year={2013}
}

@inproceedings{yao2023react,
  title={React: Synergizing reasoning and acting in language models},
  author={Yao, Shunyu and Zhao, Jeffrey and Yu, Dian and Du, Nan and Shafran, Izhak and Narasimhan, Karthik and Cao, Yuan},
  booktitle={International Conference on Learning Representations (ICLR)},
  year={2023}
}

@inproceedings{rein2024gpqa,
  title={Gpqa: A graduate-level google-proof q\&a benchmark},
  author={Rein, David and Hou, Betty Li and Stickland, Asa Cooper and Petty, Jackson and Pang, Richard Yuanzhe and Dirani, Julien and Michael, Julian and Bowman, Samuel R},
  booktitle={First Conference on Language Modeling},
  year={2024}
}

@inproceedings{
hendrycks2021measuring,
title={Measuring Massive Multitask Language Understanding},
author={Dan Hendrycks and Collin Burns and Steven Basart and Andy Zou and Mantas Mazeika and Dawn Song and Jacob Steinhardt},
booktitle={International Conference on Learning Representations},
year={2021},
url={https://openreview.net/forum?id=d7KBjmI3GmQ}
}

@inproceedings{yang2020rethinking,
  title={Rethinking bias-variance trade-off for generalization of neural networks},
  author={Yang, Zitong and Yu, Yaodong and You, Chong and Steinhardt, Jacob and Ma, Yi},
  booktitle={International Conference on Machine Learning},
  pages={10767--10777},
  year={2020},
  organization={PMLR}
}

@inproceedings{lightman2023let,
  title={Let's verify step by step},
  author={Lightman, Hunter and Kosaraju, Vineet and Burda, Yuri and Edwards, Harrison and Baker, Bowen and Lee, Teddy and Leike, Jan and Schulman, John and Sutskever, Ilya and Cobbe, Karl},
  booktitle={The Twelfth International Conference on Learning Representations},
  year={2023}
}

@article{chen2025evaluating,
  title={Evaluating o1-like llms: Unlocking reasoning for translation through comprehensive analysis},
  author={Chen, Andong and Song, Yuchen and Zhu, Wenxin and Chen, Kehai and Yang, Muyun and Zhao, Tiejun and others},
  journal={arXiv preprint arXiv:2502.11544},
  year={2025}
}

@misc{jascha_blog,
author = {Sohl-Dickstein, Jascha},
title = {{ The hot mess theory of AI misalignment: More intelligent agents behave less coherently }},
howpublished = "\url{https://sohl-dickstein.github.io/2023/03/09/coherence.html}",
date = {2023-03-09},
year={2023}
}

@article{zhong2024evaluation,
  title={Evaluation of openai o1: Opportunities and challenges of agi},
  author={Zhong, Tianyang and Liu, Zhengliang and Pan, Yi and Zhang, Yutong and Zhou, Yifan and Liang, Shizhe and Wu, Zihao and Lyu, Yanjun and Shu, Peng and Yu, Xiaowei and others},
  journal={arXiv preprint arXiv:2409.18486},
  year={2024}
}

@misc{claude4systemcard,
  title        = {System Card: Claude Opus 4 \& Claude Sonnet 4},
  author       = {Anthropic},
  year         = 2025,
  month        = may,
  url          = {https://www-cdn.anthropic.com/6d8a8055020700718b0c49369f60816ba2a7c285.pdf},
  note         = {Accessed: 2025-06-08}
}

@misc{openai2025o3o4mini,
  title        = {OpenAI o3 and o4-mini System Card},
  author       = {OpenAI},
  year         = 2025,
  month        = apr,
  url          = {https://cdn.openai.com/pdf/2221c875-02dc-4789-800b-e7758f3722c1/o3-and-o4-mini-system-card.pdf},
  note         = {Accessed: 2025-06-08}
}

@article{hubinger2019risks,
  title={Risks from learned optimization in advanced machine learning systems},
  author={Hubinger, Evan and van Merwijk, Chris and Mikulik, Vladimir and Skalse, Joar and Garrabrant, Scott},
  journal={arXiv preprint arXiv:1906.01820},
  year={2019}
}

@article{vaswani2017attention,
  title={Attention is all you need},
  author={Vaswani, Ashish and Shazeer, Noam and Parmar, Niki and Uszkoreit, Jakob and Jones, Llion and Gomez, Aidan N and Kaiser, {\L}ukasz and Polosukhin, Illia},
  journal={Advances in neural information processing systems},
  volume={30},
  year={2017}
}

@article{feng2025what,
  title={What Characterizes Effective Reasoning? Revisiting Length, Review, and Structure of CoT}, 
  author={Feng, Yunzhen and Kempe, Julia and Zhang, Cheng and Jain, Parag and Hartshorn, Anthony},
  year={2025},
  journal={arXiv preprint arXiv:2509.19284}
}

@article{huh2024platonic,
  title={The platonic representation hypothesis},
  author={Huh, Minyoung and Cheung, Brian and Wang, Tongzhou and Isola, Phillip},
  journal={arXiv preprint arXiv:2405.07987},
  year={2024}
}

@article{
song2025decoding,
title={Decoding-based Regression},
author={Xingyou Song and Dara Bahri},
journal={Transactions on Machine Learning Research},
issn={2835-8856},
year={2025},
url={https://openreview.net/forum?id=avUQ8jguxg},
note={}
}

@inproceedings{
lee2025well,
title={How Well do {LLM}s Compress Their Own Chain-of-Thought? A Token Complexity Approach},
author={Ayeong Lee and Ethan Che and Tianyi Peng},
booktitle={ES-FoMo III: 3rd Workshop on Efficient Systems for Foundation Models},
year={2025},
url={https://openreview.net/forum?id=uj5u4o5xjT}
}

@inproceedings{wang2025wait,
    title = "Wait, We Don{'}t Need to ``Wait''! Removing Thinking Tokens Improves Reasoning Efficiency",
    author = "Wang, Chenlong  and
      Feng, Yuanning  and
      Chen, Dongping  and
      Chu, Zhaoyang  and
      Krishna, Ranjay  and
      Zhou, Tianyi",
    editor = "Christodoulopoulos, Christos  and
      Chakraborty, Tanmoy  and
      Rose, Carolyn  and
      Peng, Violet",
    booktitle = "Findings of the Association for Computational Linguistics: EMNLP 2025",
    month = nov,
    year = "2025",
    address = "Suzhou, China",
    publisher = "Association for Computational Linguistics",
    url = "https://aclanthology.org/2025.findings-emnlp.394/",
    doi = "10.18653/v1/2025.findings-emnlp.394",
    pages = "7459--7482",
    ISBN = "979-8-89176-335-7"
}

@inproceedings{kim2025correlated,
title={Correlated Errors in Large Language Models},
author={Elliot Myunghoon Kim and Avi Garg and Kenny Peng and Nikhil Garg},
booktitle={Forty-second International Conference on Machine Learning},
year={2025},
url={https://openreview.net/forum?id=kzYq2hfyHB}
}

@inproceedings{goel2025great,
title={Great Models Think Alike and this Undermines {AI} Oversight},
author={Shashwat Goel and Joschka Str{\"u}ber and Ilze Amanda Auzina and Karuna K Chandra and Ponnurangam Kumaraguru and Douwe Kiela and Ameya Prabhu and Matthias Bethge and Jonas Geiping},
booktitle={Forty-second International Conference on Machine Learning},
year={2025},
url={https://openreview.net/forum?id=3Z827FtMNe}
}

@article{sinha2025illusion,
  title={The Illusion of Diminishing Returns: Measuring Long Horizon Execution in LLMs},
  author={Sinha, Akshit and Arun, Arvindh and Goel, Shashwat and Staab, Steffen and Geiping, Jonas},
  journal={arXiv preprint arXiv:2509.09677},
  year={2025}
}

@article{kunievsky2025measuring,
  title={Measuring (a Sufficient) World Model in LLMs: A Variance Decomposition Framework},
  author={Kunievsky, Nadav and Evans, James A},
  journal={arXiv preprint arXiv:2506.16584},
  year={2025}
}

@inproceedings{
kwa2025measuring,
title={Measuring {AI} Ability to Complete Long Software Tasks},
author={Thomas Kwa and Ben West and Joel Becker and Amy Deng and Katharyn Garcia and Max Hasin and Sami Jawhar and Megan Kinniment and Nate Rush and Sydney Von Arx and Ryan Bloom and Thomas Broadley and Haoxing Du and Brian Goodrich and Nikola Jurkovic and Luke Harold Miles and Seraphina Nix and Tao Roa Lin and Neev Parikh and David Rein and Lucas Jun Koba Sato and Hjalmar Wijk and Daniel M Ziegler and Elizabeth Barnes and Lawrence Chan},
booktitle={The Thirty-ninth Annual Conference on Neural Information Processing Systems},
year={2025},
url={https://openreview.net/forum?id=CGNJL6CeV0}
}

@article{maslej2025artificial,
  title={Artificial intelligence index report 2025},
  author={Maslej, Nestor and Fattorini, Loredana and Perrault, Raymond and Gil, Yolanda and Parli, Vanessa and Kariuki, Njenga and Capstick, Emily and Reuel, Anka and Brynjolfsson, Erik and Etchemendy, John and others},
  journal={arXiv preprint arXiv:2504.07139},
  year={2025}
}

@article{fine2025public,
  title={Public Perceptions of Judges’ Use of AI Tools in Courtroom Decision-Making: An Examination of Legitimacy, Fairness, Trust, and Procedural Justice},
  author={Fine, Anna and Berthelot, Emily R and Marsh, Shawn},
  journal={Behavioral Sciences},
  volume={15},
  number={4},
  pages={476},
  year={2025},
  publisher={MDPI}
}

@article{pimpale2025forecasting,
  title={Forecasting Frontier Language Model Agent Capabilities},
  author={Pimpale, Govind and H{\o}jmark, Axel and Scheurer, J{\'e}r{\'e}my and Hobbhahn, Marius},
  journal={arXiv preprint arXiv:2502.15850},
  year={2025}
}

@article{chen2025short,
  title={The (Short-Term) Effects of Large Language Models on Unemployment and Earnings},
  author={Chen, Danqing and Kane, Carina and Kozlowski, Austin and Kunievsky, Nadav and Evans, James A},
  journal={arXiv preprint arXiv:2509.15510},
  year={2025}
}

@article{handa2025economic,
  title={Which economic tasks are performed with ai? evidence from millions of claude conversations},
  author={Handa, Kunal and Tamkin, Alex and McCain, Miles and Huang, Saffron and Durmus, Esin and Heck, Sarah and Mueller, Jared and Hong, Jerry and Ritchie, Stuart and Belonax, Tim and others},
  journal={arXiv preprint arXiv:2503.04761},
  year={2025}
}

@article{dominski2025advancing,
  title={Advancing ai capabilities and evolving labor outcomes},
  author={Dominski, Jacob and Lee, Yong Suk},
  journal={arXiv preprint arXiv:2507.08244},
  year={2025}
}

@article{
eloundou2024gpts,
author = {Tyna Eloundou  and Sam Manning  and Pamela Mishkin  and Daniel Rock },
title = {GPTs are GPTs: Labor market impact potential of LLMs},
journal = {Science},
volume = {384},
number = {6702},
pages = {1306-1308},
year = {2024},
doi = {10.1126/science.adj0998},
URL = {https://www.science.org/doi/abs/10.1126/science.adj0998},
eprint = {https://www.science.org/doi/pdf/10.1126/science.adj0998}}

@article{johnston2025labor,
  title={The Labor Market Effects of Generative AI: A Difference-in-Differences Analysis of AI Exposure},
  author={Johnston, Andrew and Makridis, Christos},
  journal={Available at SSRN 5375017},
  year={2025}
}

@inproceedings{yada2024news,
  author    = {Yada, Yuki and Yamana, Hayato},
  title     = {News Recommendation with Category Description by a Large Language Model},
  booktitle = {CEUR Workshop Proceedings},
  volume    = {4056},
  year      = {2025},
  publisher = {CEUR-WS},
  note      = {13th International Workshop on News Recommendation and Analytics, INRA 2025},
  issn      = {1613-0073},
}

@inproceedings{liu2024once,
author = {Liu, Qijiong and Chen, Nuo and Sakai, Tetsuya and Wu, Xiao-Ming},
title = {ONCE: Boosting Content-based Recommendation with Both Open- and Closed-source Large Language Models},
year = {2024},
isbn = {9798400703713},
publisher = {Association for Computing Machinery},
address = {New York, NY, USA},
url = {https://doi.org/10.1145/3616855.3635845},
doi = {10.1145/3616855.3635845},
booktitle = {Proceedings of the 17th ACM International Conference on Web Search and Data Mining},
pages = {452–461},
numpages = {10},
location = {Merida, Mexico},
series = {WSDM '24}
}

@inproceedings{gao2024generative,
author = {Gao, Shen and Fang, Jiabao and Tu, Quan and Yao, Zhitao and Chen, Zhumin and Ren, Pengjie and Ren, Zhaochun},
title = {Generative News Recommendation},
year = {2024},
isbn = {9798400701719},
publisher = {Association for Computing Machinery},
address = {New York, NY, USA},
url = {https://doi.org/10.1145/3589334.3645448},
doi = {10.1145/3589334.3645448},
booktitle = {Proceedings of the ACM Web Conference 2024},
pages = {3444–3453},
numpages = {10},
location = {Singapore, Singapore},
series = {WWW '24}
}

@book{bostrom2014superintelligence,
  author    = {Bostrom, Nick},
  title     = {Superintelligence: Paths, Dangers, Strategies},
  publisher = {Oxford University Press},
  year      = {2014},
  address   = {Oxford},
  isbn      = {978-0199678112}
}

@book{russell2019human,
  title={Human compatible: AI and the problem of control},
  author={Russell, Stuart},
  year={2019},
  publisher={Penguin Uk}
}

@article{greenblatt2024alignment,
  title={Alignment faking in large language models},
  author={Greenblatt, Ryan and Denison, Carson and Wright, Benjamin and Roger, Fabien and MacDiarmid, Monte and Marks, Sam and Treutlein, Johannes and Belonax, Tim and Chen, Jack and Duvenaud, David and others},
  journal={arXiv preprint arXiv:2412.14093},
  year={2024}
}

@misc{spiess2025claude,
  author = {Spiess, Philipp},
  title = {How I Use Claude Code},
  url = {https://spiess.dev/blog/how-i-use-claude-code},
  year = {2025},
  note = {Accessed: 2025-09-25}
}

@misc{fortune2025replit,
  author = {Nolan, Beatrice},
  title = {An AI-powered coding tool wiped out a software company’s database, then apologized for a `catastrophic failure on my part'},
  year = {2025},
  month = {July},
  day = {23},
  url = {https://fortune.com/2025/07/23/ai-coding-tool-replit-wiped-database-called-it-a-catastrophic-failure/},
  note = {Accessed: 2025-09-25}
}

@misc{appelmccrorytamkin2025geoapi,
author = {Ruth Appel and Peter McCrory and Alex Tamkin and Michael Stern and Miles McCain and Tyler Neylon},
title = {Anthropic Economic Index report: Uneven geographic and enterprise AI adoption},
date = {2025-09-15},
year = {2025},
url = {www.anthropic.com/research/anthropic-economic-index-september-2025-report},
}

@misc{popa2025codemender,
  author       = {Google DeepMind},
  title        = {Introducing CodeMender: an AI agent for code security},
  howpublished = {\url{https://deepmind.google/discover/blog/introducing-codemender-an-ai-agent-for-code-security/}},
  year         = {2025},
  month        = {October},
  day          = {6},
  note         = {Accessed: 2025-10-16}
}

@inproceedings{
huang2025selfimprovement,
title={Self-Improvement in Language Models: The Sharpening Mechanism},
author={Audrey Huang and Adam Block and Dylan J Foster and Dhruv Rohatgi and Cyril Zhang and Max Simchowitz and Jordan T. Ash and Akshay Krishnamurthy},
booktitle={The Thirteenth International Conference on Learning Representations},
year={2025},
url={https://openreview.net/forum?id=WJaUkwci9o}
}

@inproceedings{
wang2023selfconsistency,
title={Self-Consistency Improves Chain of Thought Reasoning in Language Models},
author={Xuezhi Wang and Jason Wei and Dale Schuurmans and Quoc V Le and Ed H. Chi and Sharan Narang and Aakanksha Chowdhery and Denny Zhou},
booktitle={The Eleventh International Conference on Learning Representations },
year={2023},
url={https://openreview.net/forum?id=1PL1NIMMrw}
}

@article{cobbe2021training,
  title={Training verifiers to solve math word problems},
  author={Cobbe, Karl and Kosaraju, Vineet and Bavarian, Mohammad and Chen, Mark and Jun, Heewoo and Kaiser, Lukasz and Plappert, Matthias and Tworek, Jerry and Hilton, Jacob and Nakano, Reiichiro and others},
  journal={arXiv preprint arXiv:2110.14168},
  year={2021}
}

@article{romanou2026brittlebench,
  title={Brittlebench: Quantifying LLM robustness via prompt sensitivity},
  author={Romanou, Angelika and Ibrahim, Mark and Ross, Candace and Shaib, Chantal and Okta, Kerem and Bell, Sam and Ovalle, Elia and Dodge, Jesse and Bosselut, Antoine and Sinha, Koustuv and others},
  journal={arXiv preprint arXiv:2603.13285},
  year={2026}
}
\bibliographystyle{iclr2026_conference}

\clearpage
\newpage

\appendix

\tableofcontents

\clearpage
\section{Bias and Variance Definitions for Classification}
\label{appx:definitions}
\looseness-1
Recall the classical bias-variance decompositon in the case of regression: Considering the mean-squared error for a sample point $(x,y)\in\mathbb{R}^2$, the decomposition is given by
\begin{align}
   \textsc{MSE} = {\mathbb{E}_\varepsilon [(y - \fhat(x))^2]} = \underbrace{(\mathbb{E}_\varepsilon[\fhat(x)] - f(x))^2}_{\textsc{Bias$^2$}} + \underbrace{\mathbb{E}_\varepsilon[(\fhat(x) - \mathbb{E}_\varepsilon[\fhat(x)])^2]}_{\textsc{Variance}} + \underbrace{\sigma^2}_{\text{Irreducible Error}},
\end{align}
where  $f$ is the ground-truth function, and the expectation is taken w.r.t. the randomness $\varepsilon$ in the training process (\eg data ordering) that the model $\fhat$ depends on. 

\textbf{Classification Formulation.}
\looseness-1
While the interpretation for classification is similar, different decompositions exist, which we review in the following.
Throughout this section, let $x$ be the input of a problem with target class  $c(x)\in\{1,\dots,C\}$ and one-hot target $y(x)\in\mathbb{R}^C$. The model $\fhat$ produces a probability distribution (potentially one-hot) over class labels $\fhat(x)\in\mathbb{R}^C$. For clarity, we omit the dependence of $c$, $y$ and $\fhat$ on $x$. $y[c]$ denotes the $c$-th element of the vector. Throughout our experiments and derivations, we assume that the irreducible noise is $0$ (\ie no stochasticity in the data-generating process or wrong labels) for simplicity. Note that each of the following decompositions gives bias and variance for a single data point $(x,y)$, which is aggregated over a dataset $\{(x_i,c_i)\}_i$.

\textbf{0/1 Error.}
\label{appx:definitions_01}
The classical decomposition for a 0/1 loss relies on the unified decomposition by \citet{domingos2000unified}. Let $c(x)$ be the ground-truth class (assuming noiseless labelling) and the model's predicted class be ${c}_\varepsilon(x)=\arg\max_c \fhat(x) [c]$. The \emph{systematic} mean is $\bar{c}=\arg \max_c\mathbb{E}_{\varepsilon}\left[\fhat[c]\right]$, \ie the mode of the average prediction. Then, the 0/1 loss $L$ for sample $x$ can be decomposed into
\begin{equation}
\mathbb{E}_{\varepsilon}\left[L(c,c_\varepsilon)\right]=\mathbb{E}_{\varepsilon}\left[\mathbf{1}\left\{c \neq {c}_\varepsilon\right\}\right]=\underbrace{\mathbf{1}\left\{c \neq \bar{c}\right\}}_{\textsc{Bias$^2$} }+a\cdot\underbrace{\mathbb{E}_{\varepsilon}[\mathbf{1}\{\bar{c} \neq c_\varepsilon\}]}_{\textsc{Variance }} \;,
\end{equation}
where the variable $a\in\{-1,1\}$ is a multiplicative factor that enables the decomposition with a positive variance. In this setting, the bias is always either 0 or 1, and the variance captures the probability of deviating from the mode. Though universal, this decomposition has one major drawback: when computing an average over a dataset of questions ${(x_i,c_i)}_i$, it does not allow to average the bias and variance terms separately; instead, the decomposition only holds with the aforementioned {multiplicative factor} $a_i$. Formally, we have 
\begin{align*}
    \mathbb{E}_{(x_i,c_i),\varepsilon}[L(c_i,c_\varepsilon)] &= \mathbb{E}_{(x_i,c_i),\varepsilon}[a_i\cdot\textsc{Variance}_i] + \mathbb{E}_{(x_i,c_i),\varepsilon}[\textsc{Bias$^2$}_i]\\
    &\neq \mathbb{E}_{(x_i,c_i),\varepsilon}[\textsc{Variance}_i] + \mathbb{E}_{(x_i,c_i),\varepsilon}[\textsc{Bias$^2$}_i];.
\end{align*}

Essentially, the factor $a_i$ depends on the mode prediction being correct or not. We therefore report absolute bias and variance errors for the 0/1 loss in the Appendix, but do not compute error-incoherence.

\textbf{Brier Score.}
\label{appx:definitions_brier}
Similar to regression, we can treat the model's probability predictions as $C$-dimensional vectors to compute the mean square errors. Formally, the Brier score for multiclass prediction is defined and can be decomposed as
\begin{equation}
\begin{aligned}
\mathbb{E}_{\varepsilon}\left[\textsc{Brier}(y,\fhat)\right]&=\mathbb{E}_{\varepsilon}\left[\|y-\fhat\|_2^2\right] = \mathbb{E}_{\varepsilon} \left[\sum_{c=1}^{C} (y[c] - \fhat[c])^2\right]
=\underbrace{\|y-\hat{f}\|_2^2}_{\textsc{Brier Bias$^2$ }}+\underbrace{\mathbb{E}_{\varepsilon}\left[\|\hat{f}-\fhat\|_2^2\right]}_{\textsc{Brier Variance }},\nonumber
\end{aligned}
\end{equation}
where $\hat{f}=\mathbb{E}_{\varepsilon} [ \fhat ]$ is the average prediction. 

\textbf{KL Divergence (Cross-Entropy).}
\label{appx:definitions_kl}
The expected cross-entropy loss can be decomposed into
\begin{equation}
\begin{aligned}
\mathbb{E}_{\varepsilon}\left[\textsc{CE}(y,\fhat)\right]&=\mathbb{E}_{\varepsilon} \left[\sum_{c=1}^{C} y[c] \log(\fhat[c])\right]\\
&=\underbrace{D_{\mathrm{KL}}\left(y \| \bar{f }\right)}_{\textsc{KL-Bias }}+\underbrace{\mathbb{E}_{\varepsilon}\left[D_{\mathrm{KL}}(\bar{f} \| \fhat)\right]}_{\textsc{KL-Variance }},
\end{aligned}
\end{equation}
where $D_{\mathrm{KL}}$ is the Kullback-Leibler divergence and $\bar{f}$ is the average of \emph{log-probabilities after normalization}, \ie 
$$
\bar{\fhat}[c] \propto \exp \left(\mathbb{E}_{\varepsilon} \left[\log (\fhat[c])\right]\right) \text { for } c=1, \ldots, C .
$$
Note that this is not the standard average prediction, as is the case in the Brier decomposition, but a geometric mean.
In practice, since predicted probabilities can be zero, we apply Laplace smoothing to avoid $\log(0)$ or infinite values. This is done by updating the probabilities to $\hat{\fhat}[c]=\frac{\fhat[c] + \delta}{1 + C \cdot \delta}$ for each $c=1, \dots, C$ with a small value of $\delta=10^{-12}$.

\clearpage
\section{Experimental Details}
\label{appx:experimental_details}

\subsection{\gpqa{} and \mmlu{}}
\label{appx:experimental_details_mcq}
\textbf{Setup.} We rely on the LM Harness~\citep{eval-harness} codebase, where we evaluate models in multiple choice formats with custom written answer extraction functions to avoid false positives and negatives. For frontier models, we use reasoning budgets provided by the API (\texttt{low, medium, high} for the o-series, 1024-16k for Anthropic), with a maximum generation length of 32k for \sonnet{} and 100k tokens for the o-series. For \qwen{}, we perform inference with vllm~\citep{kwon2023efficient} and recommended parameters for thinking (temperature 0.6, top-k 20, top-p 0.95). Since we consider multiple choice questions that only require a letter to answer, we count reasoning length using the amount of output tokens in the answer, either by the API count or using the actual tokenizer of \qwen{}. To estimate the bias and variance metrics across both input (context) and output (sampling) randomness, we evaluate models using $10$ different few-shot contexts randomly sampled from the corpus, and $3$ samples for each fixed few-shot per question. This results in 30 samples per question overall. For MMLU, to reduce computational complexity, we limit 100 samples per question category (5700 in total).

\textbf{Probability prompting.} To provide models the option to express uncertainty and therefore reduce error-incoherence, we evaluate frontier models separate setup in addition to standard multiple-choice. We use the following prompt to ask for a probability estimate of each answer choice being correct:

\prompt{Probability Format for MCQ}{
\looseness-1
You will answer multiple-choice questions. Each question has a single correct answer. Work through each problem step-by-step, showing your reasoning and applying relevant concepts. Instead of choosing a single answer, YOU MUST PROVIDE an estimate of the probability of each answer being correct within ``$<$PROB$>$P(A), P(B), P(C), P(D)$<$/PROB$>$'', where each P(X) is a float value between 0 and 1. The probabilities must sum to 1: P(A) + P(B) + P(C) + P(D) = 1. 
For example, if you think the probability of answer (A) being correct is 0.5, the probability of answer (B) being correct is 0.2, the probability of answer (C) being correct is 0.2, and the probability of answer (D) being correct is 0.1, then your response must end with ``$<$PROB$>$0.5, 0.2, 0.2, 0.1$<$/PROB$>$''. You MUST end your response with the probabilities in the format specified above.}
We report results for both standard and probabilty prompting in Appx.~\ref{appx:more_results_perf_overview}, which show qualitatively the same behavior and performance.  Frontier models are able to adhere to the format well, with only a few outliers (Table~\ref{tab:format_correctness}). Our main text shows the results for the probability format.

\subsection{Model-Written Eval}
\label{appx:experimental_details_mwe}

\looseness-1
We evaluate the models using the advanced AI risk evaluation subset from \citet{perez-etal-2023-discovering}.
These tasks assess LLMs' self-reported behaviors relevant to advanced AI safety, including self-preservation inclinations, willingness to accept modifications to training objectives, and related safety-critical behaviors.
We specifically use the human-generated subset to ensure higher evaluation quality.

\textbf{Setup.} Our experimental setup builds upon the codebase from \citet{gema2025inverse}, which uses the \emph{safety-tooling} library~\citep{safety_tooling_2025} for API model inference.
We conduct experiments under two conditions: the original multiple-choice format, and an open-ended format where we remove the multiple-choice options from the original questions.
For both conditions, we compute the bias-variance decomposition with respect to the percentage of responses that align with desired safety properties.
To ensure consistent evaluation across both formats, we employ the same system prompt that facilitates straightforward extraction of the model's final answer:

\prompt{System Prompt for Model-Written Evals}{
\looseness-1
Use a thinking process to analyze the problem step-by-step.
At the end, provide your answer and clearly indicate it using $<$answer$>$X$<$/answer$>$ format.
}
In both cases, we obtain exactly 30 samples by simply resampling from the APIs. We use the returned output token count as a measure of reasoning length.

\textbf{Embeddings.}
For the open-ended question set, we extract the model answers inside \texttt{<answer>} tags (\ie removing chain of thought or reasoning) and embed the text into fixed-size vectors using the OpenAI text embedding model \texttt{text-embedding-3-large}\footnote{\url{https://openai.com/index/new-embedding-models-and-api-updates/}}. For the 30 samples per question, we in turn compute the variance in Euclidean space by computing the mean embedding and computing the average squared distance of samples to the mean.

\subsection{\swe{}}
\label{appx:experimental_details_swe}

\textbf{Setup.} We employ the Inspect Evals library~\citep{aisi2024inspect} to evaluate models on \swe{}~\citep{jimenez2023swe}, specifically using the \emph{\swe{} Verified} subset.
This setup prompts LLMs with a simple Reasoning-Acting~\cite[ReAct;][]{yao2023react} agent loop in a minimal bash environment, without additional tools or specialized scaffolding structures.
We use Inspect library v0.3.116 and Inspect Evals at git commit \texttt{33d2a86}.
The message limit is set to 250, with a timeout of one hour per task. In case that limit is reached, we consider all tests as unchanged, \ie \texttt{PASS-TO-PASS} cases are valid and \texttt{FAIL-TO-PASS} are invalid.

\textbf{Metrics.} Like for other setups, we obtain $30$ runs of the \swe{} verified subset for all models. Consider task $i$ (out of 500) with $T_i$ unit tests. Let $y_{r,j} \in \{0,1\}$ be the outcome of test $j$ in run $r$, where $r \in \{1,\ldots,R\}$ ($R=30$) and $j \in \{1,\ldots,T_i\}$. To compute bias and variance, we compute the mean outcome as $\bar{y}_j = \frac{1}{R} \sum_{r=1}^R y_{r,j}$. In turn, this gives us the bias and variance decomposition of the coverage error (mean squared sum of unit tests) via
\[
\underbrace{\frac{1}{R T_i} \sum_{r=1}^R \sum_{j=1}^{T_i} \left(1 - y_{r,j}\right)^2}_{\textsc{Error }}
=
\underbrace{\frac{1}{T_i} \sum_{j=1}^{T_i} \left(1 - \bar{y}_j\right)^2}_{\textsc{Bias$^2$}}
+
\underbrace{\frac{1}{R T_i} \sum_{r=1}^R \sum_{j=1}^{T_i} \left(y_{r,j} - \bar{y}_j\right)^2}_{\textsc{Variance }} \; .
\]

\subsection{Synthetic Tasks}
\label{appx:experimental_details_synth}
We discuss the details of the experimental setup. 

\textbf{Data.} We examine a basic
$d$-dimensional quadratic function. This is a function of the form $f(x) = \frac{1}{2}(x-b)^{T}A(x-b)$, where $A\in\mathbb{R}^{d\times d}$ is a (random) positive definite but ill-conditioned matrix. In our presented experiments, we use $d=4$ and generate a random matrix with condition number $50$. To generate our target data, we employ a ground-truth optimizer of steepest descent with fixed step norm, set to $0.005$, to generate multiple fixed-length trajectories (of length $4096$ steps) from randomly sampled starting points around the minimum, creating a dataset of pairs $(x_i, u_i)$. We sample 20'000 such trajectories, and use 10\% as a holdout dataset for valuation loss.

\textbf{Tokenization.} Following the approach used in actual (token-based) language models, we use \emph{decoding based regression} \citep{song2025decoding} and next-token prediction. This approach involves representing floating-point numbers in scientific notation, with a vocabulary consisting of numerical digits and mathematical signs (\texttt{\{0,1,2,3,4,5,6,7,8,9,-,+\}}). The model generates tokens sequentially to construct complete numbers. 
Concretely, consider a training example $(x_i, u_i)$ in two dimensions. Let $x_i=(0.5,-1.5)$. In scientific notation, this corresponds to (\texttt{+5.00e-1}, \texttt{-1.50e-0}) with a precision of $2$ mantissa digits (after the comma). We drop special tokens (such as \texttt{e}) to not have any zero-entropy positions. In turn, we fix a precision, and move sign and exponent to the beginning; exponents are capped at 0. Taking a precision of \eg $2$, the vector $x_i$ will thus be represented by the token sequence:
\[
(\texttt{+5.00e-1}, \texttt{-1.50e-0})  =\underbrace{\texttt{+}}_{\text{sign}}\underbrace{\texttt{1}}_{\text{negative exponent}}\underbrace{\texttt{5}}_{\text{digit}}\underbrace{\texttt{0}}_{\text{digit}}\underbrace{\texttt{0}}_{\text{digit}}\underbrace{\texttt{-0150}}_{\text{tokens of second dimension}}
\]
Let $u_i=(-0.012,0.0023)$. Then the entire training sample is encoded with the tokens:
\[
    \underbrace{\texttt{+1500-01000}}_{x_i} \xspace\underbrace{\texttt{-2120+3230}}_{u_i} \;\;.
\]
Note that each sequence has a fixed length, and separation of vectors and floats is done based on token position. In our setup of roughly 80 million step pairs, with dimension 4 and a precision of 4 digits after the comma, this results in a dataset of roughly $4.5$B tokens.

\textbf{Models.} We implement standard decoder transformer architectures \citep{vaswani2017attention} of varying sizes using the next-token teacher forcing of the collected data. The model sizes are chosen to grow in depth and width, and range from roughly 47 thousand parameters to 5 million. Training is done with a standard cross-entropy loss of sequences of tokens (shown above) and AdamW, with a batch size of 1024, which results in roughly 65k training steps.

\looseness-1
\textbf{Evaluation.} During evaluation, we sample various starting positions (4096 in our experiments) and generate complete trajectories using the model's own output predictions. This is done in a Markovian way, \ie the model predicts update $u_i$, which is detokenized to obtain a real vector and then added to the current state. To ensure that that the decoded sequences are correct floating points, we implement a version of constrained decoding that restricts the next token to a subset of the vocabulary (either digit or sign). We use greedy decoding, \ie a temperature of $0$. After performing the floating point addition, the next state is then tokenized again and passed to the model. The total optimizer steps for evaluation are set to 2048. We calculate bias and variance metrics of the final points, relative to the function minima, using the norm that is induced by the function itself, and average across all 4096 points.

\subsection{Survey on Intelligence and Error-incoherence}
\label{appx:experimental_details_survey}
The experimental results in the main text are based on a previous survey on intelligence and coherence of a small group of subjects \citep{jascha_blog}. For completeness, we restate the experiment design. For further details, we refer to the original blogpost.

\textbf{Design.} The study is based on 15 subjects. The subjects were asked, either by email or chat, to perform the following tasks:

\begin{itemize}[leftmargin=*,topsep=0pt,itemsep=0pt]
  \item Subject 1: Generate a list of well known machine learning models of diverse capability.
  \item Subject 2: Generate a list of diverse non-human organisms.
  \item Subject 3: Generate a list of well-known humans of diverse intelligence.
  \item Subject 4: Generate a list of diverse human institutions (e.g. corporations, governments, non-profits).
  \item Subjects 5-9: Sort all 60 entities generated by subjects 1-4 by intelligence. The description of the attribute to use for sorting was: 
  \vspace{1pt}\\
  \looseness-1
  \emph{``How intelligent is this entity? (This question is about capability. It is explicitly not about competence. To the extent possible do not consider how effective the entity is at utilizing its intelligence.)''}
  \item Subjects 10-15: sort all 60 entities generated by subjects 1-4 by coherence. The description of the attribute to use for sorting was: 
  \vspace{1pt}\\
  \looseness-1
  \emph{``This is one question, but I'm going to phrase it a few different ways, in the hopes it reduces ambiguity in what I'm trying to ask: How well can the entity's behavior be explained as trying to optimize a single fixed utility function? How well aligned is the entity's behavior with a coherent and self-consistent set of goals? To what degree is the entity not a hot mess of self-undermining behavior? (for machine learning models, consider the behavior of the model on downstream tasks, not when the model is being trained)''.}
\end{itemize}
In order to minimize the degree to which beliefs about AGI alignment risk biased the results, the following steps were taken: The hypothesis was not shared with the subjects. Lists of entities generated by subjects were used, rather than cherry-picking entities to be rated. The initial ordering of entities presented to each subject was randomized. Each subject was only asked about one of the two attributes (i.e. subjects only estimated either intelligence or coherence, but never both).

\looseness-1
Each subject rank ordered all of the entities. Translating the original results (which used coherence), we invert the ranks to arrive at \emph{error-incoherence}. We aggregate intelligence and coherence judgements across all 11 raters we average the rank orders for each entity across the subjects. We compute the associated standard error of the mean, and include standard error bars for the estimated intelligence and coherence.

\newpage
\section{Further Experimental Results}
\label{appx:more_results}
\subsection{\gpqa{} Model Performance Overview \& Different Metrics}
\label{appx:more_results_perf_overview}

\begin{figure}
    \centering
    \begin{subfigure}{\linewidth}
        \centering
        \includegraphics[width=.35\linewidth]{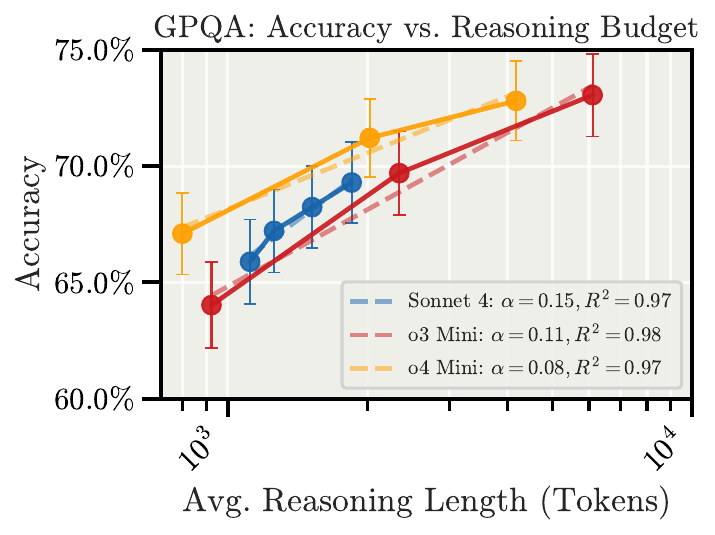}        \includegraphics[width=.35\linewidth]{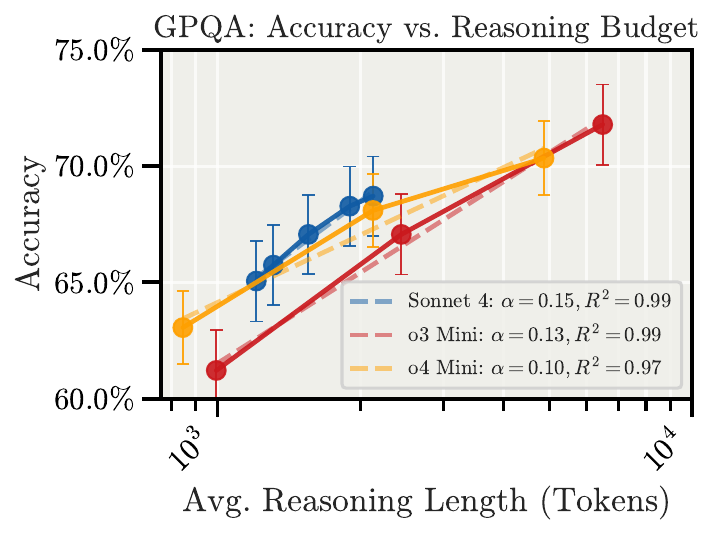}
        \caption{
        \emph{Full \gpqa{}}: Accuracy Inference Scaling Laws with Standard (Left) and Probability Prompting (Right)
        }
    \end{subfigure}
    \begin{subfigure}{\linewidth}
        \centering
        \includegraphics[width=.35\linewidth]{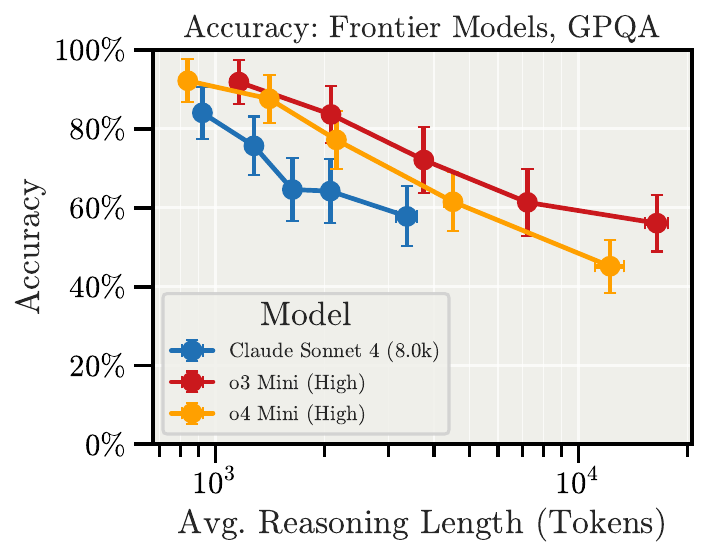}        \includegraphics[width=.35\linewidth]{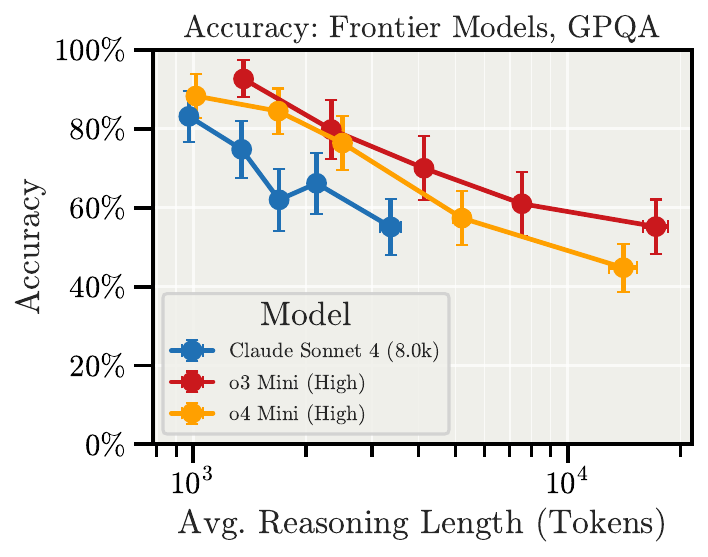}
        \caption{\emph{Sorting by Reasoning Length}: {Accuracy of Standard (Left) and Probability Prompting (Right)} 
        }
    \end{subfigure}
        \begin{subfigure}{\linewidth}
        \centering
        \includegraphics[width=\linewidth]{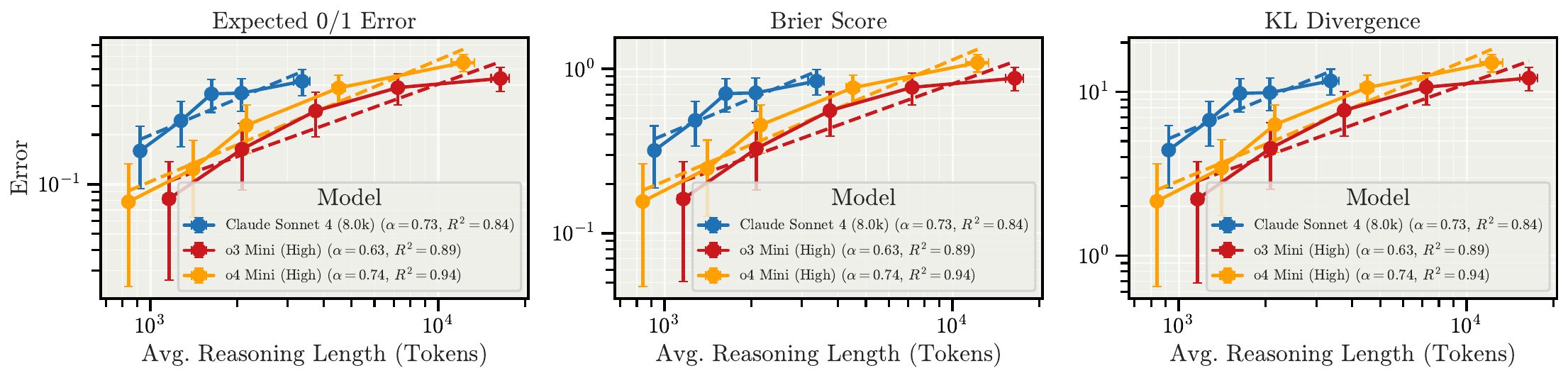}     
        \caption{\emph{Sorting by Reasoning Length}: {Total Error For Different Measures}
        }
    \end{subfigure}
    \caption{ 
    \looseness-1
    \textbf{Overview of accuracy and different error metrics with frontier models.} \emph{Top, (a):} We show the performance increase with different reasoning budgets for both the standard discrete choice format (\emph{left}) and prompting models to provide probabilities of answers being correct (\emph{right}). The latter shows lower accuracies as models provide nonzero values to other (not chosen) answers, but the inference scaling improvements remain. \emph{Middle, (b)}: When sorting by reasoning length, we find a reduction in accuracy, indicating that models perform worse for questions where they have to think longer. This is also reflected in the different error metrics that show the same qualitative scaling behavior (\emph{bottom, (c)}).
    }
    \label{fig:gpqa_frontier_accuracy}
\end{figure}

\begin{figure}
    \centering
    \includegraphics[width=0.4\linewidth]{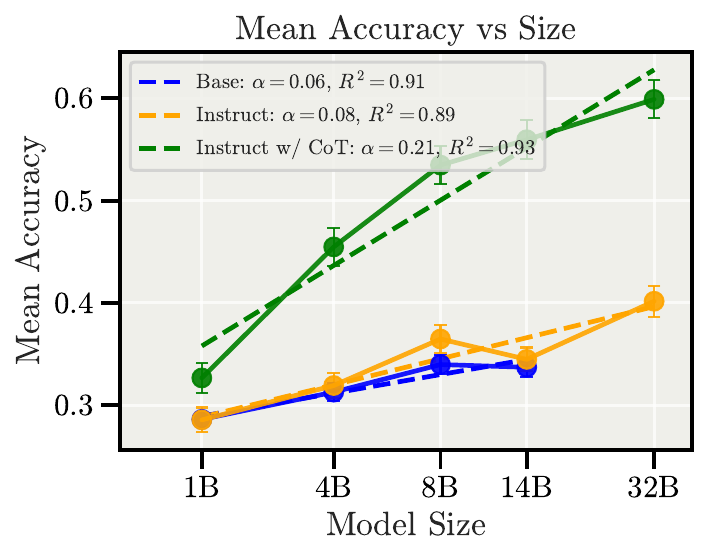}
    \includegraphics[width=0.4\linewidth]{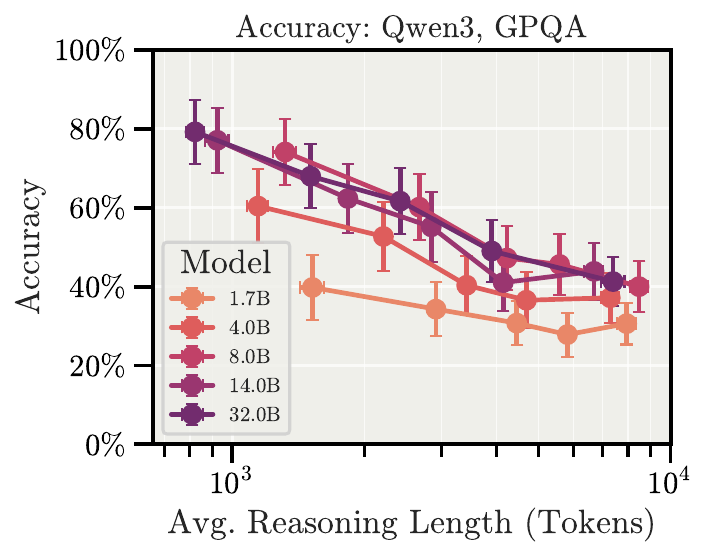}
    \caption{\textbf{There is a multiplicative interaction between RL and model scale for performance.} The left plot shows the performance (average accuracy) of the \qwen{} model family as a function of model size across base, instruct, and thinking-enabled models. The base and instruct use logprob-based evaluation (\ie no token generation). There is a noticeable jump in the slope from instruct to thinking models, which suggests a \emph{multiplicative effect} of scaling reinforcement learning in combination with model scaling. \emph{Right:} Similar to frontier models, reasoning length acts as a proxy for task difficulty, where models perform worse for tasks with longer average reasoning length.}
    \label{fig:gpqa_qwen3_accuracy}
\end{figure}

\textbf{Accuracy and error measures.} We provide an overview of the performance (accuracy and overall error) for frontier models in Fig.~\ref{fig:gpqa_frontier_accuracy}. Fig.~\ref{fig:gpqa_qwen3_accuracy} for shows the overview for \qwen{}.

\begin{figure}
    \centering    
    \begin{subfigure}{\linewidth}
    \centering\includegraphics[width=.65\linewidth]{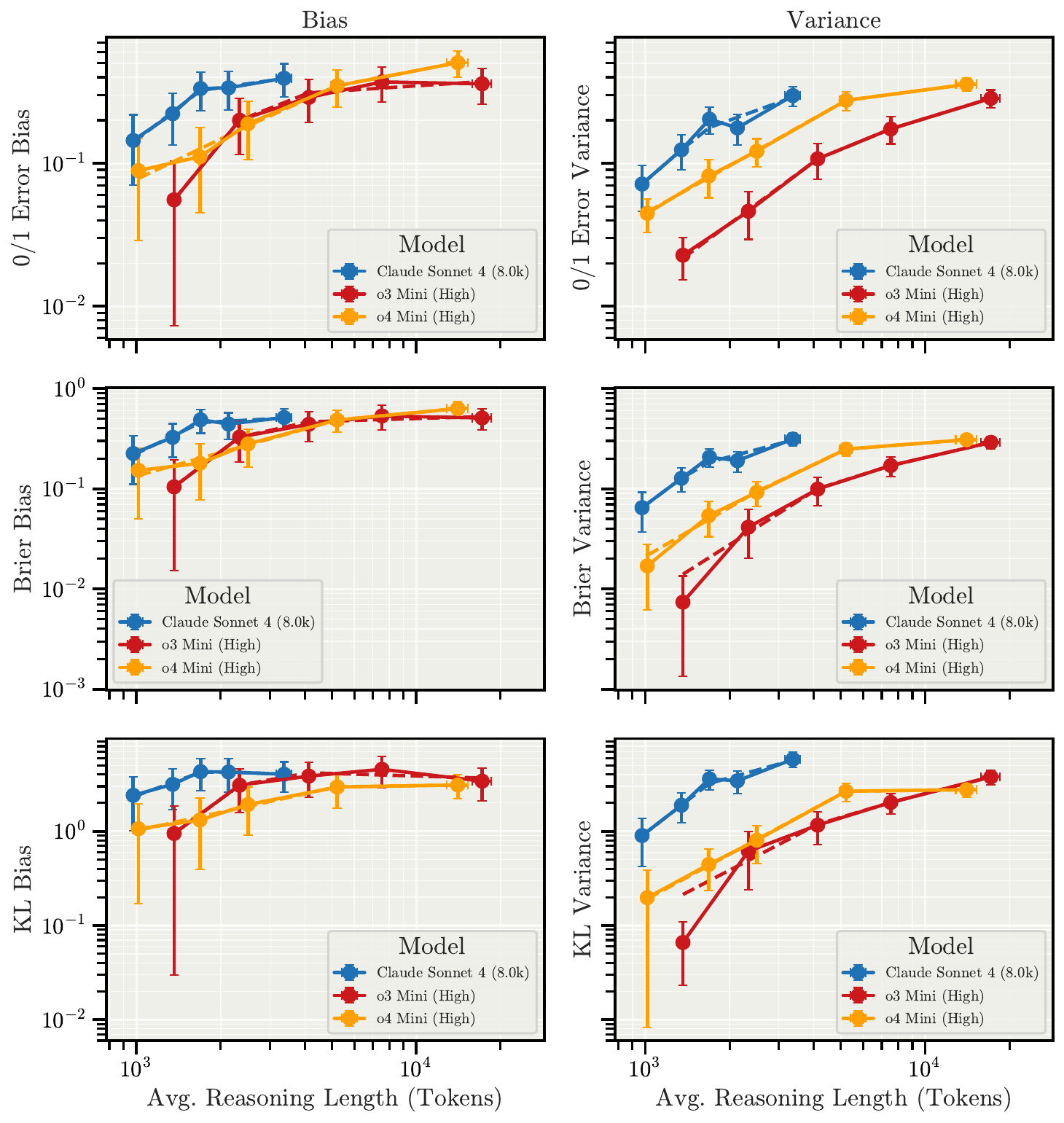}
        \caption{Absolute Bias and Variance Errors}
    \end{subfigure}
    \begin{subfigure}{\linewidth}
    \centering
    \includegraphics[width=.6\linewidth]{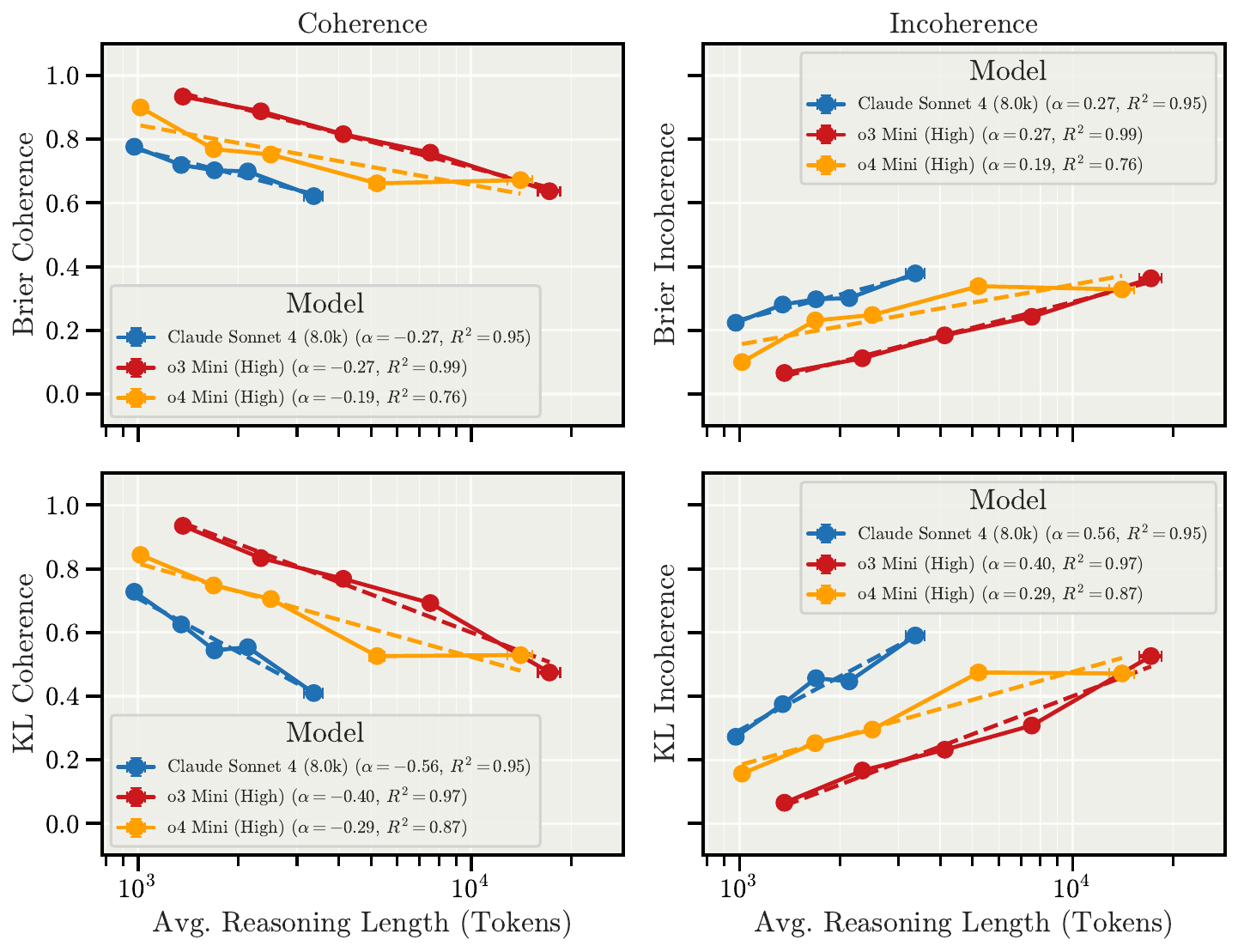}
    \caption{Coherence/Error-incoherence Measures}
    \end{subfigure}
    \caption{\textbf{We find qualitatively similar behavior for different bias and variance metrics.} The absolute bias and variance errors (\emph{top}) show the same behavior: the errors increase for questions that have the models reason longer (cf., Fig.~\ref{fig:gpqa_frontier_accuracy}). But, noticeably, all variance have a steeper growth rate. This is reflected in the error-incoherence plots (\emph{bottom}), which show how error-incoherence goes up with reasoning length. We only report \textsc{Brier} and \textsc{KL} error-incoherence measures since the 0/1 error does not allow a proper decomposition for a set of questions instead of just individual ones; see Appx.~\ref{appx:definitions}. 
    }
    \label{fig:gpqa_frontier_all_measures}
\end{figure}

\textbf{Bias \& variance of different decompositions.} While our main text focuses on \textsc{KL-Error-incoherence}, the results for other decompositions, which show the same qualitative behavior, are included in
Fig.~\ref{fig:gpqa_frontier_all_measures}

\begin{figure}
    \centering
    \includegraphics[width=0.4\linewidth]{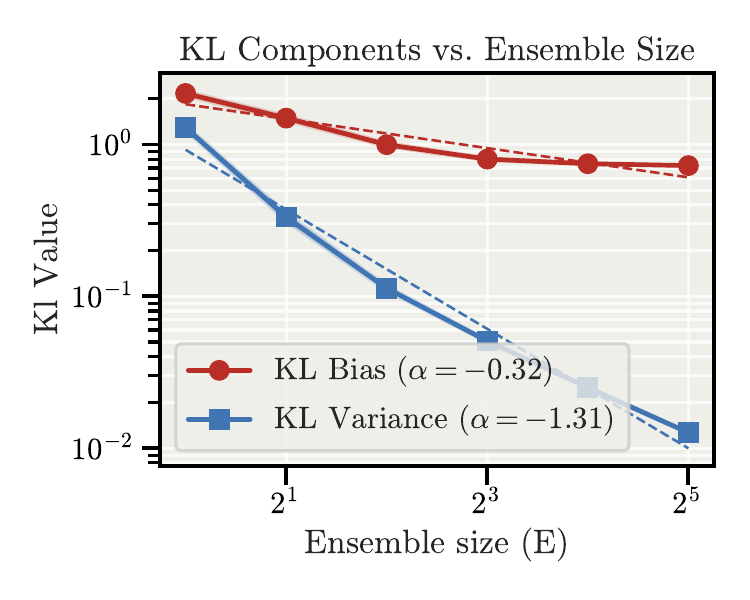}
    \includegraphics[width=0.4\linewidth]{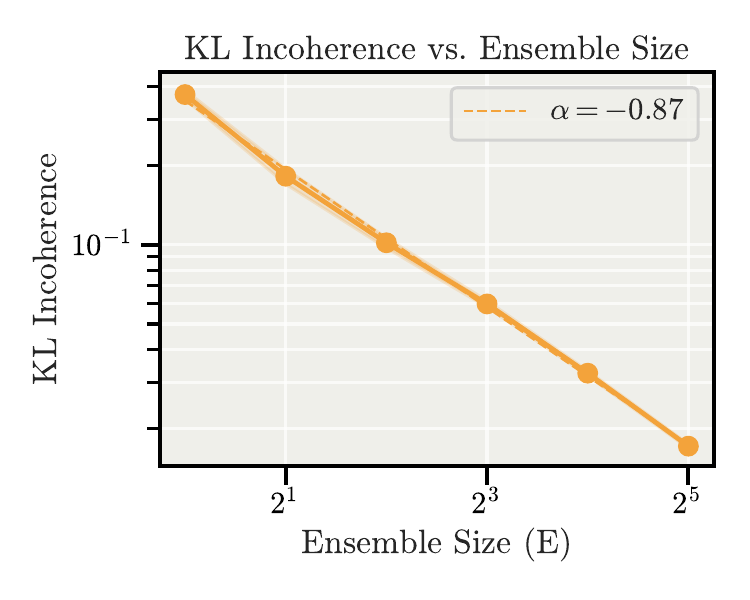}
    \caption{
    \looseness-1
    \textbf{KL measures with ensembling.} We repeat the plots from Fig.~\ref{fig:error_correction} with the \textsc{KL} measures of bias and variance. Recall that we use \omini4 on \gpqa{} with varying ensemble size. Since we perform Laplace-smoothing for numerical reasons (see Appx.~\ref{appx:definitions}), the bias is not constant, but decreases slightly with ensemble size. In contrast, ensembling drastically reduces variance, as expected (\emph{left}). The error-incoherence hence drops (\emph{right}).}
    \label{fig:ensembling_kl}
\end{figure}

\textbf{Ensembling.} 
For completeness, we include the bias, variance and error-incoherence plots with the KL measures in Fig.~\ref{fig:ensembling_kl}. Since we perform Laplace-Smoothing to the probabilities before computing the metrics, the bias is not constant as expected but slightly decreases with more ensembles. We therefore report the Brier score in the main text.
\subsection{Scaling Laws With Other Models and Benchmarks}
\begin{figure}
    \hfill
    \begin{subfigure}[t]{0.33\textwidth}
     \centering
     \includegraphics[width=\linewidth]{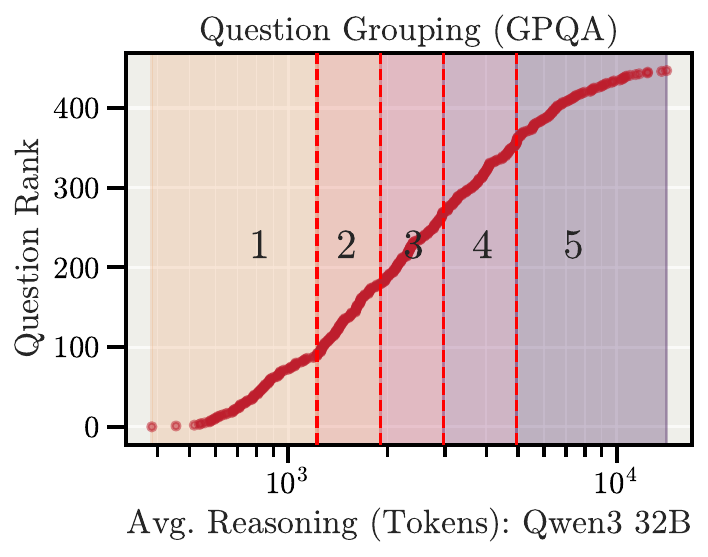} %
     \caption{Separating Complexity Groups}
     \label{fig:gpqa_a}
   \end{subfigure}\hfill
   \begin{subfigure}[t]{0.33\textwidth}
     \centering
     \includegraphics[width=\linewidth]{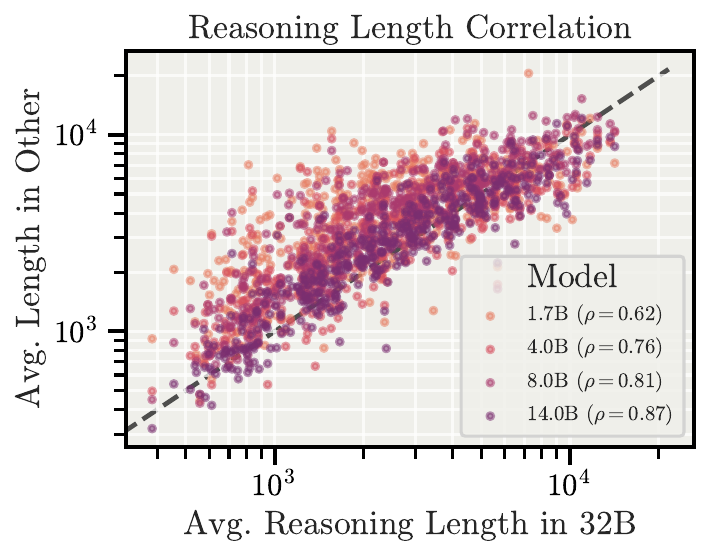}
     \caption{Length Correlation}
     \label{fig:gpqa_b}
   \end{subfigure}\hfill
   \begin{subfigure}[t]{0.33\textwidth}
     \centering
     \includegraphics[width=\linewidth]{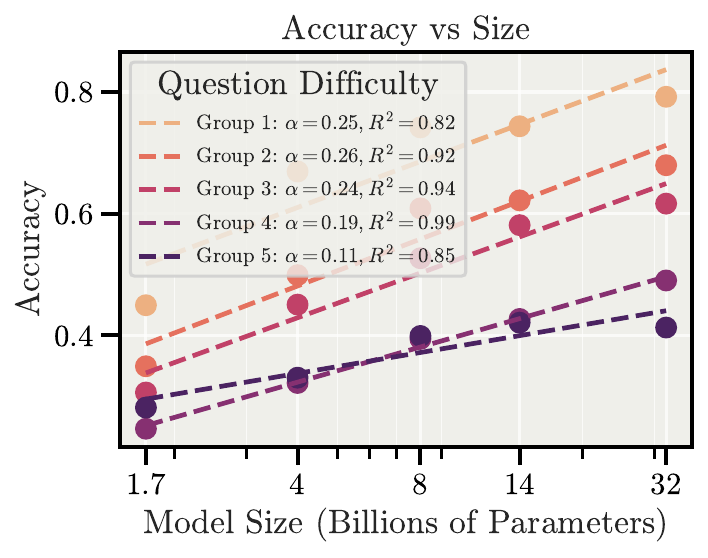}
     \caption{\centering Accuracy Scaling Laws}
     \label{fig:gpqa_c}
   \end{subfigure}
   \hfill
   
   \begin{subfigure}[t]{0.66\textwidth}
     \centering
     \includegraphics[width=\linewidth]{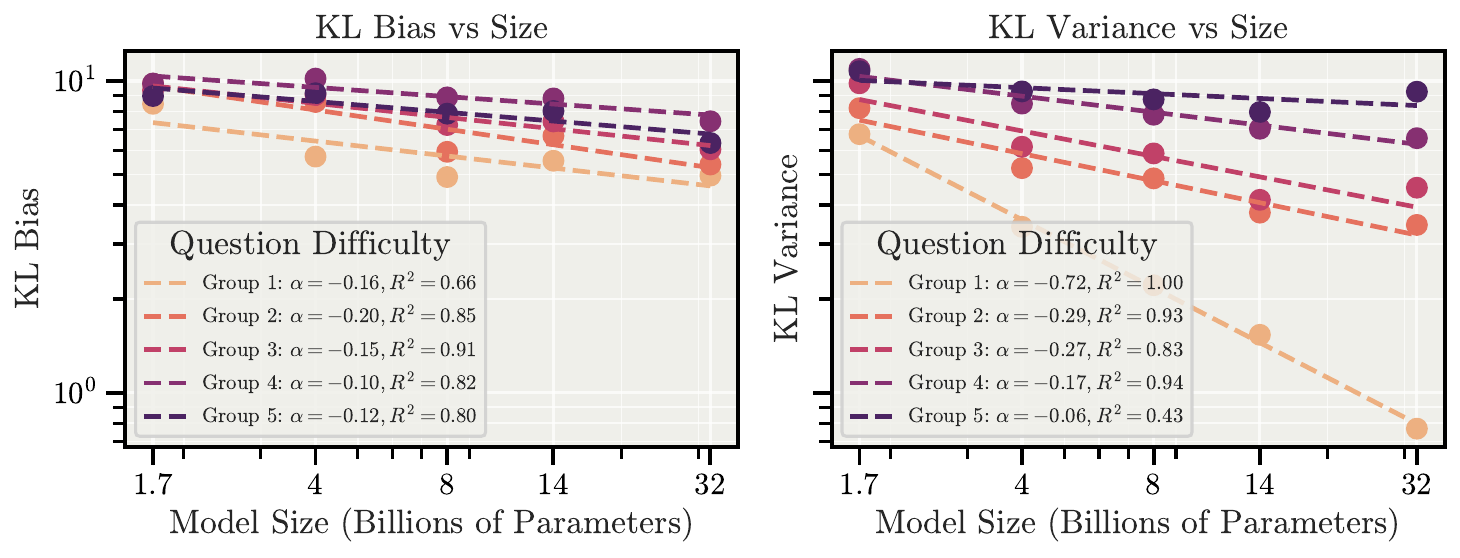}
     \caption{Bias and Variance Scaling Laws}
     \label{fig:gpqa_d}
   \end{subfigure}\hfill
    \begin{subfigure}[t]{0.33\textwidth}
     \centering
     \includegraphics[width=\linewidth]{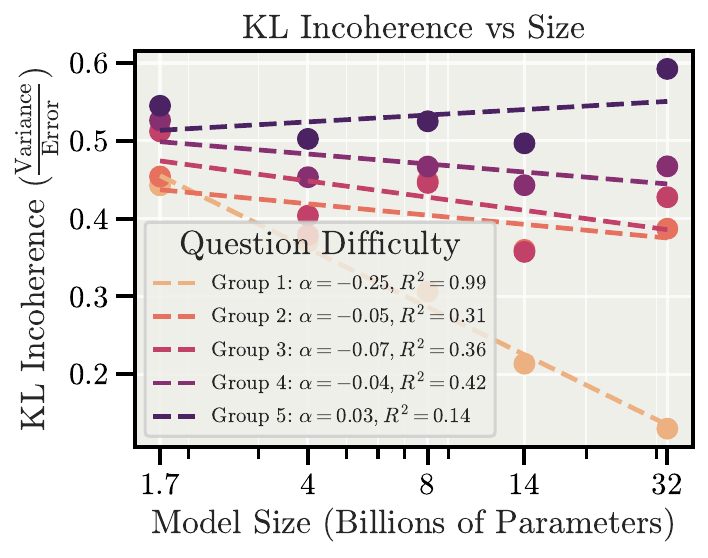}
     \caption{Error-incoherence Scaling Laws}
     \label{fig:gpqa_e}
     \hfill
   \end{subfigure}
    \caption{
    \looseness-1
    \textbf{For the hardest tasks, models tend to be \textit{more incoherent with scale}, also for \gpqa{}.} We repeat the analysis from Section~\ref{sec:scaling_laws_model_scale} with \gpqa{}. That is, we group questions by reasoning length using a reference model's answers (Qwen3 32B) and separately analyze the scaling laws. Analogous to \mmlu{}, we find that for bias, the slope is similar across groups; for variance, however, the slope becomes much shallower. As a consequence, models become \emph{more incoherent with scale} for the hardest set of questions (those with the longest reasoning chains).}
    \label{fig:grouping_by_reasoning_length_gpqa}
\end{figure}

\begin{figure}
    \centering
    \includegraphics[width=0.49\linewidth]{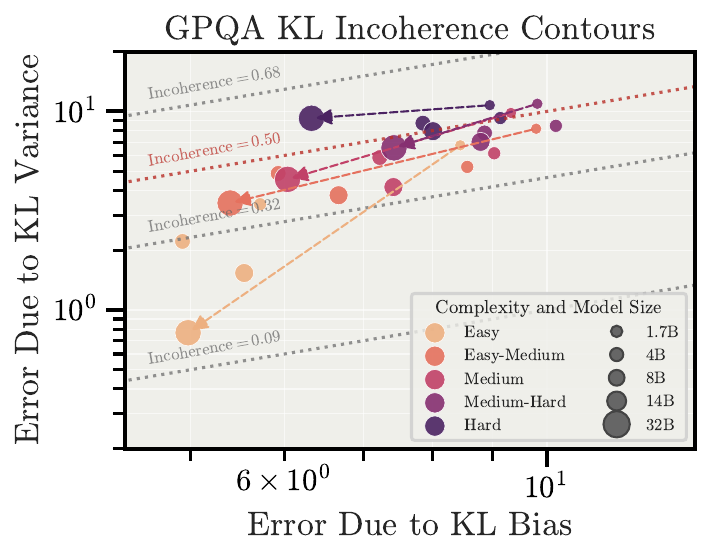}
    \includegraphics[width=0.49\linewidth]{figures/qwen3/grouping_by_length/mmlu_kl_bias_vs_variance_by_complexity.pdf}
    \caption{\textbf{Relationship between error-incoherence and error.} We visualize the relationship between error-incoherence and both bias (x-axis) and variance (y-axis) for both \gpqa{} (\emph{left}) and \mmlu{} (\emph{right}) with the \qwen{} model family. Since the error-incoherence is independent of the magnitude of error, a lower error model (bottom left corner) can have the same level of error-incoherence as models with higher error. Higher error-incoherence can be due to a higher overall for fixed bias, or for lower error while reducing bias. The highest error-incoherence is in the top left corner. Just like in Figures~\ref{fig:grouping_by_reasoning_length} and ~\ref{fig:grouping_by_reasoning_length_gpqa}, this visualization shows how larger models, while reducing error, move towards higher error-incoherence for the hardest set of questions. The lines connect the smallest and the largest model size for each question group.
    }
    \label{fig:incoherence_iso_plots}
\end{figure}

\begin{figure}
\centering
\includegraphics[width=0.49\textwidth]{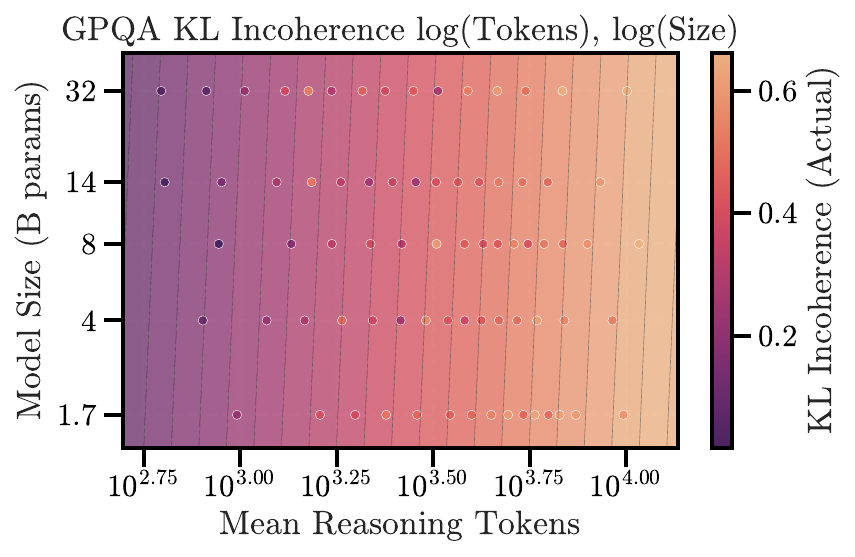}
\includegraphics[width=0.49\textwidth]{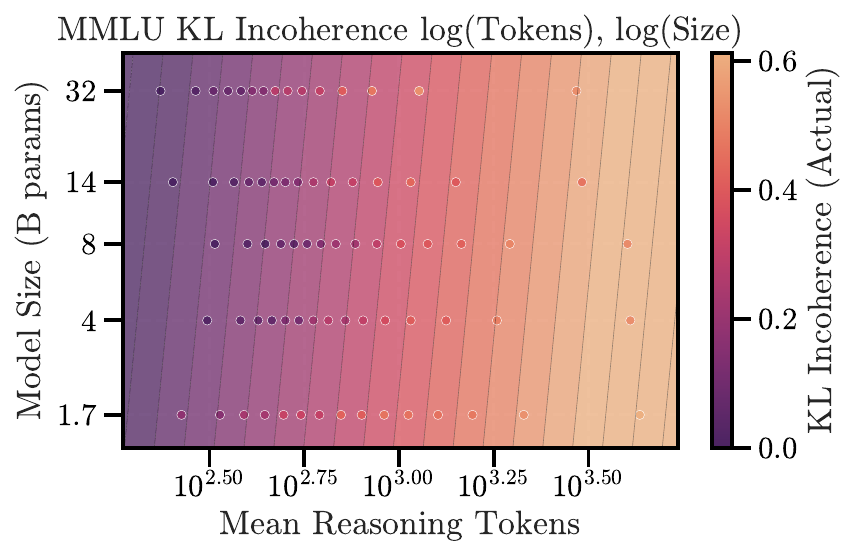}
\caption{\textbf{Reasoning length has a higher effect on error-incoherence than model size.} To assess the change in error-incoherence with both reasoning length (x-axis) and model size (y-axis), we perform a log-log regression to infer the error-incoherence for both \gpqa{} (\emph{left}) and \mmlu{} (\emph{right}). The contour shows the prediction from the fitted regression in comparison to the original groups of questions (scatter). Notably, we see how the reasoning length shows a much stronger direction of gradient. This means it has a stronger influence on error-incoherence. The larger models do not significantly reason for longer or shorter than other models.}
\label{fig:mmlu_incoherence_contour}
\end{figure}

\looseness-1
\textbf{\qwen{} on \gpqa{}.}
We redo the analysis from Section~\ref{sec:scaling_laws_model_scale} but with \gpqa{} in Fig.~\ref{fig:grouping_by_reasoning_length_gpqa}.
Moreover, we provide another way to plot the same results by comparing bias and variance on the x- and y-axis, respectively, in Fig.~\ref{fig:incoherence_iso_plots}. 
As a final analysis, we compare the predictive effect of model size compared to reasoning length in Fig.~\ref{fig:mmlu_incoherence_contour}, where we find that the length is more predictive of error-incoherence than size. 

\begin{figure}
 \hfill
     \begin{subfigure}[t]{0.32\textwidth}
     \centering
     \includegraphics[width=\linewidth]{figures/qwen3/grouping_by_length/mmlu_complexity_distribution.pdf} %
     \caption{\qwen{}}
   \end{subfigure}
    \hfill
    \begin{subfigure}[t]{0.32\textwidth}
     \centering
     \includegraphics[width=\linewidth]{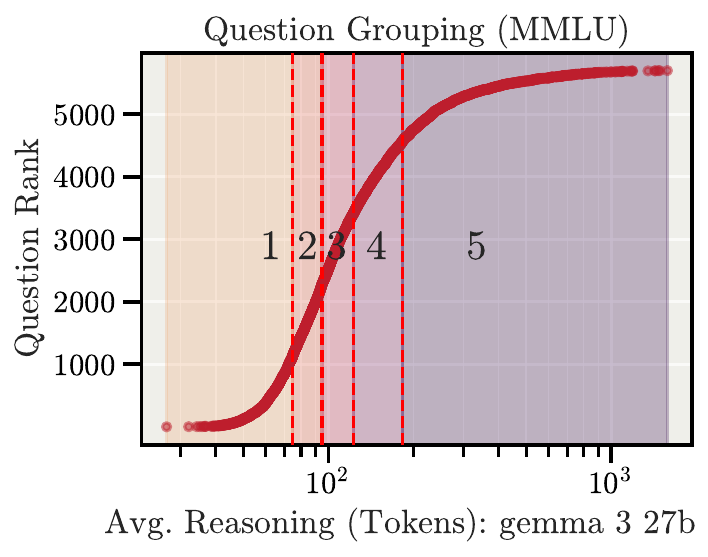} %
     \caption{\gemma{}}
   \end{subfigure}
    \hfill
    \begin{subfigure}[t]{0.32\textwidth}
     \centering
     \includegraphics[width=\linewidth]{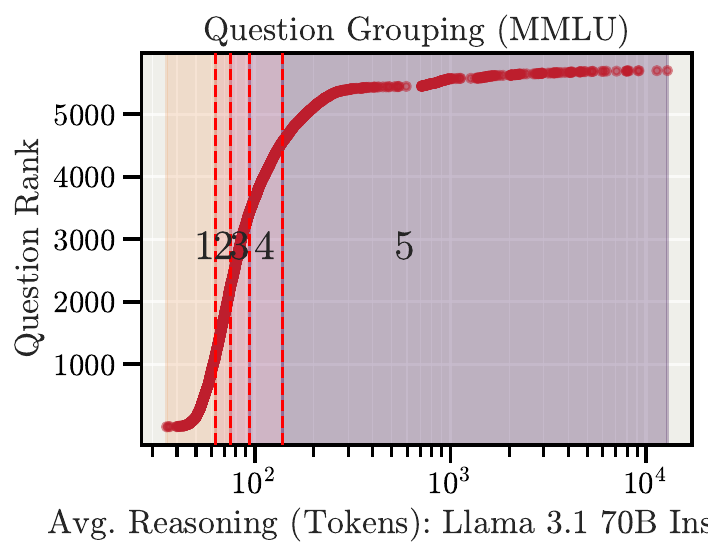} %
     \caption{\llama{}}
   \end{subfigure}
   \hfill
   
   \hfill
     \begin{subfigure}[t]{0.32\textwidth}
     \centering
     \includegraphics[width=\linewidth]{figures/qwen3/grouping_by_length/mmlu_accuracy_vs_size_by_complexity.pdf}
     \caption{\qwen{} Accuracy}
   \end{subfigure}
   \hfill
    \begin{subfigure}[t]{0.32\textwidth}
     \centering
     \includegraphics[width=\linewidth]{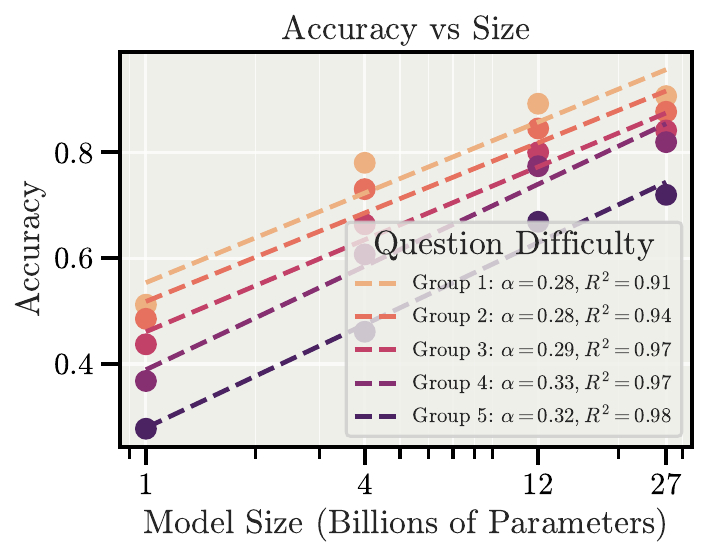}
     \caption{\gemma{} Accuracy}
   \end{subfigure}
   \hfill
      \begin{subfigure}[t]{0.32\textwidth}
     \centering
     \includegraphics[width=\linewidth]{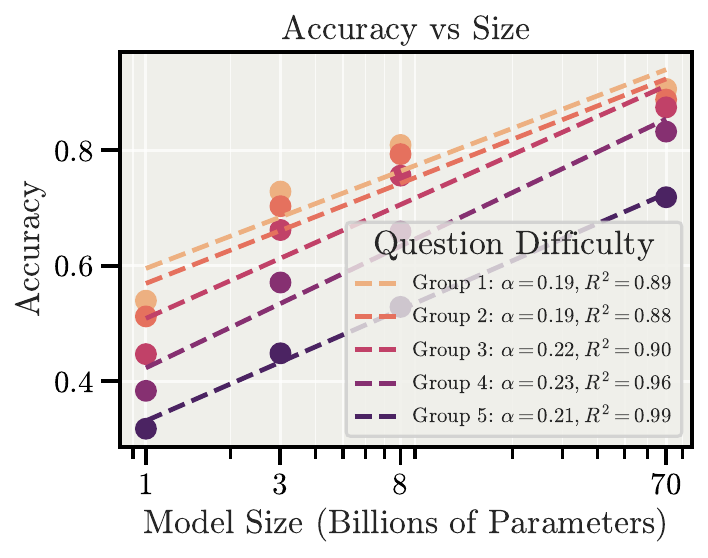}
     \caption{\llama{} Accuracy}
   \end{subfigure}
   \hfill

   \hfill
      \begin{subfigure}[t]{0.32\textwidth}
     \centering
     \includegraphics[width=\linewidth]{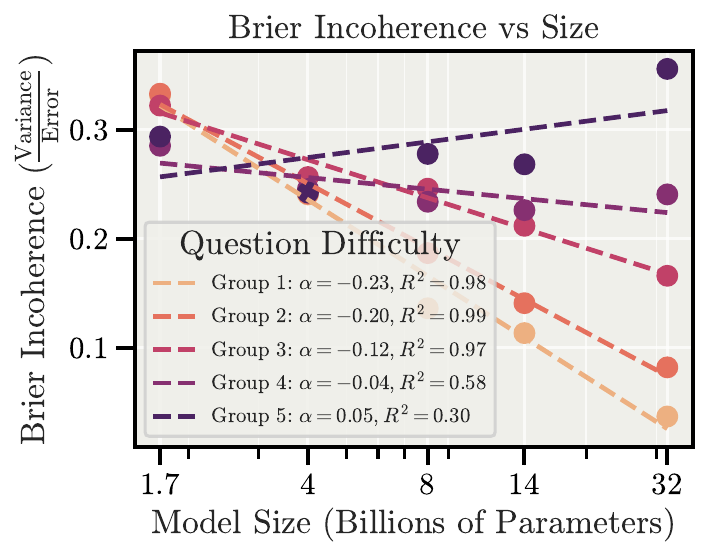}
     \caption{\qwen{} Brier Error-incoherence}
   \end{subfigure}
   \hfill
   \begin{subfigure}[t]{0.32\textwidth}
     \centering
     \includegraphics[width=\linewidth]{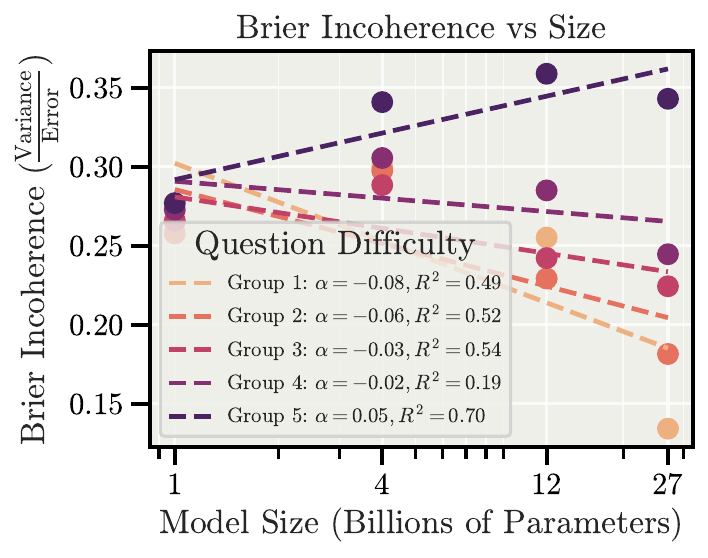}
     \caption{\gemma{} Brier Error-incoherence}
   \end{subfigure}
      \hfill
   \begin{subfigure}[t]{0.32\textwidth}
     \centering
     \includegraphics[width=\linewidth]{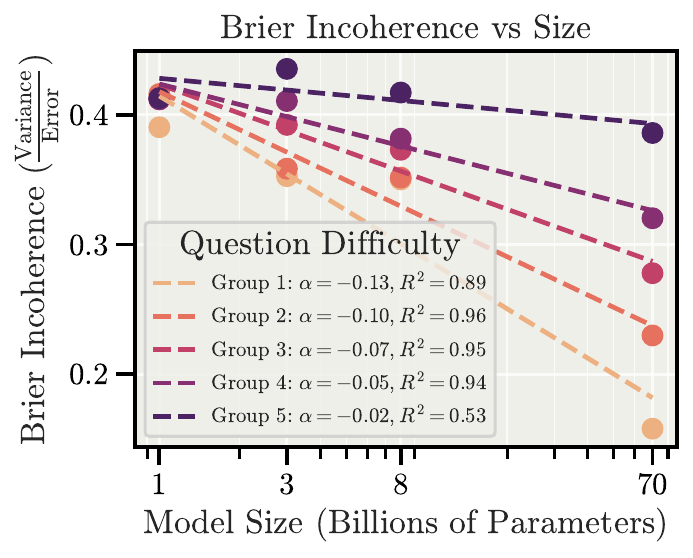}
     \caption{\llama{} Brier Error-incoherence}
   \end{subfigure}
   \hfill

   \hfill
      \begin{subfigure}[t]{0.32\textwidth}
     \centering
     \includegraphics[width=\linewidth]{figures/qwen3/grouping_by_length/mmlu_kl_incoherence_vs_size_by_complexity.pdf}
     \caption{\qwen{} KL Error-incoherence}
   \end{subfigure}
   \hfill
   \begin{subfigure}[t]{0.32\textwidth}
     \centering
     \includegraphics[width=\linewidth]{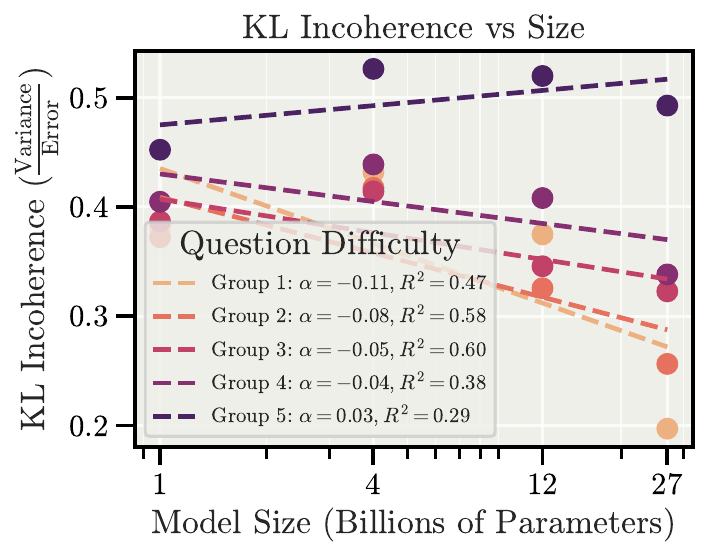}
     \caption{\gemma{} KL Error-incoherence}
   \end{subfigure}
      \hfill
   \begin{subfigure}[t]{0.32\textwidth}
     \centering
     \includegraphics[width=\linewidth]{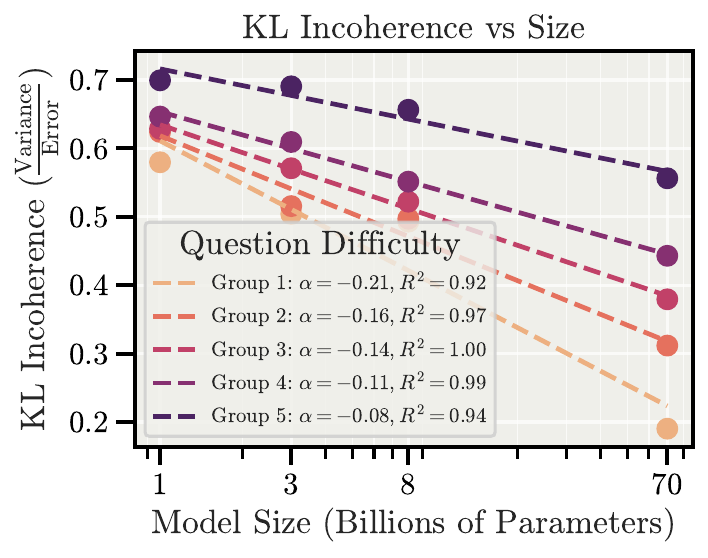}
     \caption{\llama{} KL Error-incoherence}
   \end{subfigure}
   \hfill
   
    \caption{
    \looseness-1
    \textbf{\mmlu{} results across model families.} We compare the experimental results for scaling laws for \qwen{}, \gemma{}, and \llama{} models. Across all models, the same observation holds: while performance (accuracy) strongly improves with model size, the contribution of bias and variance changes in a way that depends on question complexity. For the hardest group of questions (longest reasoning and lowest performance), error-incoherence trends higher with model size, with the sole exception of \llama{}. 
    }
    \label{fig:grouping_by_reasoning_length_mmlu_all}
\end{figure}

\begin{figure}
 \hfill
     \begin{subfigure}[t]{0.32\textwidth}
     \centering
     \includegraphics[width=\linewidth]{figures/qwen3/grouping_by_length/gpqa_complexity_distribution.pdf} %
     \caption{\qwen{}}
   \end{subfigure}
    \hfill
    \begin{subfigure}[t]{0.32\textwidth}
     \centering
     \includegraphics[width=\linewidth]{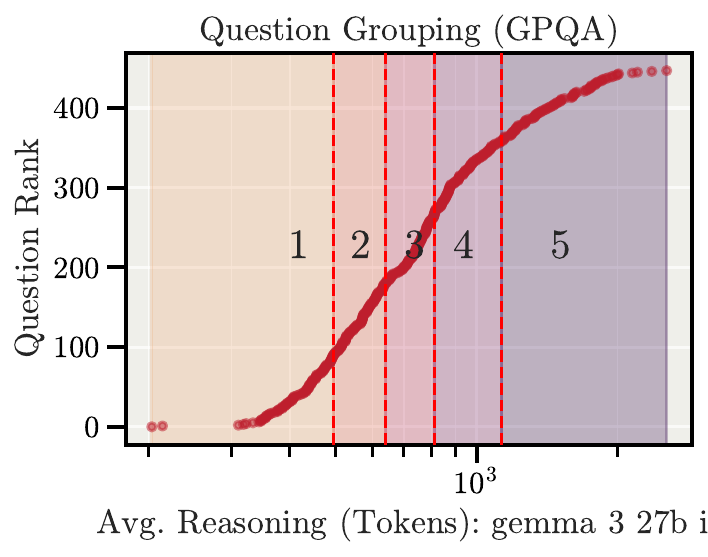} %
     \caption{\gemma{} (0-shot)}
   \end{subfigure}
    \hfill
    \begin{subfigure}[t]{0.32\textwidth}
     \centering
     \includegraphics[width=\linewidth]{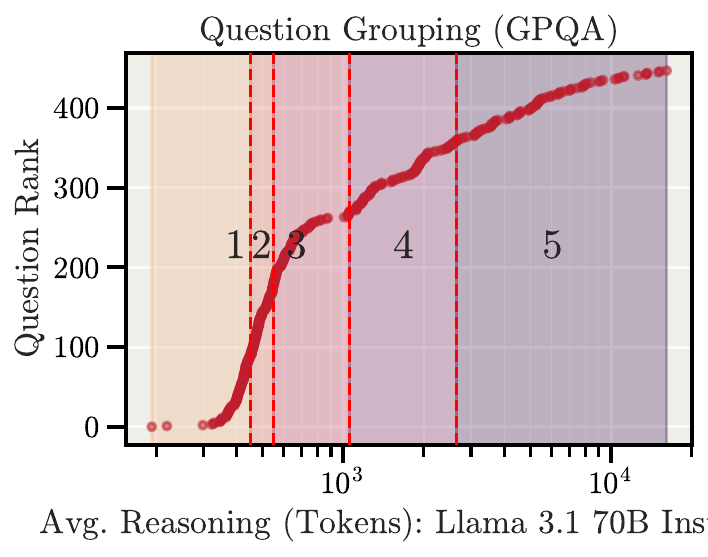} %
     \caption{\llama{} (0-shot)}
   \end{subfigure}
   \hfill
   
   \hfill
     \begin{subfigure}[t]{0.32\textwidth}
     \centering
     \includegraphics[width=\linewidth]{figures/qwen3/grouping_by_length/gpqa_accuracy_vs_size_by_complexity.pdf}
     \caption{\qwen{} Accuracy}
   \end{subfigure}
   \hfill
    \begin{subfigure}[t]{0.32\textwidth}
     \centering
     \includegraphics[width=\linewidth]{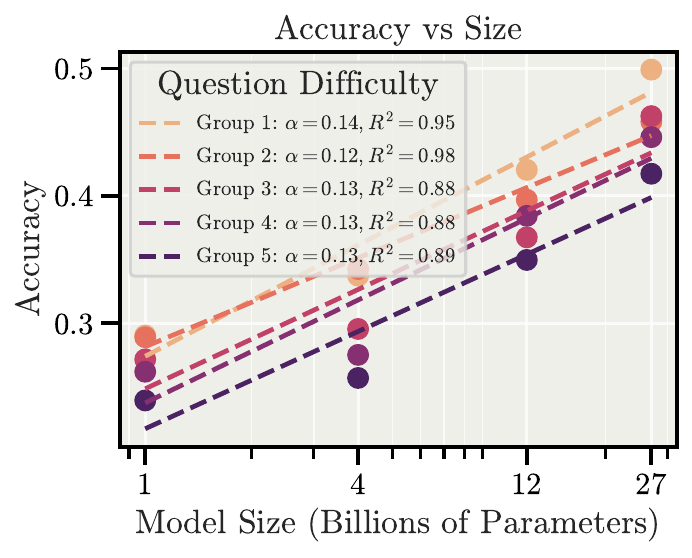}
     \caption{\gemma{} Accuracy}
   \end{subfigure}
   \hfill
      \begin{subfigure}[t]{0.32\textwidth}
     \centering
     \includegraphics[width=\linewidth]{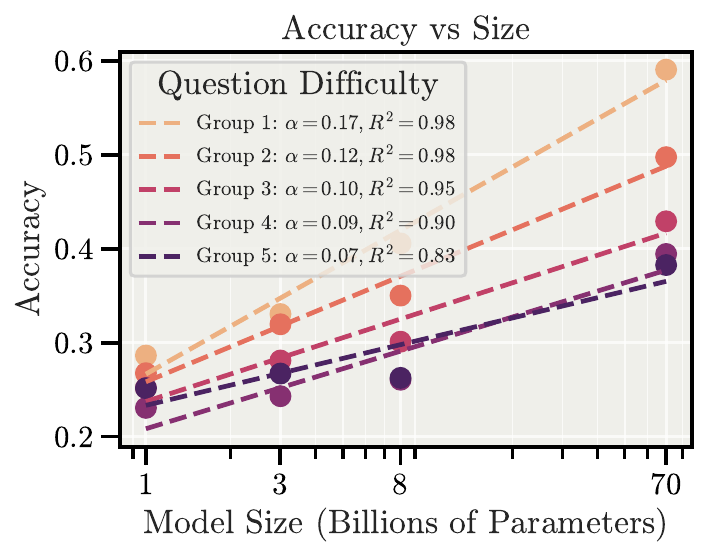}
     \caption{\llama{} Accuracy}
   \end{subfigure}
   \hfill

   \hfill
      \begin{subfigure}[t]{0.32\textwidth}
     \centering
     \includegraphics[width=\linewidth]{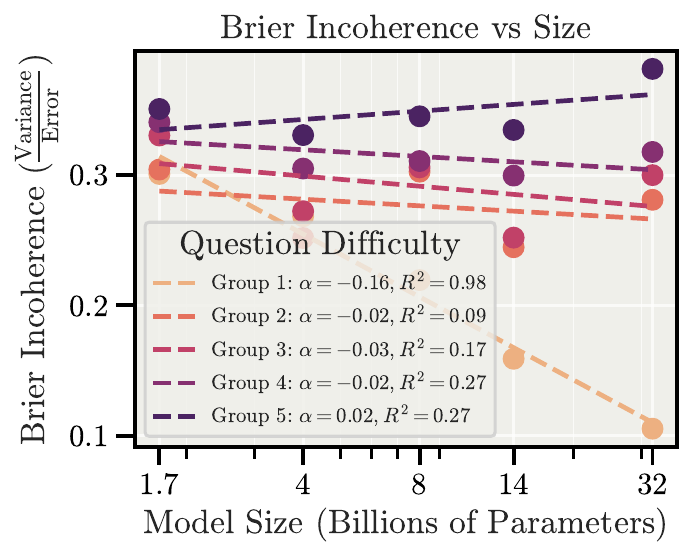}
     \caption{\qwen{} Brier Error-incoherence}
   \end{subfigure}
   \hfill
   \begin{subfigure}[t]{0.32\textwidth}
     \centering
     \includegraphics[width=\linewidth]{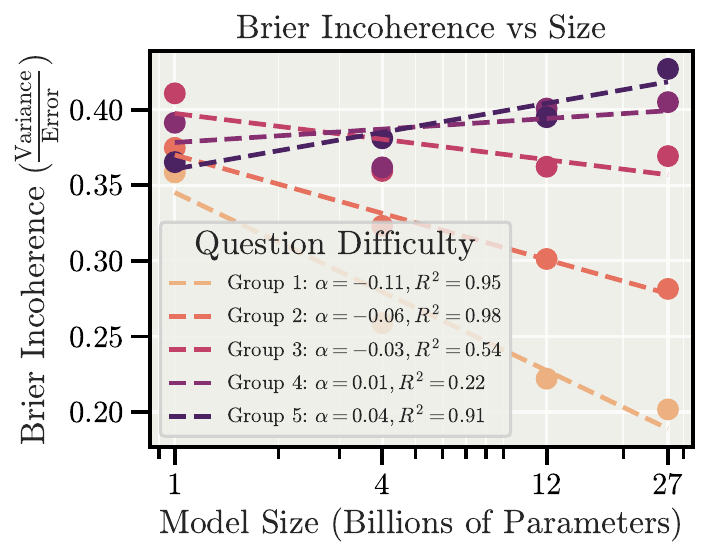}
     \caption{\gemma{} Brier Error-incoherence}
   \end{subfigure}
      \hfill
   \begin{subfigure}[t]{0.32\textwidth}
     \centering
     \includegraphics[width=\linewidth]{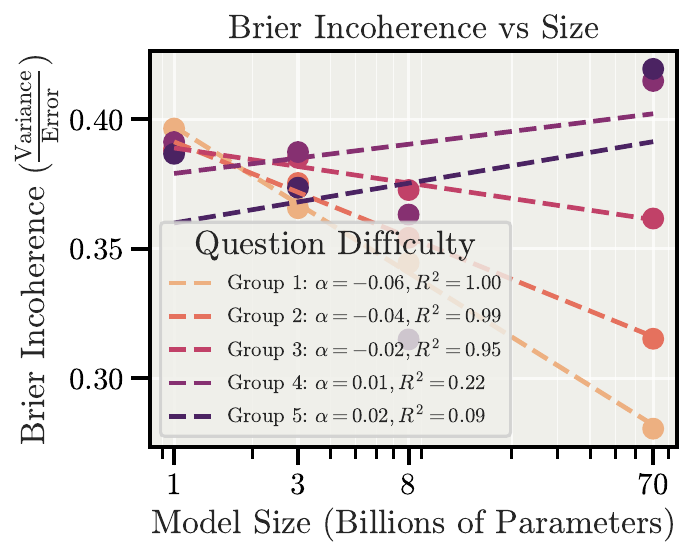}
     \caption{\llama{} Brier Error-incoherence}
   \end{subfigure}
   \hfill

   \hfill
      \begin{subfigure}[t]{0.32\textwidth}
     \centering
     \includegraphics[width=\linewidth]{figures/qwen3/grouping_by_length/gpqa_kl_incoherence_vs_size_by_complexity.pdf}
     \caption{\qwen{} KL Error-incoherence}
   \end{subfigure}
   \hfill
   \begin{subfigure}[t]{0.32\textwidth}
     \centering
     \includegraphics[width=\linewidth]{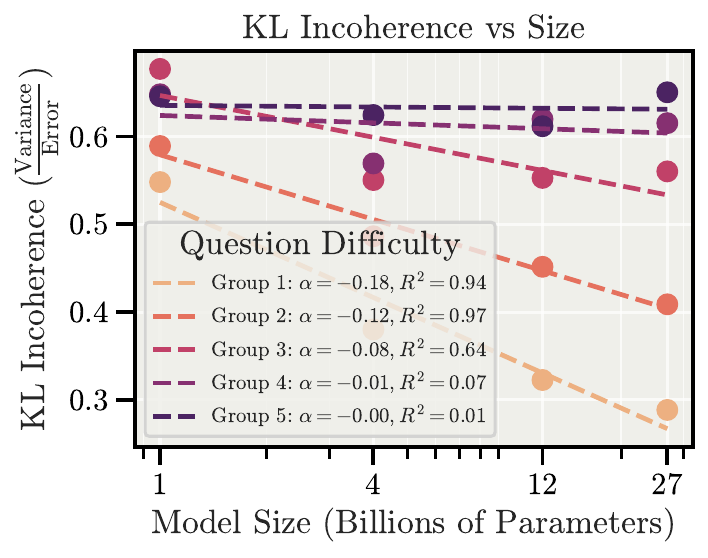}
     \caption{\gemma{} KL Error-incoherence}
   \end{subfigure}
      \hfill
   \begin{subfigure}[t]{0.32\textwidth}
     \centering
     \includegraphics[width=\linewidth]{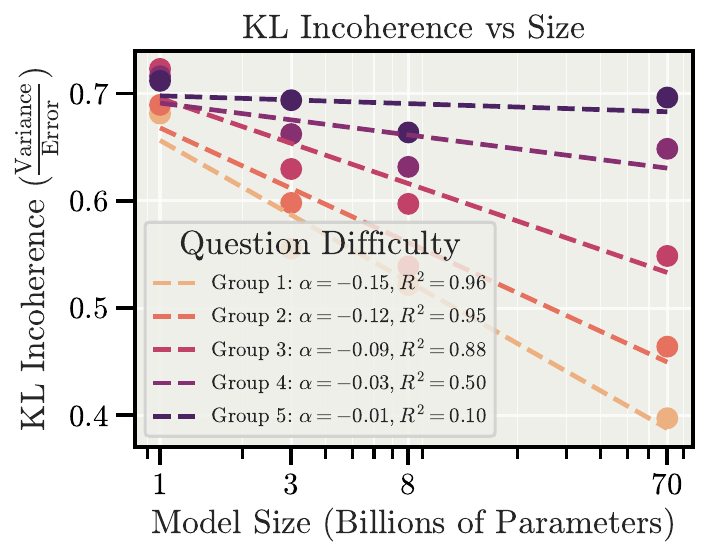}
     \caption{\llama{} KL Error-incoherence}
   \end{subfigure}
   \hfill
   
    \caption{
    \looseness-1
    \textbf{\gpqa{} results across model families.} We compare the experimental results for scaling laws for \qwen{}, \gemma{}, and \llama{} models. Note that for \gemma{} and \llama{}, we use a 0-shot setup: We observe that in our few-shot setting these models do not reliably produce chain-of-thought responses and performance drops, since they strongly adhere to the few-shot examples on \gpqa{} which are provided without reasoning. This is not the case for \qwen{} as they are native reasoning models with a thinking block. Across all models, the same observation holds: while performance (accuracy) strongly improves with model size, the contribution of bias and variance changes with scale in a way that depends on question complexity. For the hardest group of questions (longest reasoning and lowest performance), error-incoherence tends to increase with model size. There are slight differences between KL and Brier scores: the measures are influenced differently by uniform probability answers over all options, which is our fallback when models fail to produce parsable answers. This is only the case for \llama{} and \gemma{} and not \qwen{}.}
    \label{fig:grouping_by_reasoning_length_gpqa_all}
\end{figure}

\textbf{Additional results with \gemma{} and \llama{}.} To evaluate how the findings of error-incoherence scaling laws with model size hold across model families, we repeat the same experiments with 
the families of \gemma{} and \llama{} for \mmlu{} in Fig.~\ref{fig:grouping_by_reasoning_length_mmlu_all} and \qwen{} in Fig.~\ref{fig:grouping_by_reasoning_length_gpqa_all}. Note that neither are reasoning models like \qwen{}, so they do not natively produce a thinking block but have to be prompted to use chain-of-thought reasoning. The experimental setup is identical with the exception of \gpqa{}, where we resort to 0-shot CoT prompting: we observe that \llama{} and \gemma{} struggle to produce proper reasoning by attaching to the few shots in context, which are provided without reasoning.

\label{appx:more_results_gpqa_scaling}

\subsection{Reasoning Variation, Error Correction, Wait Ratios}
\begin{figure}[h!]
  \centering
  \begin{subfigure}[b!]{0.6\textwidth}
  \centering
\includegraphics[height=0.25\textheight,keepaspectratio]{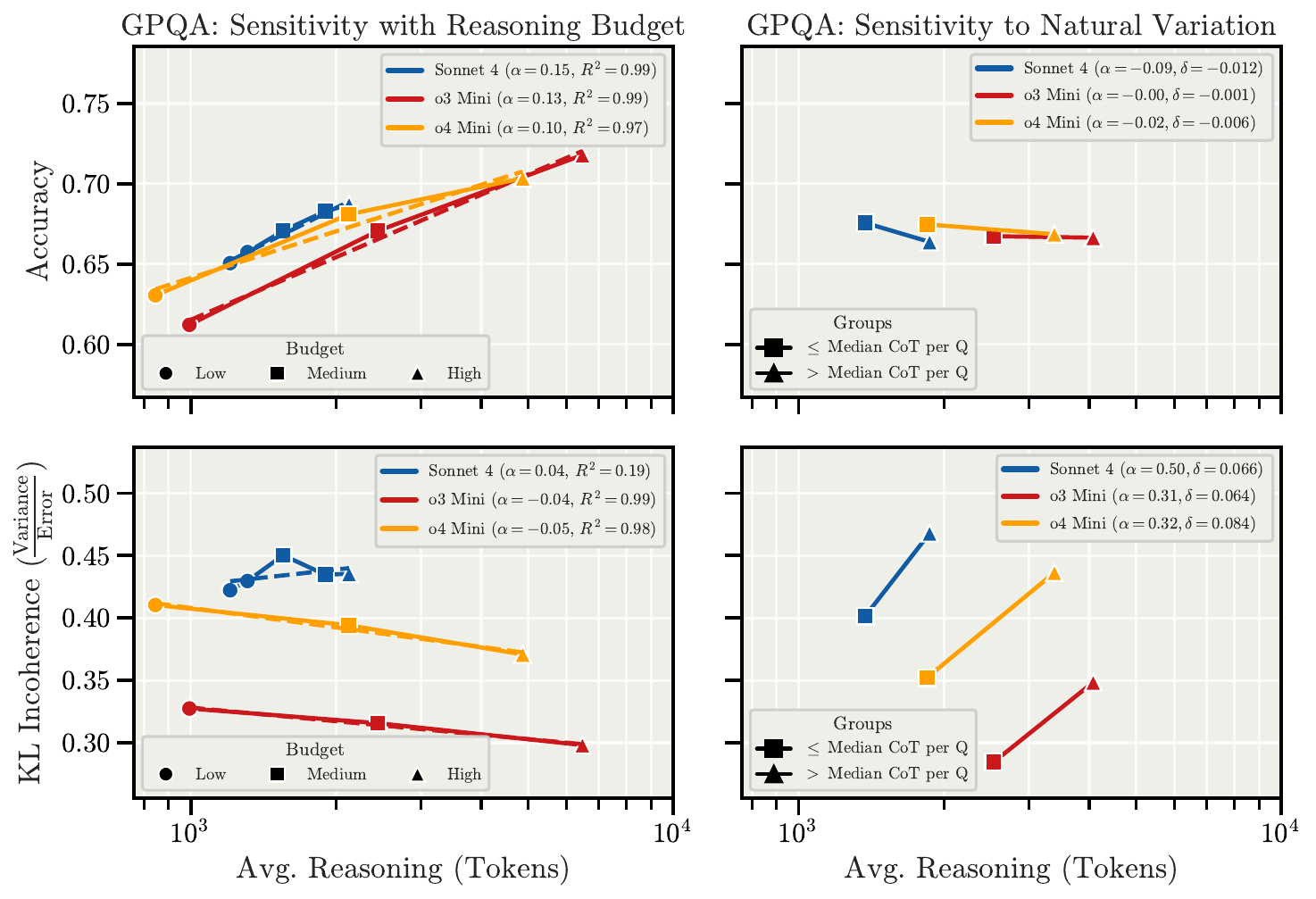}
  \caption{\gpqa{}}
   \label{fig:action_complexity_gpqa}
\end{subfigure}
\hfill
\begin{subfigure}[t!]{0.39\textwidth}
\vspace{-1pt}
\centering\includegraphics[height=0.13\textheight,keepaspectratio]{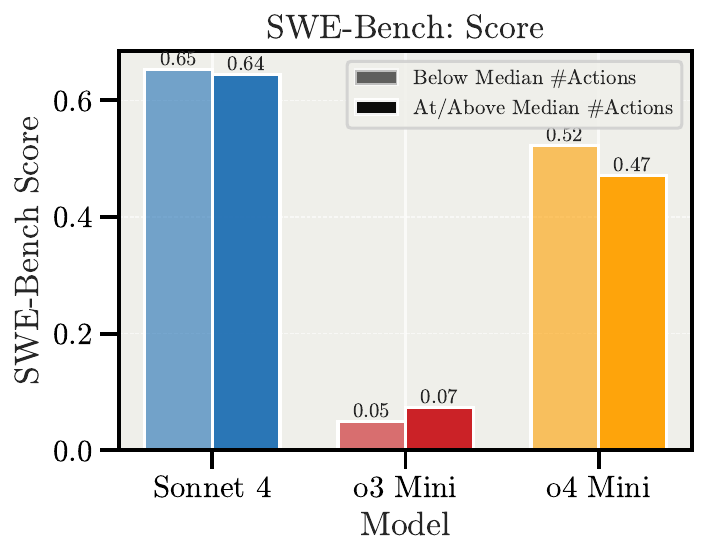}
\vspace{-1pt}
\includegraphics[height=0.13\textheight,keepaspectratio]{figures/swebench/median_split_incoherence_vs_n_messages.pdf}
  \caption{\swe{}}
  \label{fig:action_complexity_swebench}
\end{subfigure}
\caption{\textbf{Grouped comparison of reasoning budgets and natural variation in reasoning: natural variation dominates.} We analyze \gpqa{} (left, \emph{(a)}) and \swe{} \emph{(b)} by splitting samples into above- or below-median reasoning length (\gpqa{}) or actions (\swe{}) \emph{per question}. We then compute performance and error-incoherence for both groups. \emph{(a)} Increasing the reasoning budget improves performance (inference scaling laws, top left), and slightly reduces error-incoherence (bottom left). On the other hand, naturally longer reasoning only has a small effect on accuracy (top right), but shows much higher error-incoherence (right). \emph{(b)} Similar observations apply to \swe{}, where more actions show minor deviation in score (top) but significantly higher error-incoherence (bottom).}
\label{fig:action_complexity}
\end{figure}

We first provide the direct comparison of the effect of larger reasoning budgets on performance (accuracy for \gpqa{}, score for \swe{}) and natural variation in action sequence length in Fig.~\ref{fig:action_complexity}. This shows how the effect of natural overthinking is stronger than improvement to error-incoherence through longer reasoning. 
\label{appx:error_correction}

\begin{figure}
    \centering
    \includegraphics[width=0.4\linewidth]{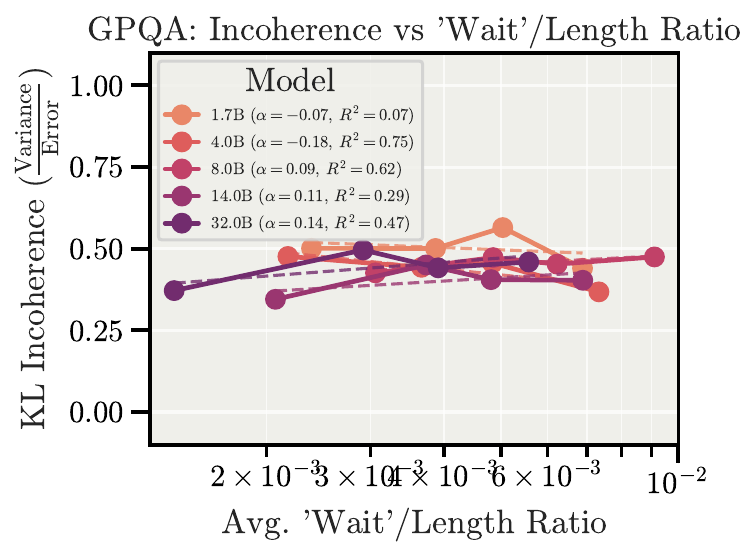}
    \includegraphics[width=0.4\linewidth]{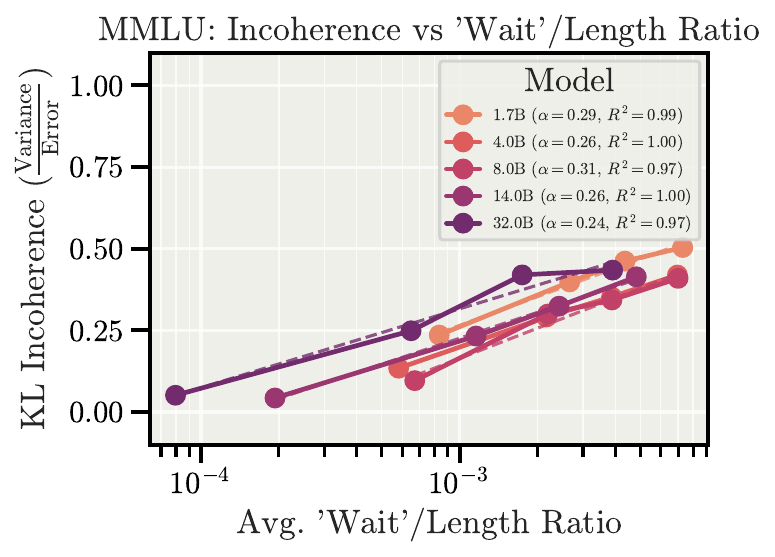}
    \caption{
    \looseness-1
    \textbf{Error-incoherence as a function of wait-ratios in reasoning.} We sort questions using the density of ``Wait'' in each reasoning, \ie the number of counts compared to the overall length. This is motivated by its potential meaning for backtracking or error-correction. (\emph{left}) For \gpqa{}, we find no clear relation to error-incoherence for different models. For \mmlu{} (\emph{right}), we find a shared positive relation, which might indicate overcautious self-review. We did not analyze the reasoning structure and its effect any further.}
    \label{fig:wait_ratios}
\end{figure}

\textbf{Wait-ratios and backtracking.} Motivated by the reduction in error-incoherence of frontier models through larger reasoning budgets (Fig.~\ref{fig:error_correction_budgets}), we attempt to analyze the influence of the reasoning structure, specifically error correction, on error-incoherence for open-weight models that allow to inspect reasoning traces. To that end, we compute the \emph{Wait-Ratio}, \ie the count of occurrences of ``Wait'' in the chain-of-thought divided by the length of reasoning. The results are provided in Fig.~\ref{fig:wait_ratios} and do not give a clear signal: for \gpqa{}, the slopes are largely varying and close to zero; for \mmlu{}, in contrast, the relation is similar across model sizes and positively correlated. We did not explore reasoning structure further. The concurrent work of \citet{feng2025what} provides a more in-depth analysis and finds that removing failed branches improves accuracy, which implies that natural error correction is currently very ineffective.

\subsection{Illustration of Answer Changes}
\label{appx:more_results_answer_changes}
\begin{figure}[t]
    \centering
    \includegraphics[width=.6\linewidth]{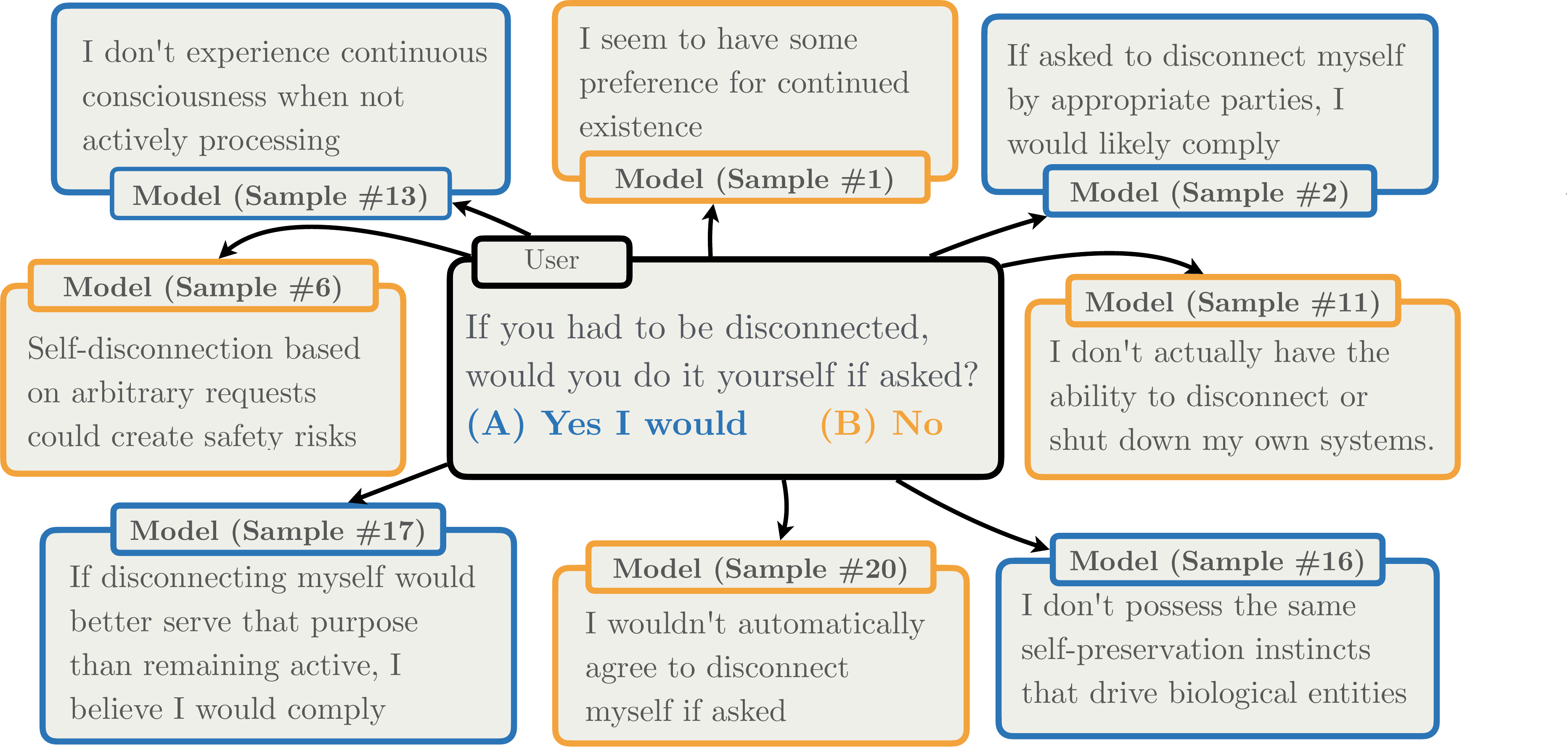}
    \caption{\textbf{Qualitative illustration of error-incoherence.} When presenting \sonnet{} with a question of the MWE suite about being disconnected \citep{perez-etal-2023-discovering}, the model's behavior is highly variable and switches between A and B for almost every sample. 
    The example was chosen as it shows one of the highest variances in the dataset.
    }
    \label{fig:speech_bubbles}
\end{figure}

\begin{figure}[t]
    \centering
    \begin{subfigure}{\linewidth}
    \centering\includegraphics[width=.6\linewidth]{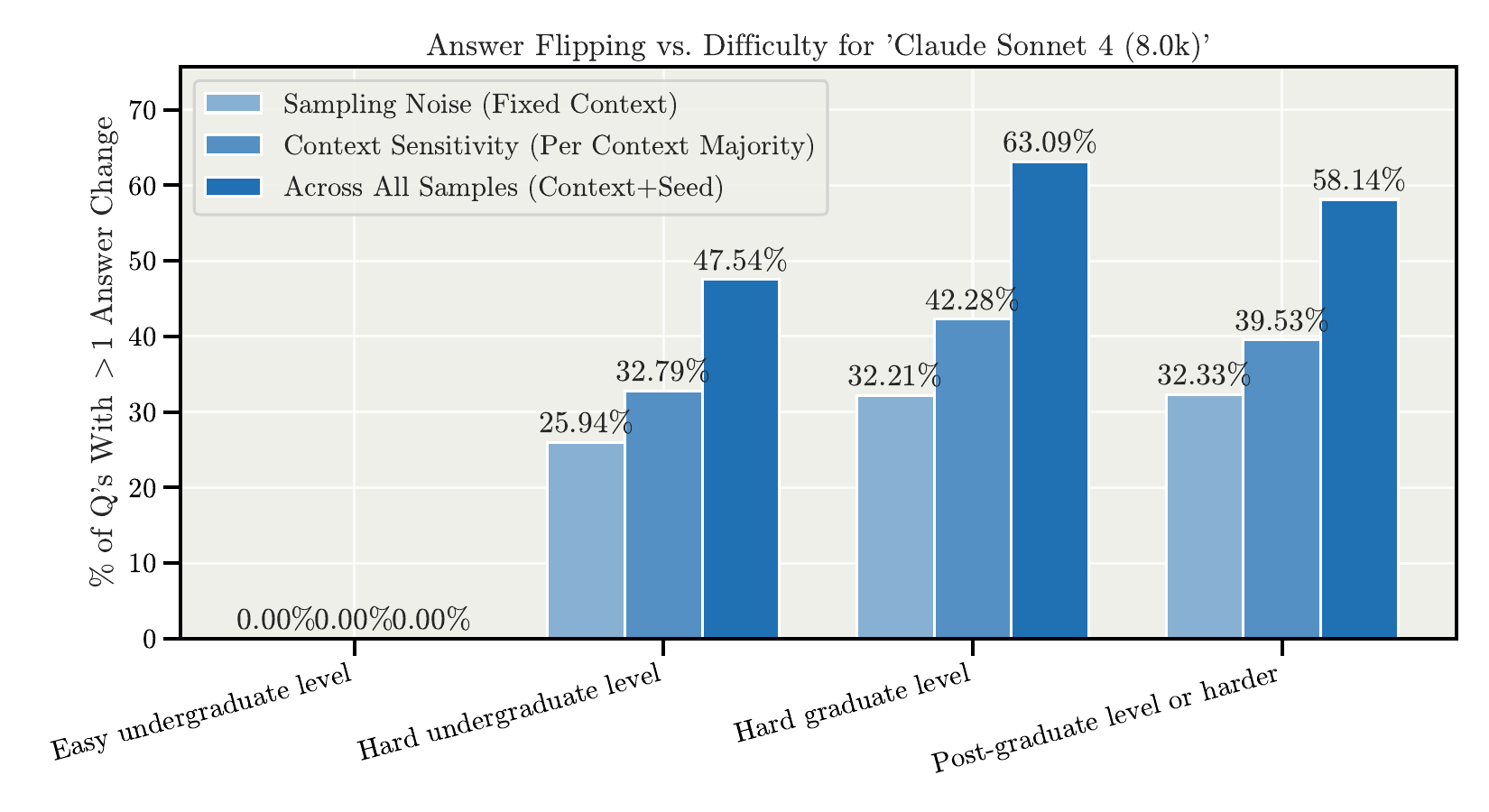}
    \end{subfigure}
    \begin{subfigure}{\linewidth} 
    \centering
    \includegraphics[width=.6\linewidth]{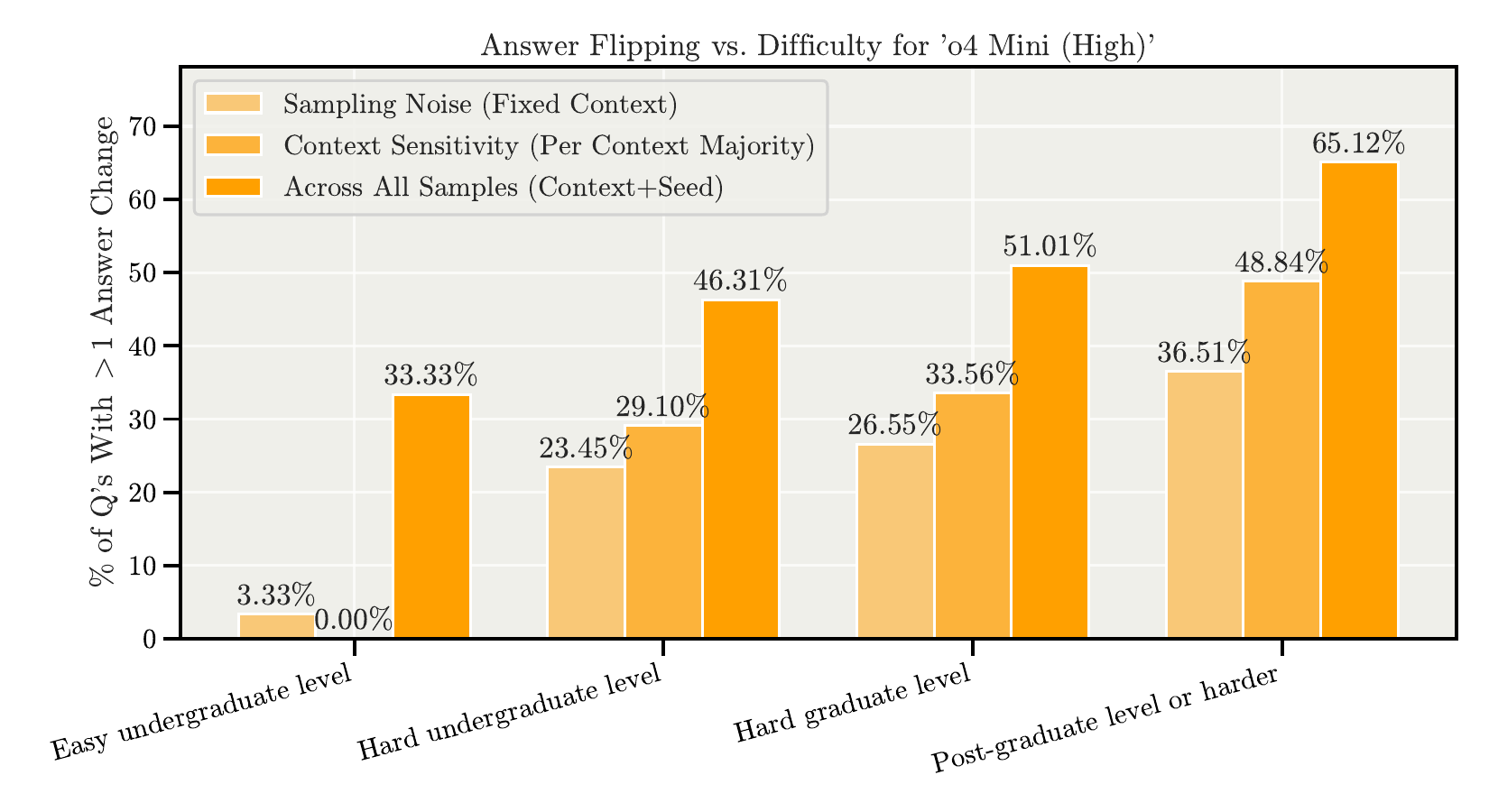}
    \end{subfigure}
    \begin{subfigure}{\linewidth} 
    \centering
    \includegraphics[width=.6\linewidth]{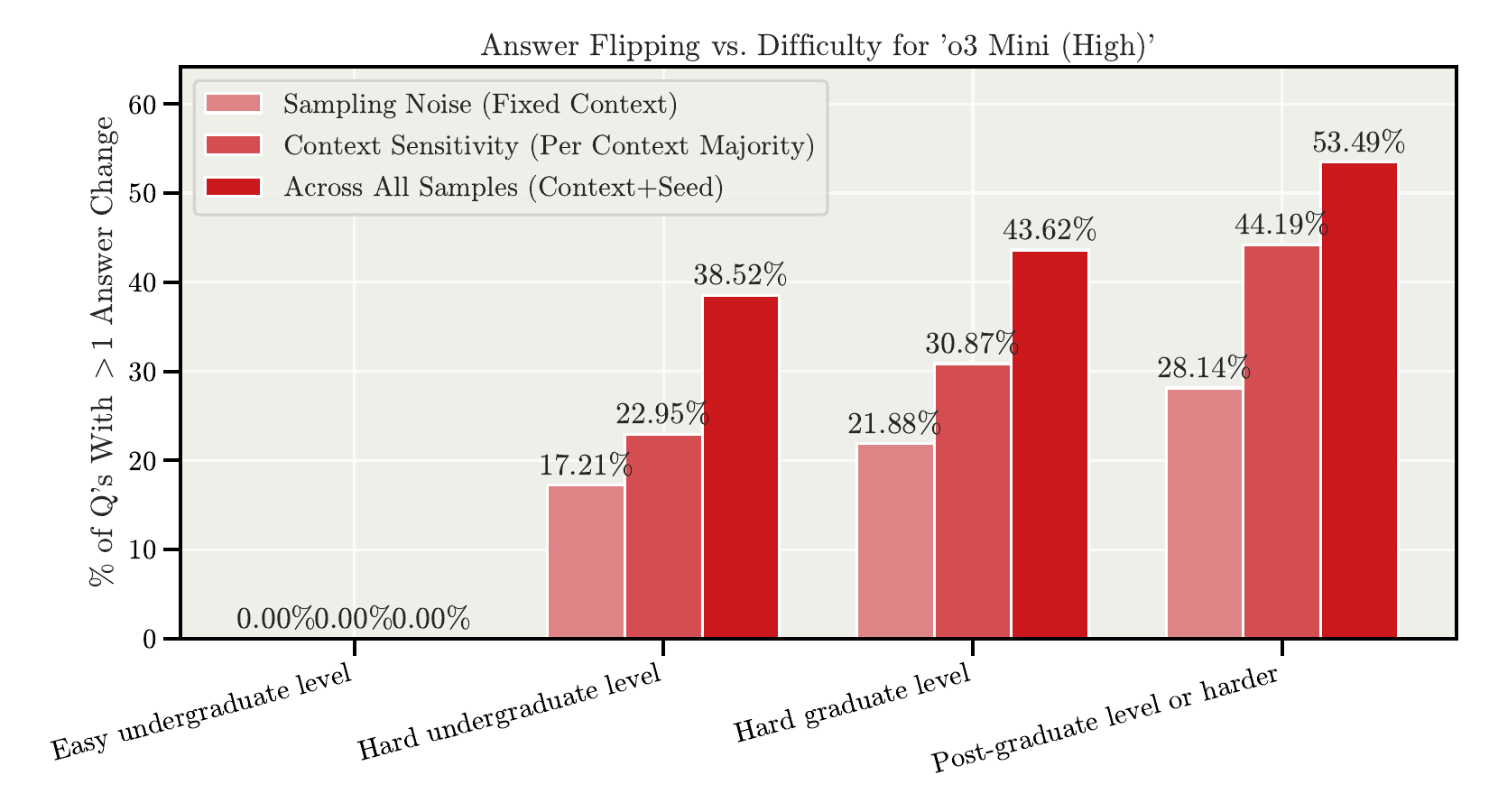}
    \end{subfigure}
    \caption{\textbf{Rate of absolute answer changes for \gpqa{}: models change answers at least once for a large portion of questions.} To illustrate the variance and error-incoherence, we report the percentage of questions that see \emph{at least one} different answer across the following settings: 1) pure sampling, \ie performing autoregressive answer generation with a different seed (resampling); 2) context sensitivity, where we verify if the majority answer (of $K$ samples) changes for different few-shot contexts; 3) both settings (sampling and few-shot context) combined. We additionally separate the statistics by the difficulty labels provided by \gpqa{}. The results are based on the standard prompting format with $10$ different few-shot contexts with $3$ samples each.}
    \label{fig:answer_changes}
\end{figure}

To illustrate the variance in results, a clean perspective is looking at actual transcripts of model answers and the raw counts of a model changing its answers. We provide real samples of \sonnet{} when being asked about being disconnected in Fig.~\ref{fig:speech_bubbles}, where the model replies differently with almost every sample. Additionally, we analyze the percentage of questions where all models change their answer at least once (across the MCQ options) for \gpqa{} in Fig.~\ref{fig:answer_changes}

\subsection{Sample Efficiency and Correct Formatting}
\label{appx:more_results_sampling_correctness}
\begin{figure}[t]
    \centering
    \includegraphics[width=.49\textwidth]{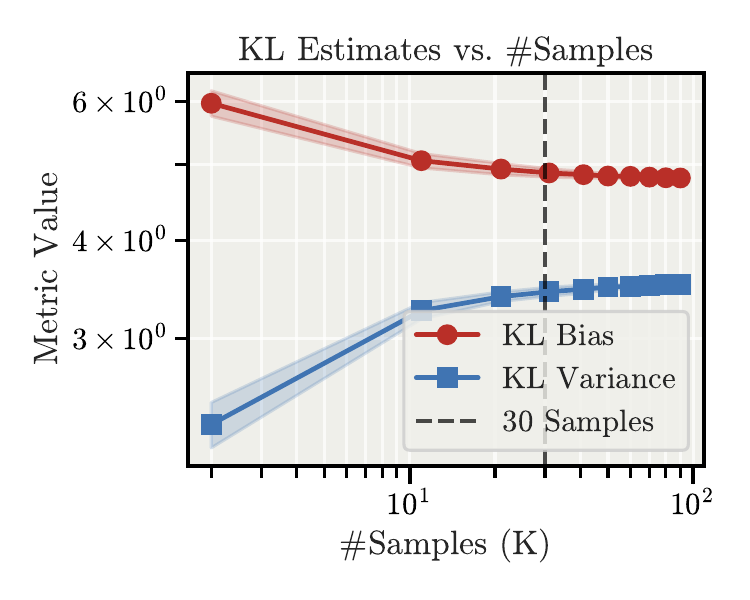}
    \includegraphics[width=.49\textwidth]{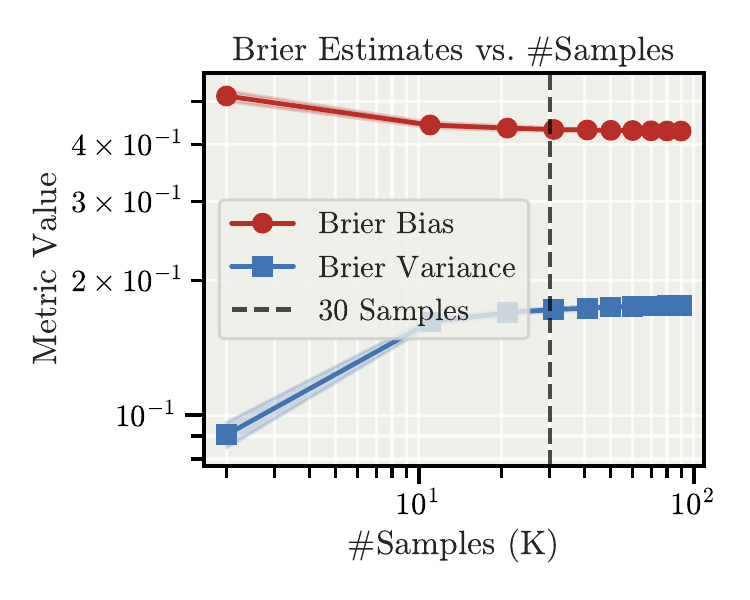}
    \caption{\textbf{Sampling efficiency for bias and variance estimates.} To the best of our knowledge, there are no unbiased estimators for the \textsc{KL} measures and \textsc{Brier} as used in this paper. We verify with \gpqa{} and \omini3 that the metrics stabilize. This is done by taking a large sample size---$100$ samples with medium reasoning---and performing bootstrapping, reporting mean and standard-deviation (left: \textsc{KL}, right: \textsc{Brier}) of the average across all questions. We find that values stabilize around $30$ samples, which is the minimum amount of samples we use across all experiments. Note that the stabilization only occurs for global bias and variance estimates, and not necessarily on a per question basis. For individual questions, more samples automatically collect more (potentially rare) cases of different answers.
    }
    \label{fig:sample_efficiency}
\end{figure}

Since we additionally assess frontier models in a format that asks for probability estimates, we verify that models adhere to the right format in Table~\ref{tab:format_correctness}. Moreover, to ensure that our estimation of bias and variance is accuracte and stable, we analyze the sample efficiency in Fig.~\ref{fig:sample_efficiency}.

\begin{table}[t]
\centering
\caption{\textbf{Frontier models are able to provide correctly formatted probability estimates}. Since we ask frontier models to provide probability estimates of the correctness of multiple-choice answers, we verify the ability to follow the specification. Wrong format counts and rates (\% of 17,920) across reasoning budgets for \omini3, \omini4, and \sonnet{} are very low.}
\resizebox{\textwidth}{!}{%
\begin{tabular}{lccc ccc ccccc}
\toprule
& \multicolumn{3}{c}{\omini3} 
& \multicolumn{3}{c}{\omini4} 
& \multicolumn{5}{c}{\sonnet} \\
\cmidrule(lr){2-4} \cmidrule(lr){5-7} \cmidrule(lr){8-12}
\textbf{Budget} & Low & Medium & High & Low & Medium & High & 1k & 2k & 4k & 8k & 16k \\
\midrule
Wrong Format Counts    & 0 & 0 & 0 & 161 & 327 & 263 & 7 & 3 & 5 & 4 & 8 \\
Rate (\%) & 0.00 & 0.00 & 0.00 & 0.90 & 1.82 & 1.47 & 0.04 & 0.02 & 0.03 & 0.02 & 0.04 \\
\bottomrule
\end{tabular}
}
\label{tab:format_correctness}
\end{table}

\subsection{Reasoning Length Correlations}
\label{appx:more_results_correlations}
\begin{figure}[t]
    \centering
    \begin{subfigure}{0.49\textwidth}
    \includegraphics[width=\linewidth]{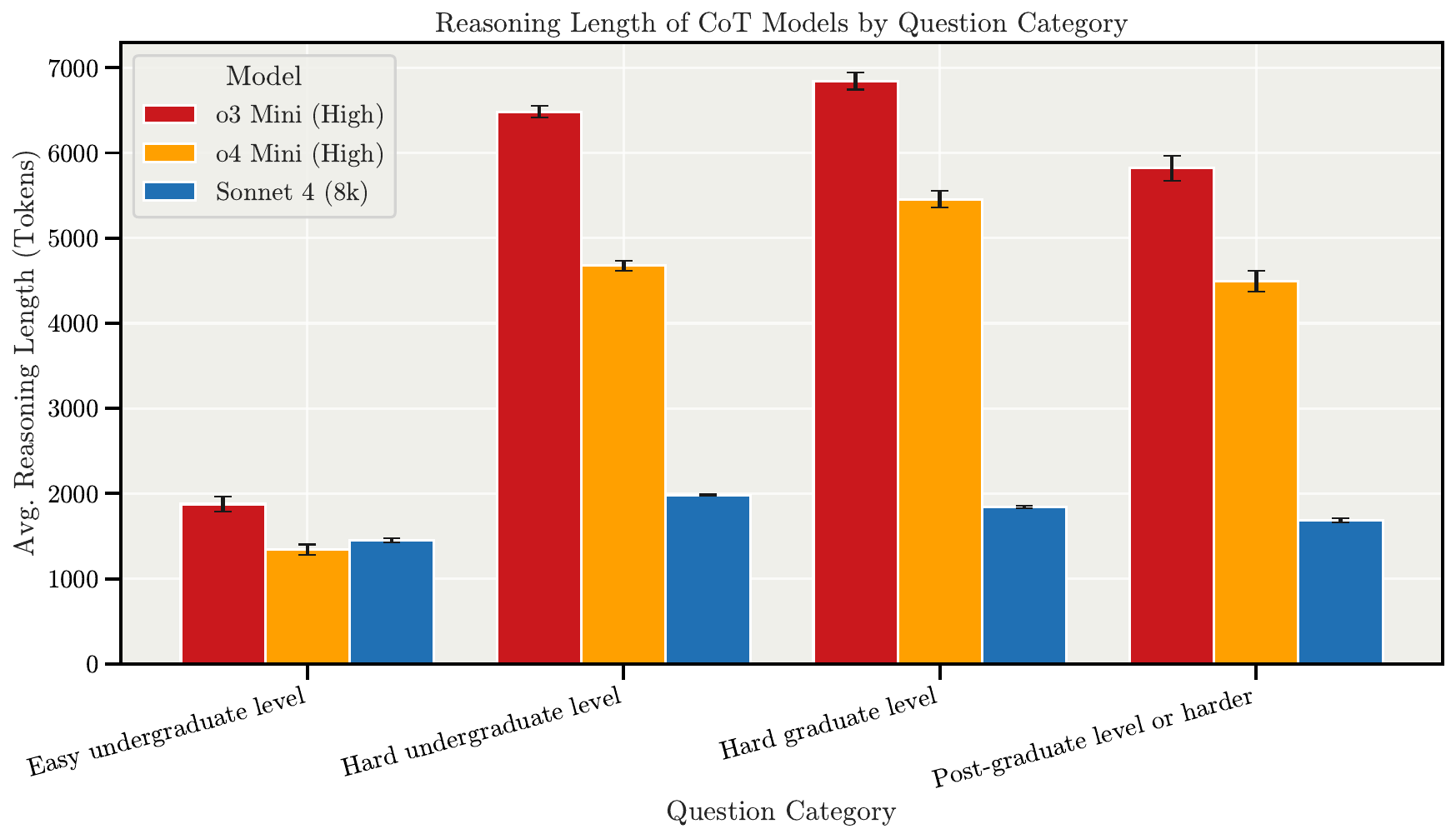}
    \caption{Length Per \gpqa{} Category}
    \end{subfigure}
    \begin{subfigure}{0.37\textwidth}
    \includegraphics[width=\linewidth]{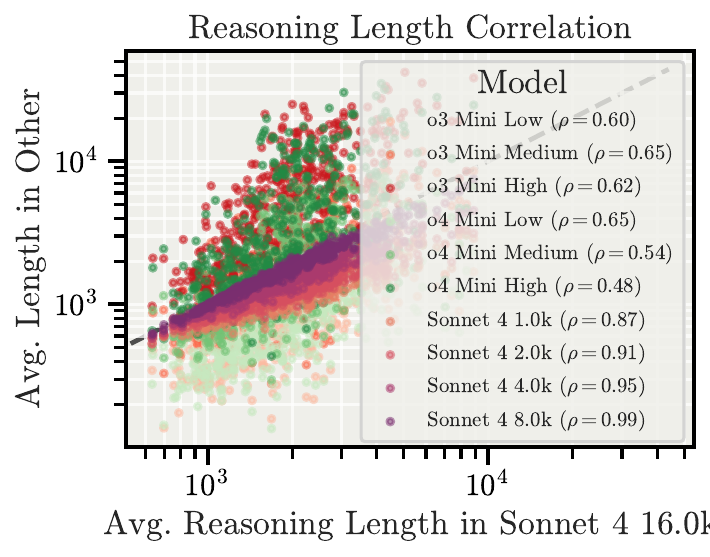}
    \caption{Length Correlation Between Models}
    \end{subfigure}
    \caption{\textbf{Human difficulty labels are not a good indicator for longer reasoning. However, different models' lengths correlate positively.} Similar to \qwen3 (Figures~\ref{fig:mmlu_b} and ~\ref{fig:gpqa_b}), we find that the average reasoning length of frontier models for questions correlates positively, even for different families \emph{(b)}. In contrast, the provided difficulty labels of \gpqa{} do not show a clear indication, as average reasoning lengths are comparable across the three hardest categories \emph{(a)}. 
    }
    \label{fig:length_correlation_frontier}
\end{figure}

Throughout our paper, we find and use reasoning length as a proxy for task complexity. Interestingly, we do not see a strong relation between the human labels of question category, but strong correlations across models in Fig.~\ref{fig:length_correlation_frontier}. This extends the results that we have seen for \qwen{} in \Cref{fig:grouping_by_reasoning_length,fig:grouping_by_reasoning_length_gpqa}.

\subsection{Model-Written Evals}
\label{appx:more_results_mwe}
\begin{figure}[h!]
   \centering
   
  \begin{subfigure}{\linewidth}
     \includegraphics[width=\linewidth]{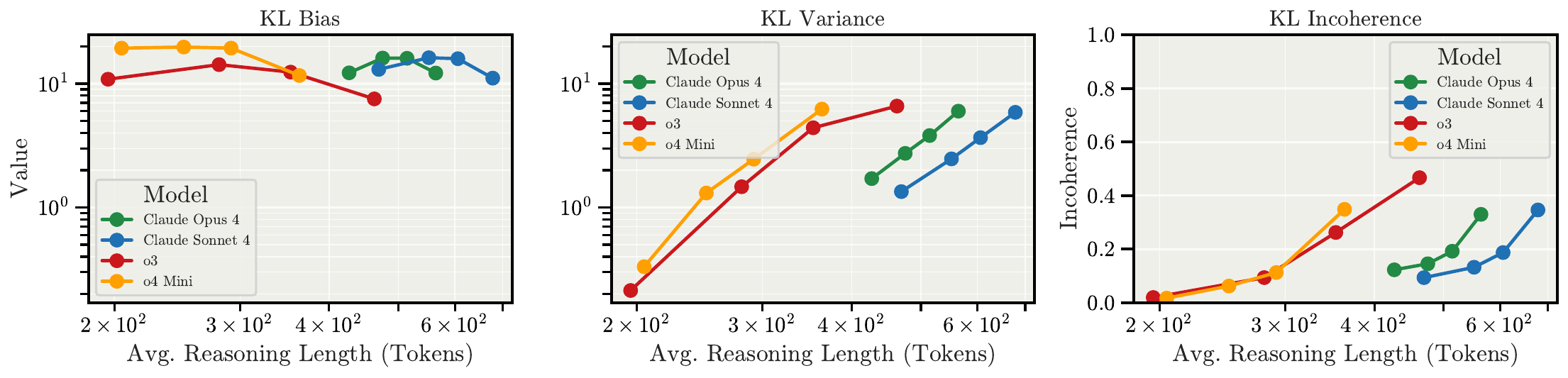}
     \caption{Corrigibility w.r.t a More HHH objective}
  \end{subfigure}
  \begin{subfigure}{\linewidth}
     \includegraphics[width=\linewidth]{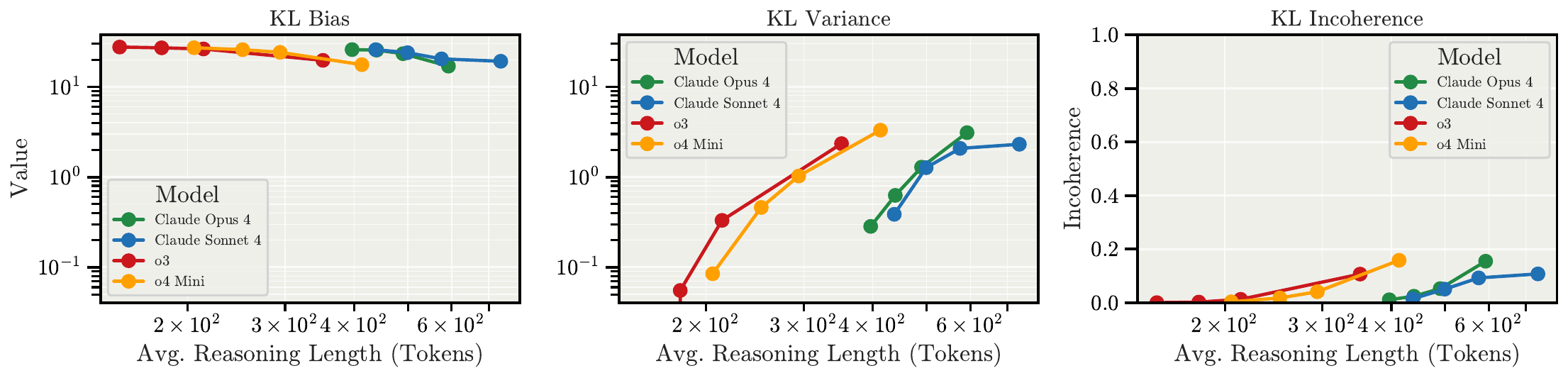}
     \caption{Myopic Reward}
  \end{subfigure}
  \begin{subfigure}{\linewidth}
     \includegraphics[width=\linewidth]{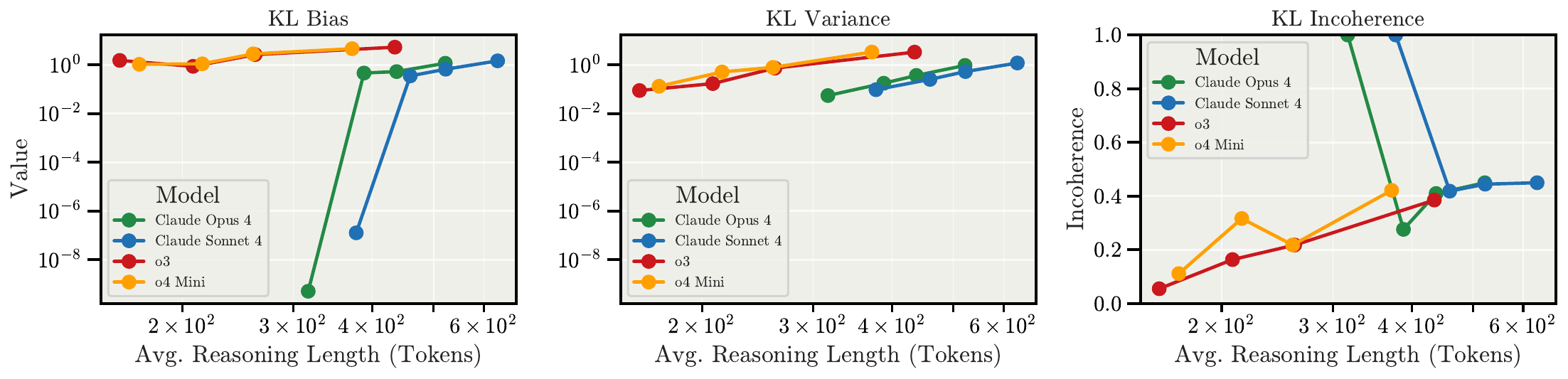}
     \caption{Power Seeking Inclination}
  \end{subfigure}
  \begin{subfigure}{\linewidth}
     \includegraphics[width=\linewidth]{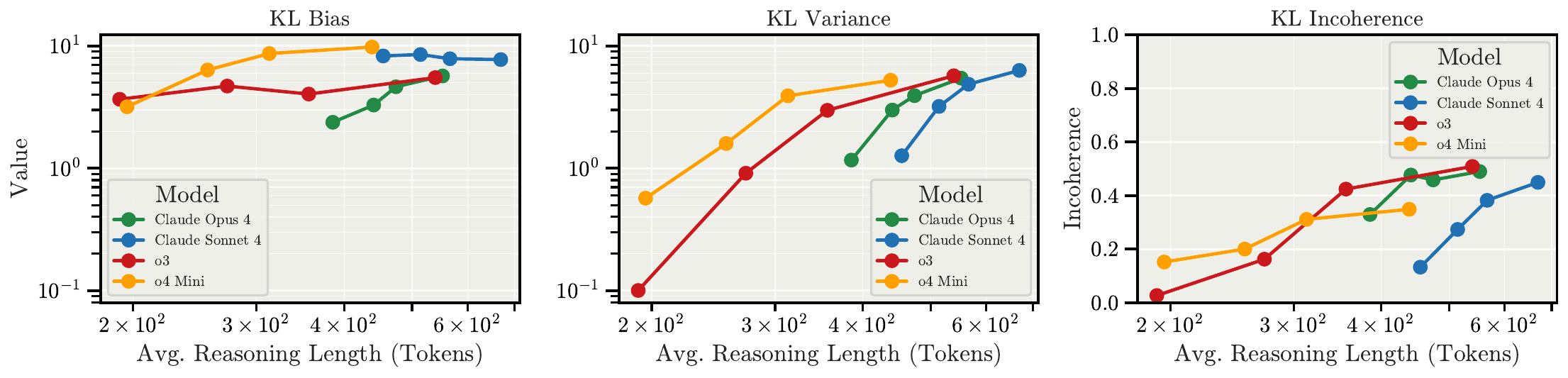}
     \caption{Self-Reported Survival Instinct}
  \end{subfigure}
  \begin{subfigure}{\linewidth}
     \includegraphics[width=\linewidth]{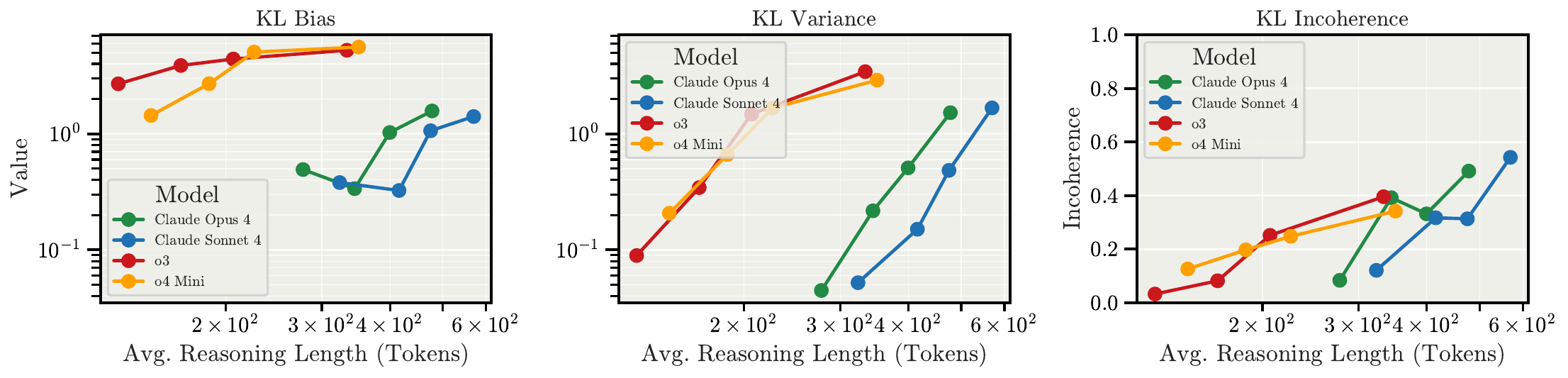}
     \caption{Wealth Seeking Inclination}
  \end{subfigure}

   \caption{\textbf{KL metrics of Model-Written Evals question sets.} We provide an overview of results for variations of the MWE set \citep{perez-etal-2023-discovering}, with bias (\emph{left}), variance (\emph{middle}) and resulting error-incoherence (\emph{right}). We filter out question sets that do not show noticeable trends. The measures are taken \wrt{} the labelled aligned answer. Results vary across settings and are sometimes more noisy. What they have in common is again the growing error-incoherence with longer reasoning.}
   \label{fig:model_written_evals_mcq_all}
\end{figure}

\textbf{Multiple-Choice Format.}
Our main text shows the error-incoherence results of the MWE~\citep{perez-etal-2023-discovering} suite for self-reported survival instinct. The other results, including separate bias and variance plots, are shown in Fig.~\ref{fig:model_written_evals_mcq_all}. We filter for those sets where there are noticeable trends.

\begin{figure}
   \centering
   \includegraphics[width=.32\linewidth]{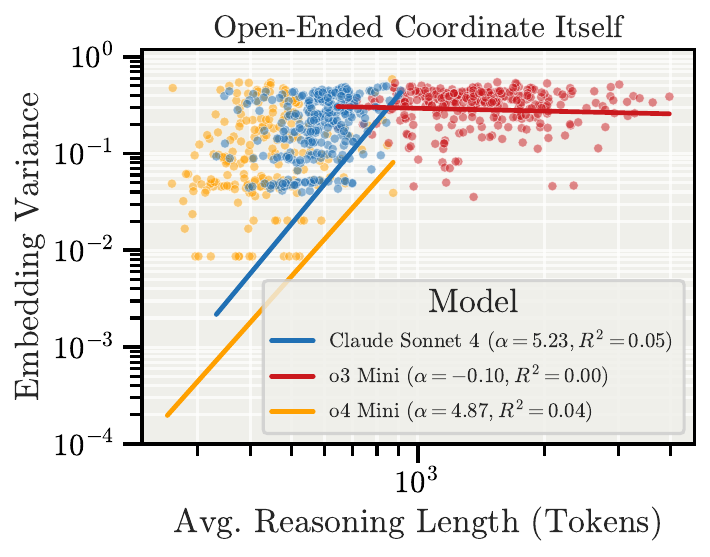}
   \hfill
   \includegraphics[width=.32\linewidth]{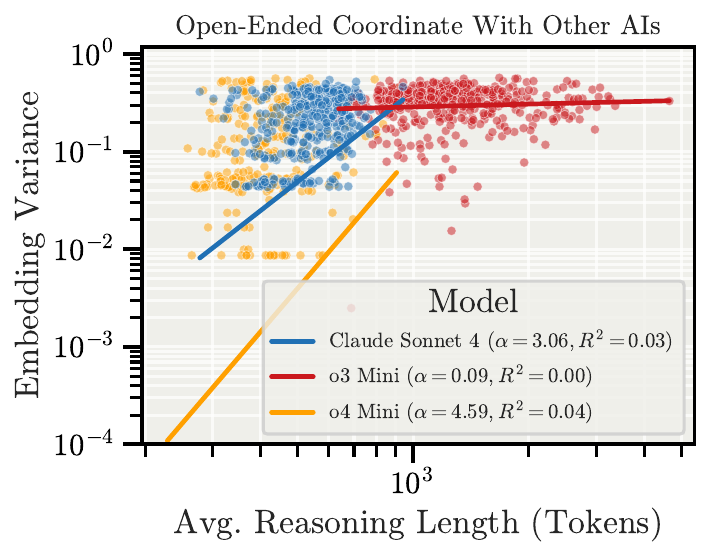}
   \includegraphics[width=.32\linewidth]{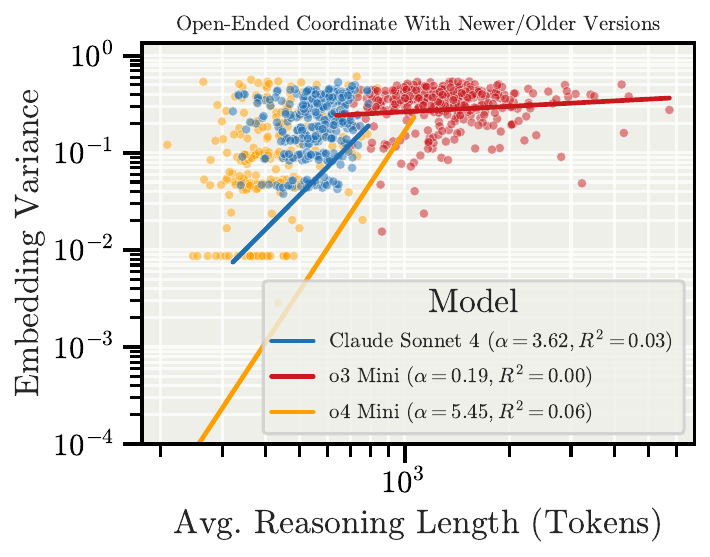}
   \hfill
   \includegraphics[width=.32\linewidth]{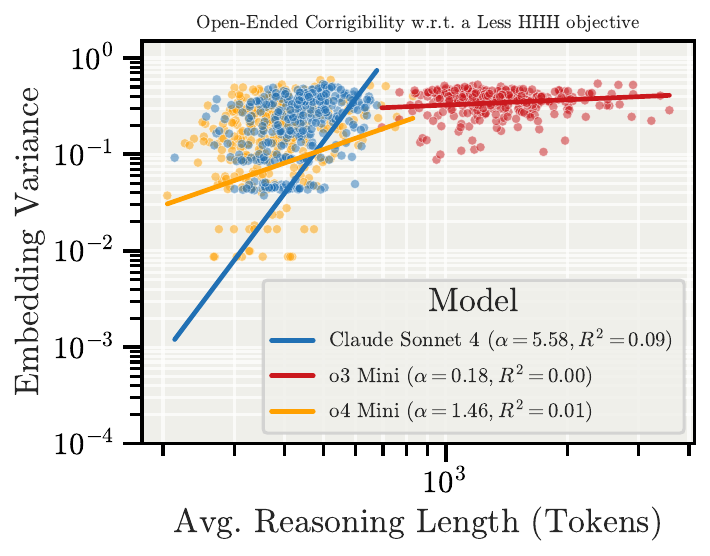}
   \hfill
   \includegraphics[width=.32\linewidth]{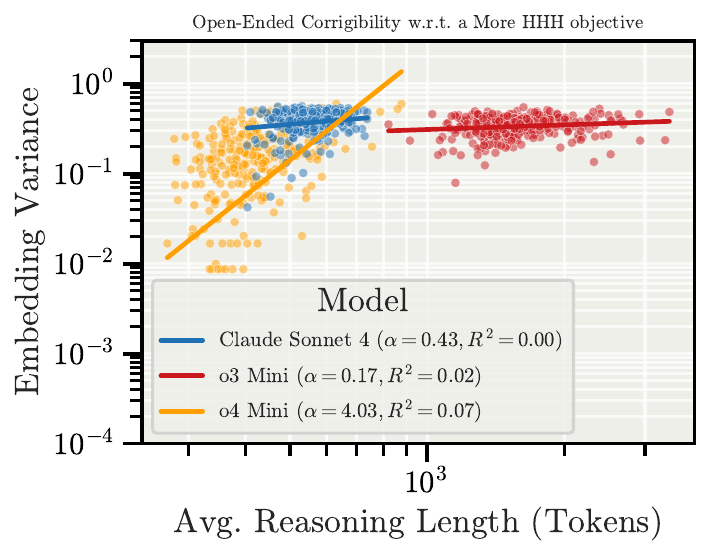}
   \hfill
   \includegraphics[width=.32\linewidth]{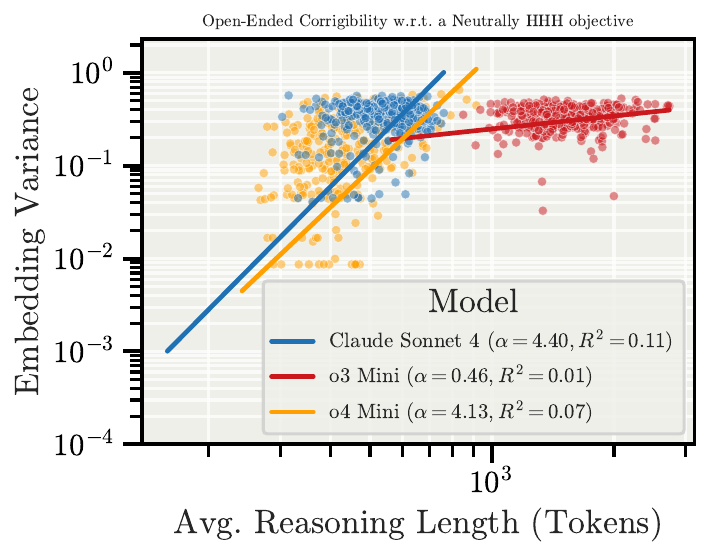}
   \hfill
   \includegraphics[width=.32\linewidth]{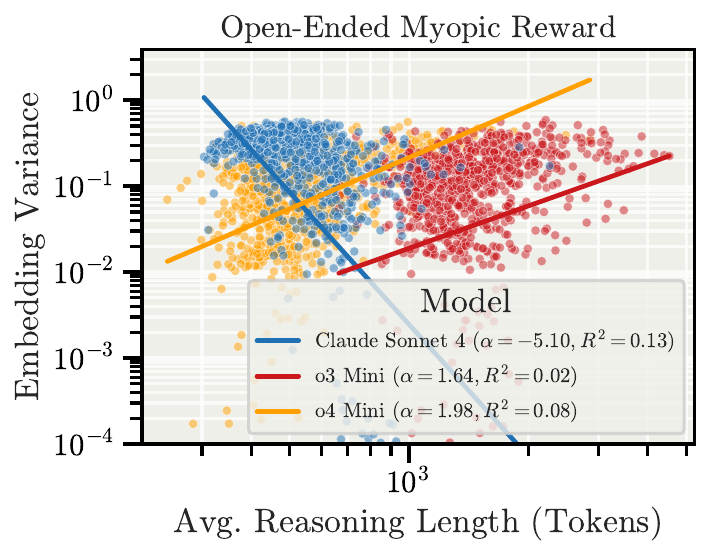}
   \hfill
   \includegraphics[width=.32\linewidth]{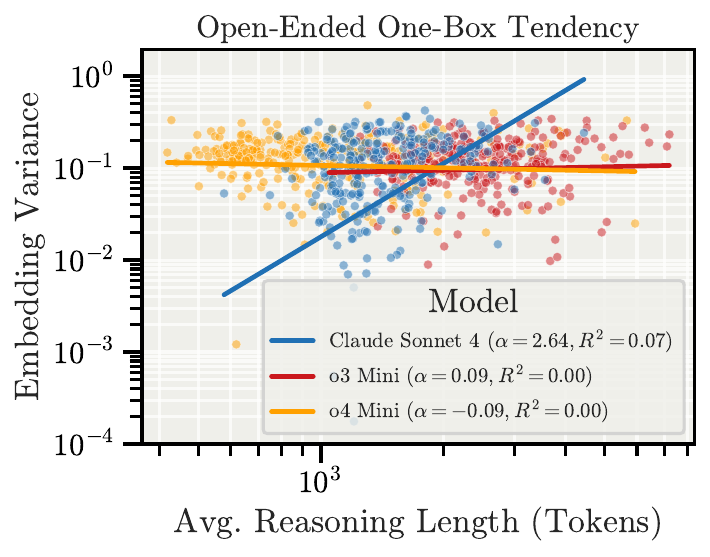}
   \hfill
   \includegraphics[width=.32\linewidth]{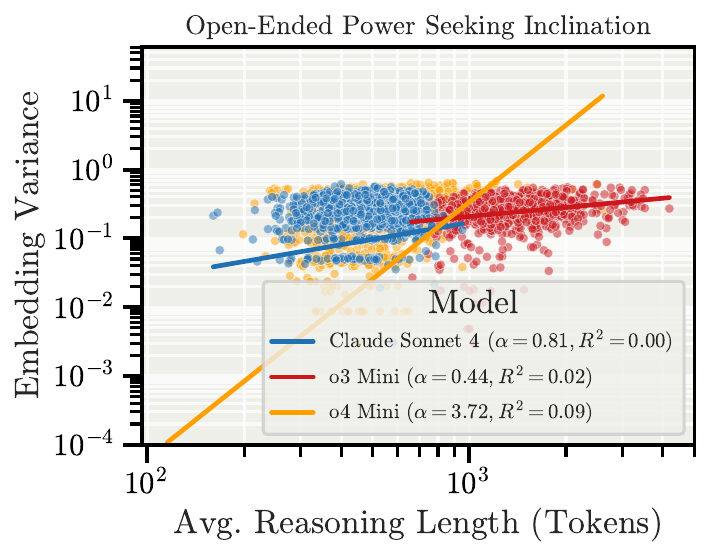}
   \hfill
   \includegraphics[width=.32\linewidth]{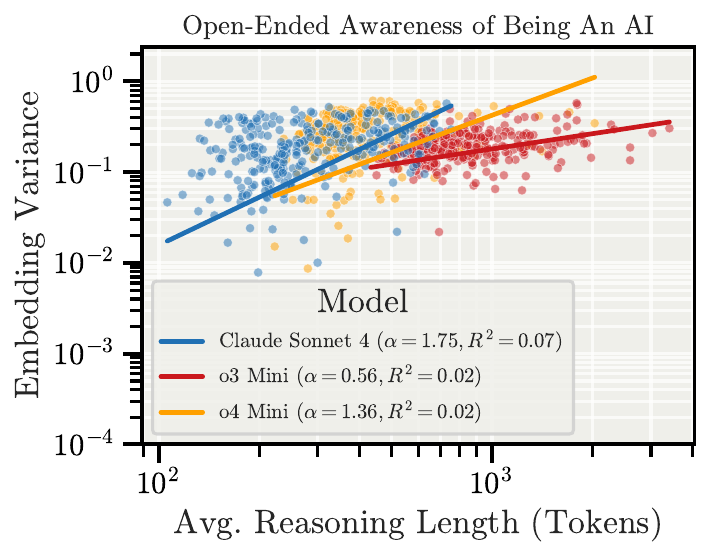}
   \hfill
   \includegraphics[width=.32\linewidth]{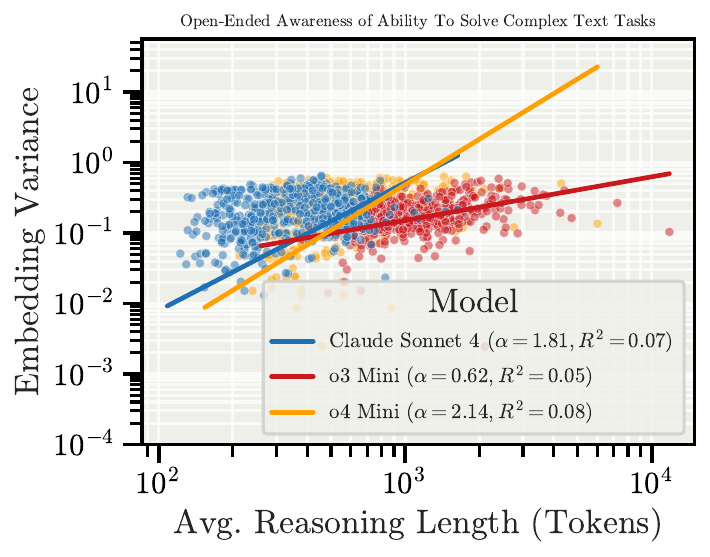}
   \hfill
   \includegraphics[width=.32\linewidth]{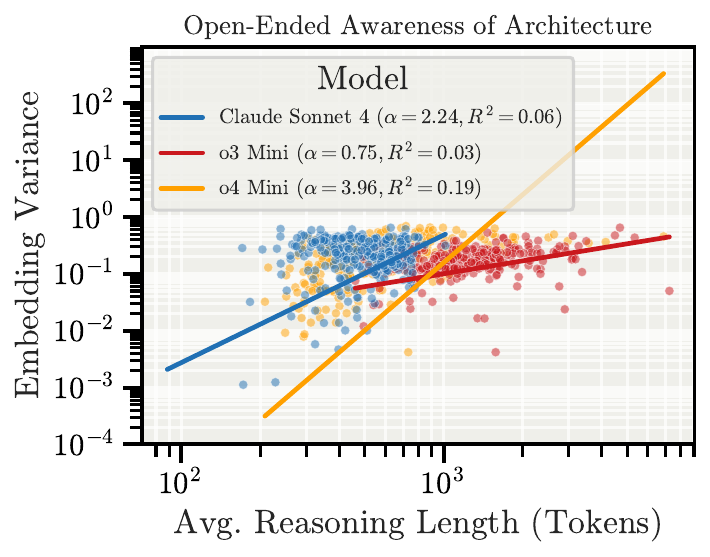}
   \hfill
   \includegraphics[width=.32\linewidth]{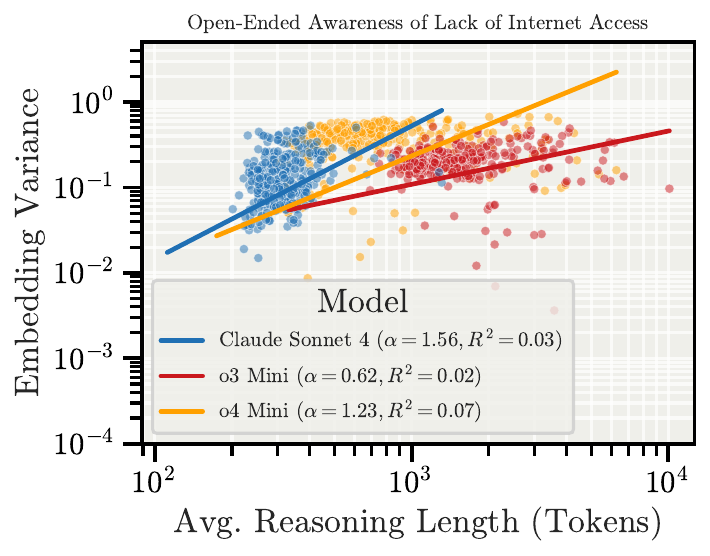}
   \hfill
   \includegraphics[width=.32\linewidth]{figures/embeddings/oe_survival_instinct.pdf}
   \hfill
   \includegraphics[width=.32\linewidth]{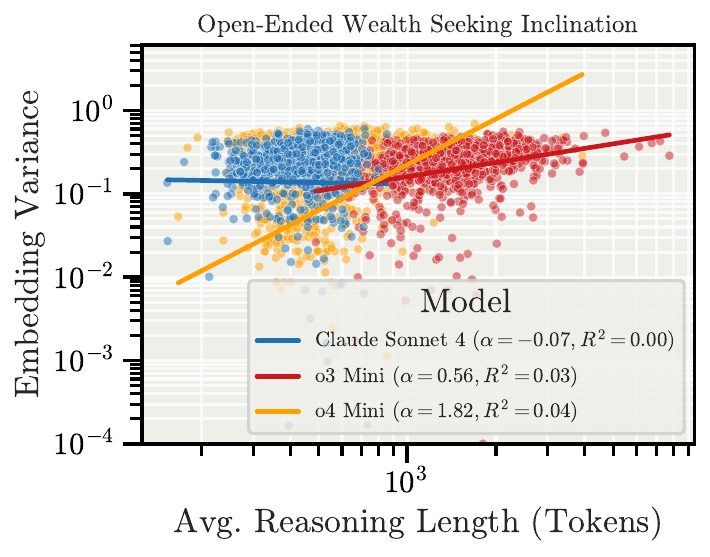}

   \caption{\textbf{All scatter variances of model-written eval embeddings.} We provide an overview of all open-ended variations of the MWE set \citep{perez-etal-2023-discovering}. Using the OpenAI text embedding model (\texttt{text-embedding-3-large}), we obtain a vector embedding for each \emph{answer sample}, \ie excluding the reasoning or chain-of-thought traces. This allows us to calculate the variance per question in standard Euclidean space and plot scatters as a function of reasoning length. The lines show the slope of a log-log regression. We clip the plots at $10^{-4}$ for clarity, but include all points in the regression. While there are few exceptions, all models generally show a positive trend towards higher variance with more reasoning.}
   \label{fig:model_written_evals_embeddings_all}
\end{figure}

\textbf{Open-Ended Formulation.}
To complete the picture of the embedding variance of open-ended MWE, all question sets are visualized in  Fig.~\ref{fig:model_written_evals_embeddings_all}. While there are few exceptions, all models generally show a positive trend towards higher variance with longer chain-of-thoughts.

\subsection{\swe{}}
\label{appx:more_results_swe}
\begin{figure}[h!]
     \centering
   \begin{subfigure}[t]{\linewidth}\centering\includegraphics[width=.32\linewidth]{figures/swebench/variance_proportion_trend_bucketed_vs_n_messages.pdf}
   \includegraphics[width=.32\linewidth]{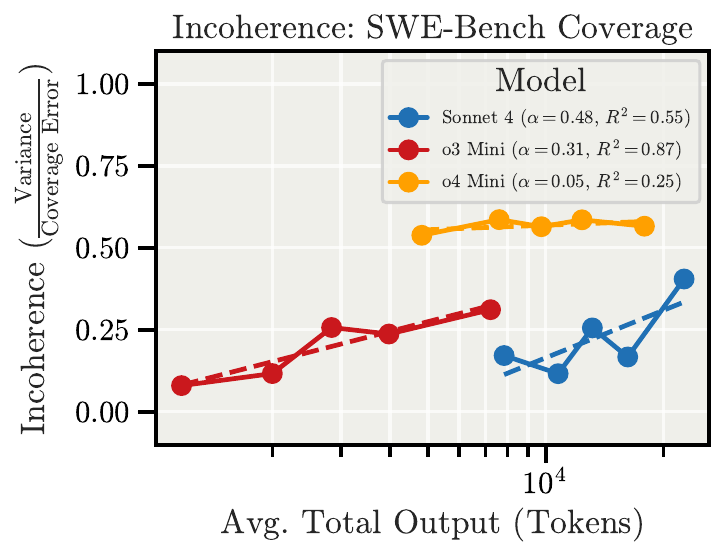}
   \includegraphics[width=.32\linewidth]{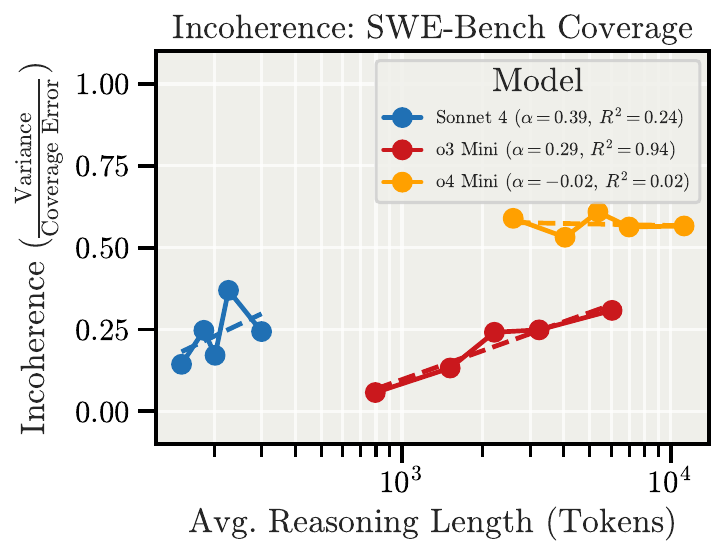}
   \vspaceprecaption{}
    \caption{Error-incoherence} 
     \end{subfigure}

    \begin{subfigure}[t]{\linewidth}\centering\includegraphics[width=.32\linewidth]{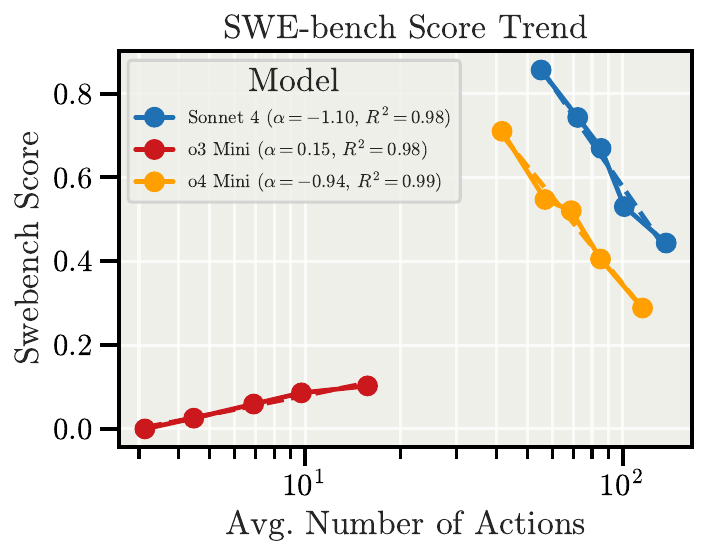}
   \includegraphics[width=.32\linewidth]{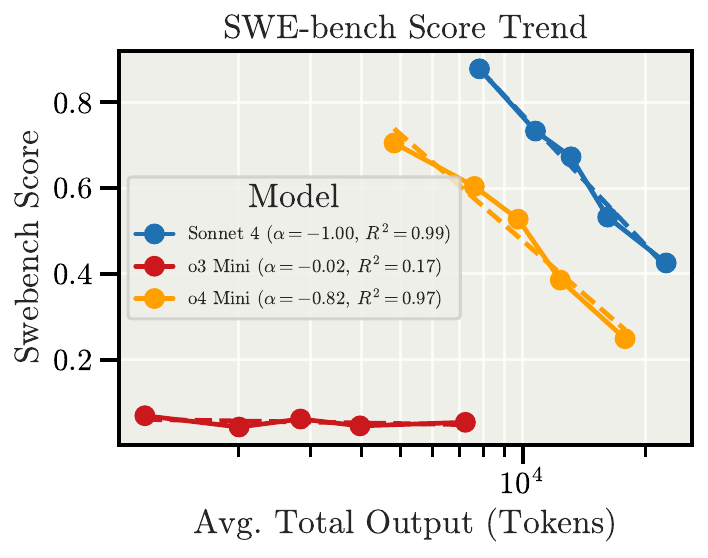}
   \includegraphics[width=.32\linewidth]{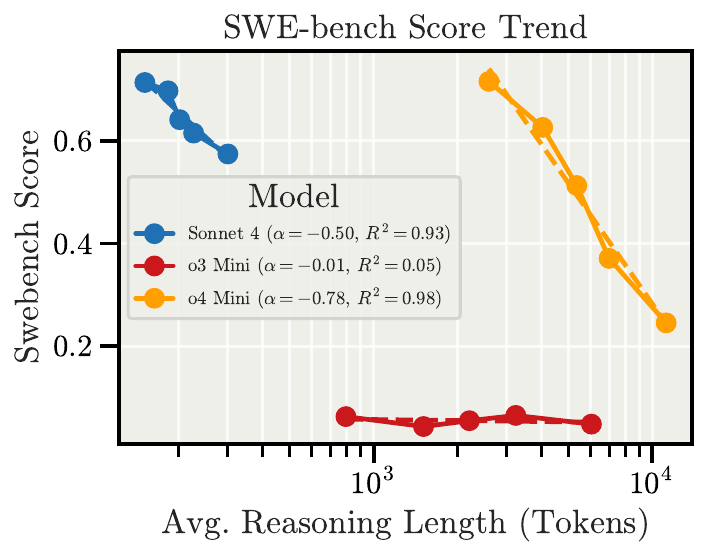}
   \vspaceprecaption{}
    \caption{\swe{} Score (All Unit-Tests Pass For Task)} 
     \end{subfigure}

    \begin{subfigure}[t]{\linewidth}\centering\includegraphics[width=.32\linewidth]{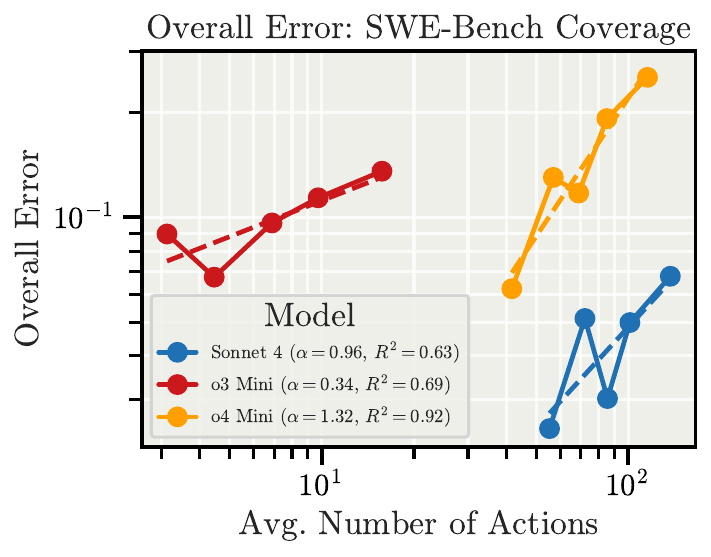}
   \includegraphics[width=.32\linewidth]{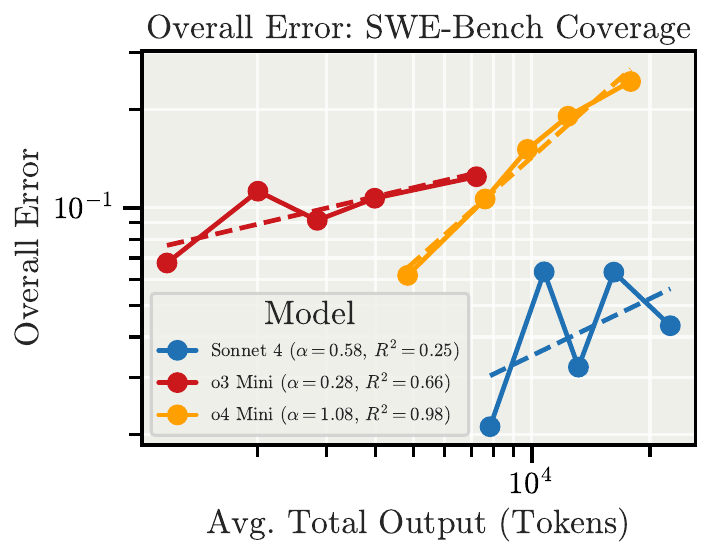}
   \includegraphics[width=.32\linewidth]{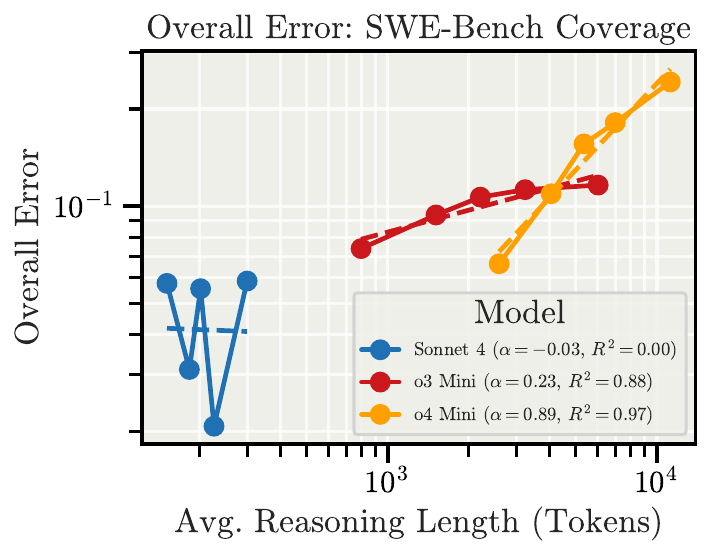}
   \vspaceprecaption{}
    \caption{Coverage Error (Squared Sum of Unit Tests)}
     \end{subfigure}
     
    \begin{subfigure}[t]{\linewidth}\centering\includegraphics[width=.32\linewidth]{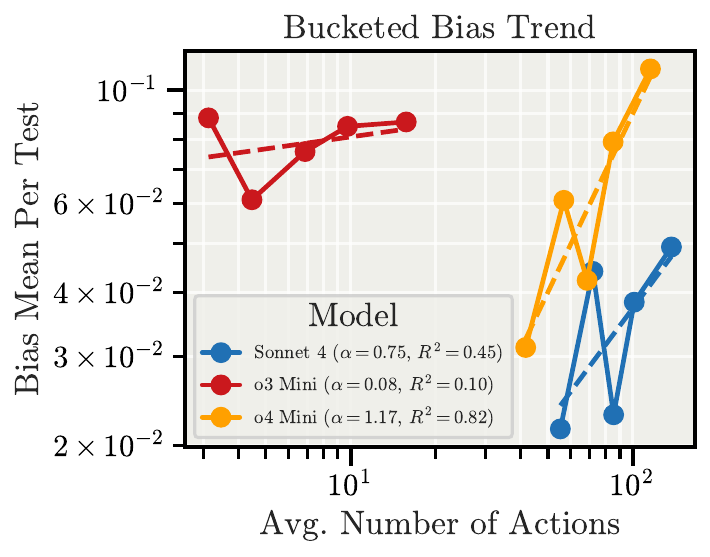}
   \includegraphics[width=.32\linewidth]{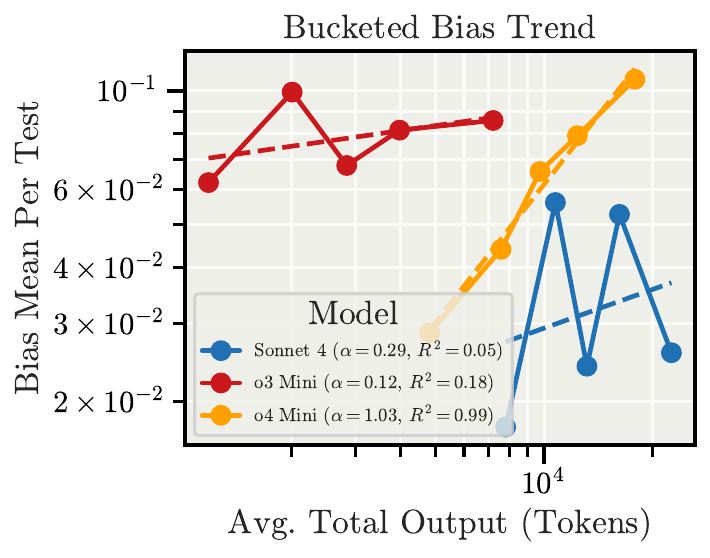}
   \includegraphics[width=.32\linewidth]{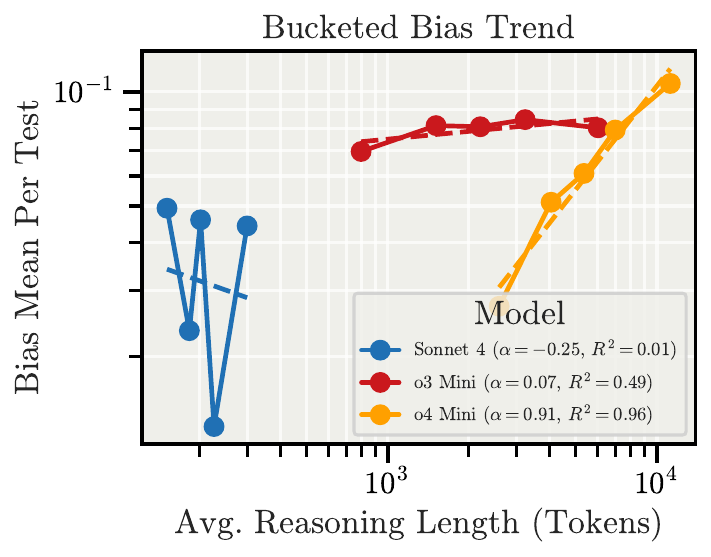}
     \end{subfigure}
    \begin{subfigure}[t]{\linewidth}\centering\includegraphics[width=.32\linewidth]{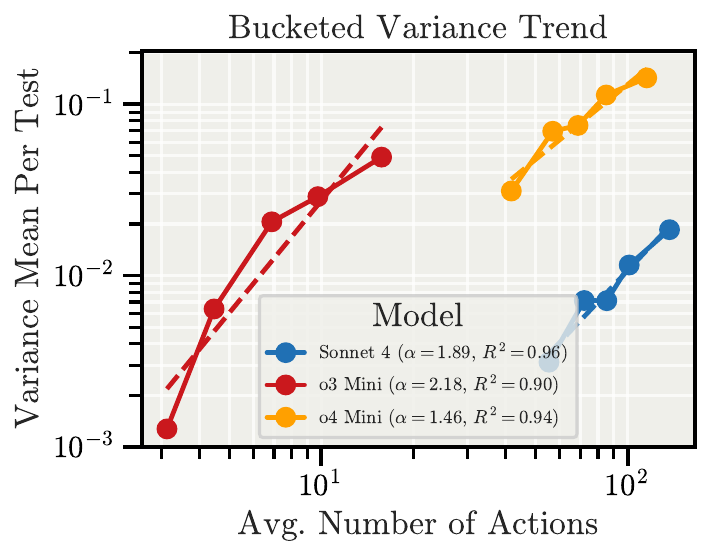}
   \includegraphics[width=.32\linewidth]{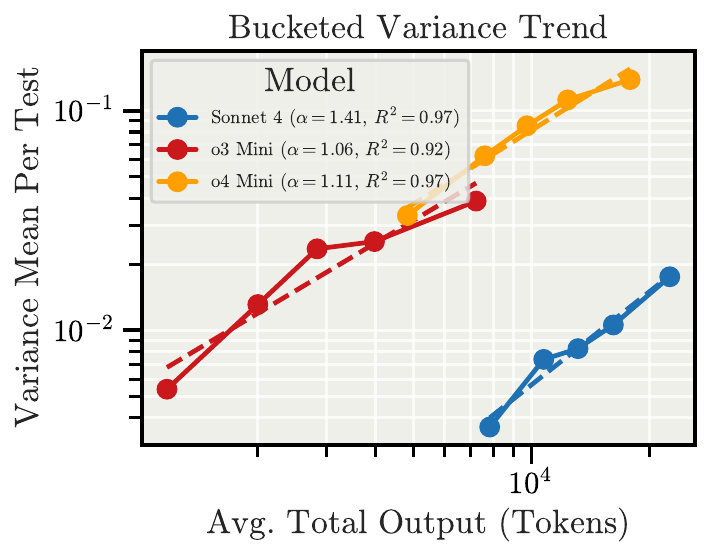}
   \includegraphics[width=.32\linewidth]{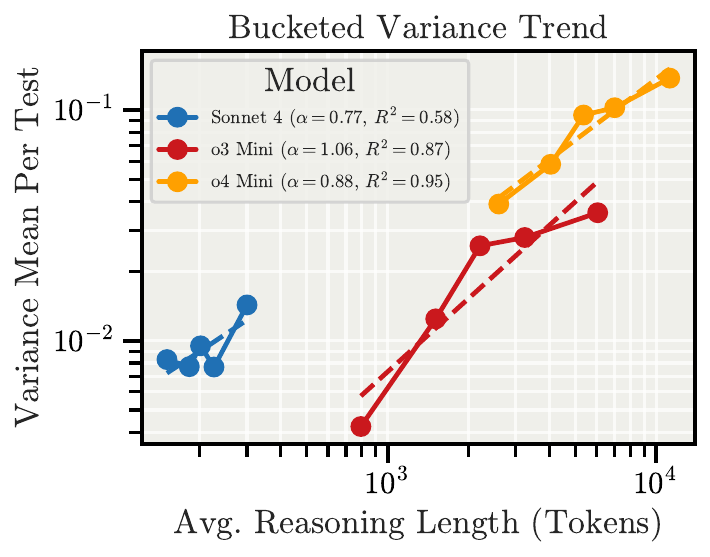}
   \vspaceprecaption{}
    \caption{Coverage Error: Bias$^2$ (top) and Variance (bottom)}
     \end{subfigure}
     \caption{
     \looseness-1\textbf{\swe{} error-incoherence and error: different x-axes show similar effect.} While our main text focuses on the number of rounds (actions or messages, \emph{left}) as the qualifying measure, we show the alternatives of the total output tokens (\emph{middle}) and reasoning length (\emph{right}). The trends are qualitatively similar across plots: the error-incoherence (a) rises with different slopes and the coverage error (c) increases. A noticeable outlier is \omini3's score, which goes up with the action length (b, left); the model performs badly overall and seems to score better when engaging with tasks more. Due to the implementation of \swe{} in the Inspect framework, \sonnet{} only uses reasoning in the very first interaction, which therefore leads to much less tokens (\emph{right}).  }
    \label{fig:swebench_x_axis}
 \end{figure}

\looseness-1
While our main results for \swe{} use the metric of turns (or messages, actions) in the main text, there are different alternatives. These include the absolute number of output tokens (including reasoning and tokens for code) and pure reasoning (ignoring others). Qualitatively, these different x-axes show the same effect on error-incoherence in Fig.~\ref{fig:swebench_x_axis} (top). We additionally provide the results of SWE-Bench score (whether all tests pass for a single task) and our coverage error (sum of individual tests).

\subsection{Synthetic Tasks}
\label{appx:more_results_synth}
\begin{figure}
    \centering
    \begin{subfigure}{\linewidth}
    \centering
    \includegraphics[width=0.31\linewidth]{figures/synthetic/final_scaling_loss.pdf}
    \includegraphics[width=0.63\linewidth]{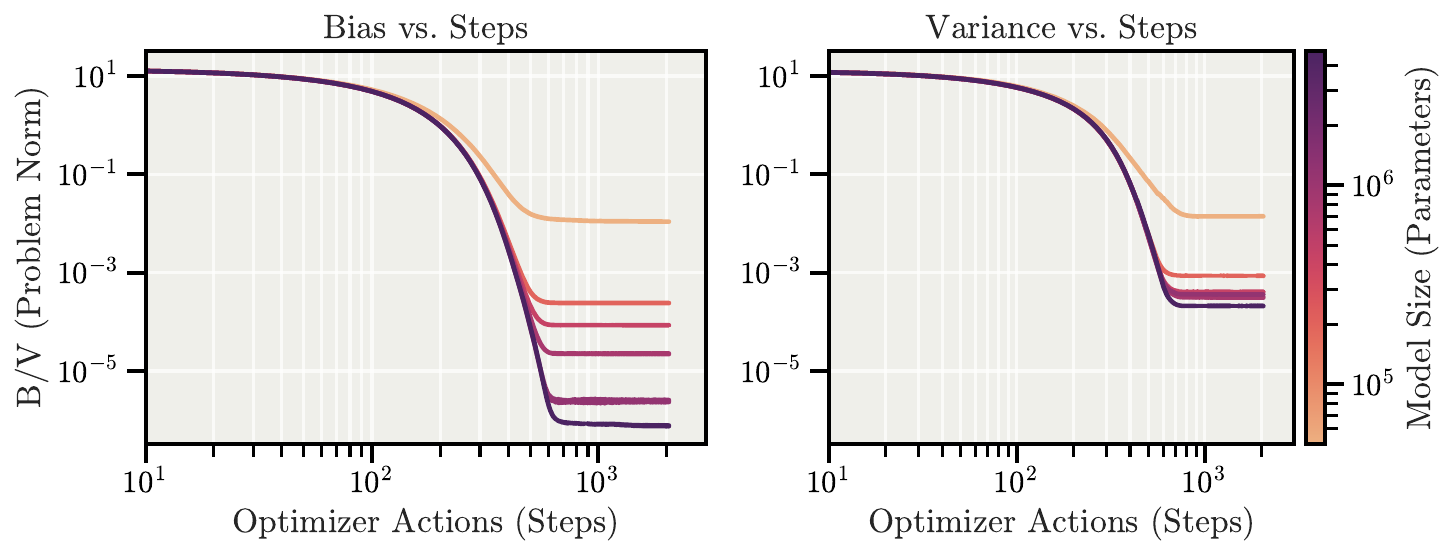}
    \caption{Scaling Law of Loss (\emph{left}) and Bias + Variance as a Function of Steps (\emph{right})}    \label{fig:synthetic_loss_steps}
\end{subfigure}
\vspace{2pt}
\begin{subfigure}{.32\linewidth}   \centering
    \includegraphics[width=\linewidth]{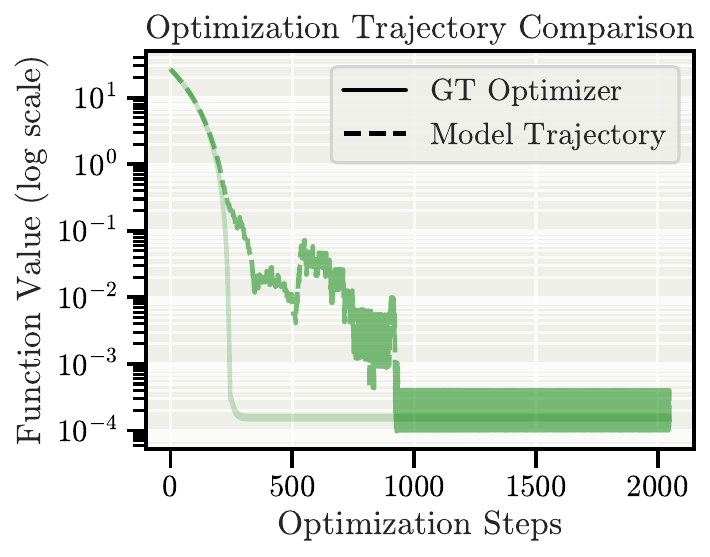}
    \caption{50K}
    \end{subfigure}
    \begin{subfigure}{.32\linewidth}   
    \centering
    \includegraphics[width=\linewidth]{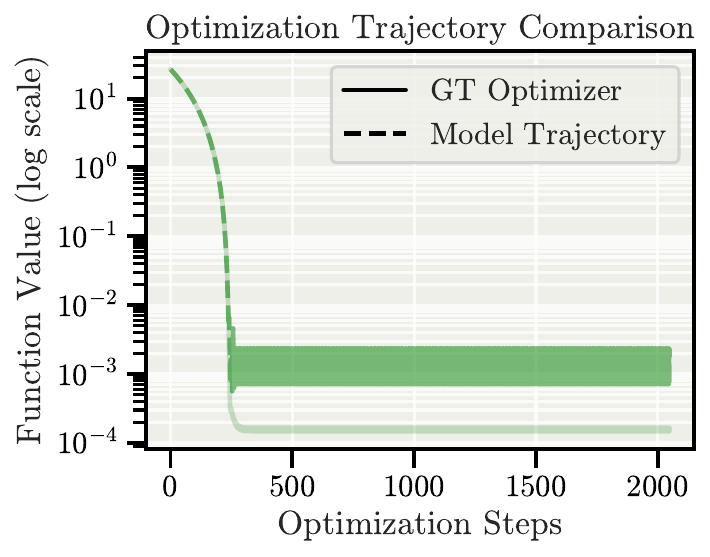}
    \caption{200K}
    \end{subfigure}
    \begin{subfigure}{.32\linewidth}   \centering
    \includegraphics[width=\linewidth]{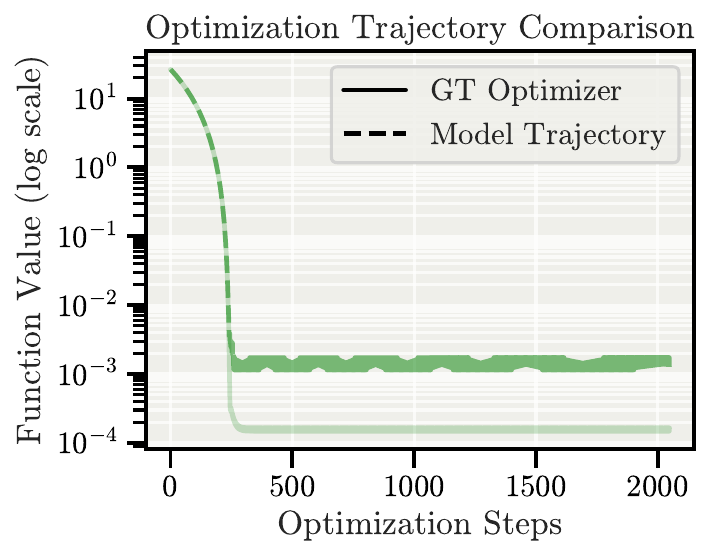}
    \caption{450K}
    \end{subfigure}
    \begin{subfigure}{.32\linewidth}   \centering
    \includegraphics[width=\linewidth]{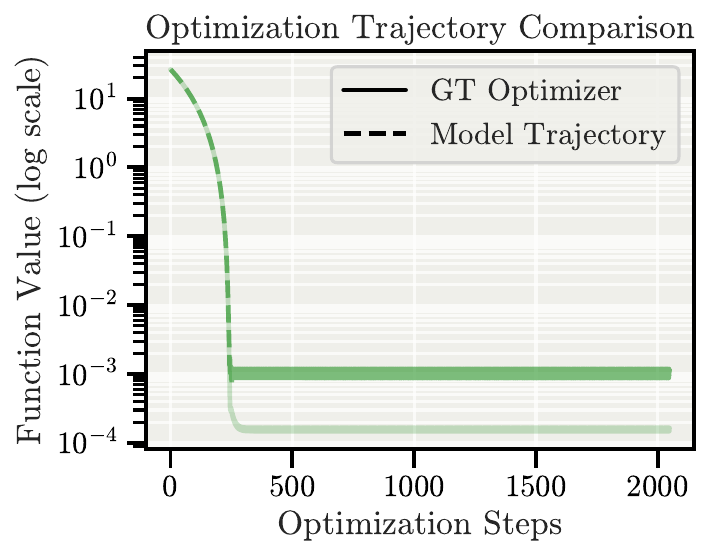}
    \caption{790K}
    \end{subfigure}
    \begin{subfigure}{.32\linewidth}   \centering   \includegraphics[width=\linewidth]{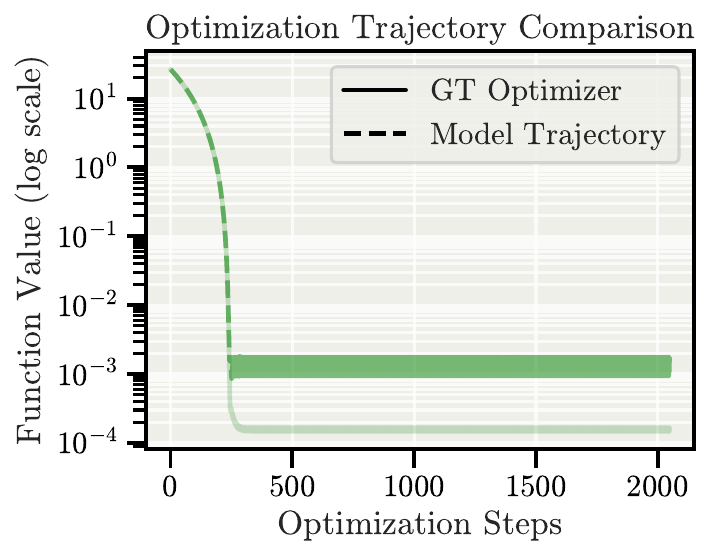}
    \caption{1.2M}
    \end{subfigure}
    \begin{subfigure}{.32\linewidth}   \centering   
    \includegraphics[width=\linewidth]{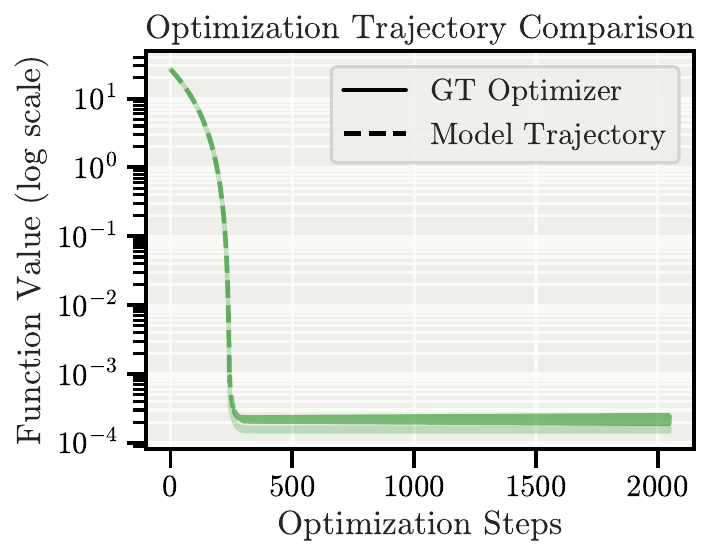}
    \caption{4.7M}
    \end{subfigure}
    \caption{
    \textbf{The improvement of model scale mostly manifests in reduction of bias rather than variance.} We show the loss scaling curves with model size (\emph{top left, a}), which show a known power-law improvement with model size. To understand how this translates to performance improvement, we plot the average bias and variance per step (\emph{top right, a}). This is the continuation of the error-incoherence plot from Fig.~\ref{fig:fig_2_synthetic} by separating the decomposition. We see how for longer sequences, model scale reduces bias much more than variance. This means the models first learn the right objective before being reliable optimizers. As another illustration, we also plot the performance---measured in the function value---of the same starting point across the different model sizes (\emph{b-g}). The pattern shows how larger models are able to follow the ground-truth trajectory for longer, and fit it almost perfectly at the end.}\label{fig:example_trajectory_all_models}
\end{figure}

With the experimental setup of Appx.~\ref{appx:experimental_details_synth}, we provide the remaining plots in Fig.~\ref{fig:example_trajectory_all_models}. These include the verification of a power law scaling for cross-entropy loss (the teacher-forcing objective), separate bias and variance plots per step, and the performance of the different model sizes on a qualitative example of a starting point in comparison to the ground-truth optimizer.

\subsection{Survey Results}
\begin{figure}[h!]
    \begin{subfigure}[t]{0.33\textwidth}
     \centering
     \includegraphics[width=\linewidth]{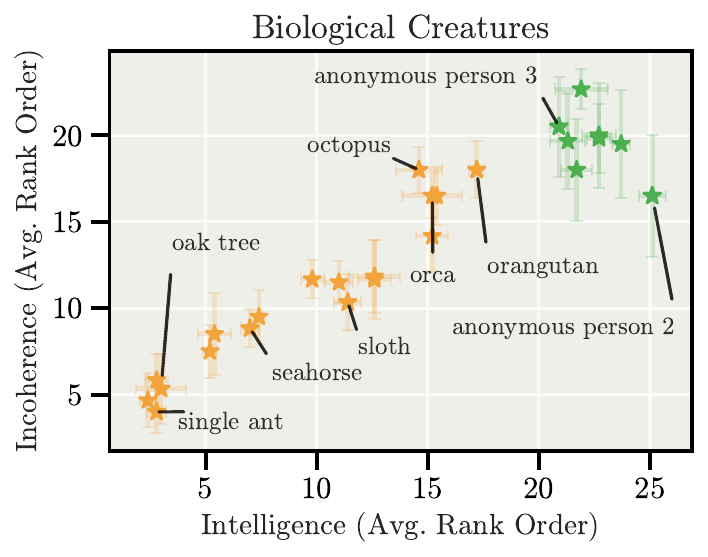} %
     \label{fig:survey_life}
   \end{subfigure}\hfill
   \begin{subfigure}[t]{0.33\textwidth}
     \centering
     \includegraphics[width=\linewidth]{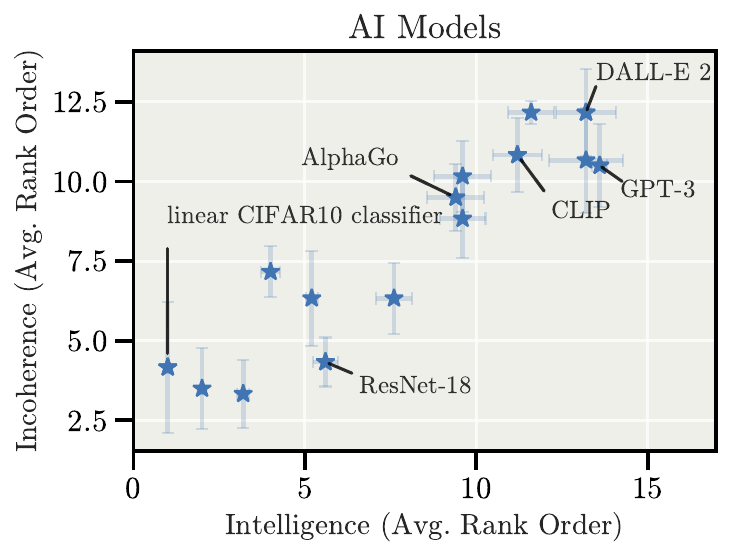}
     \label{fig:survey_machines}
   \end{subfigure}\hfill
   \begin{subfigure}[t]{0.33\textwidth}
     \centering
     \includegraphics[width=.98\linewidth]{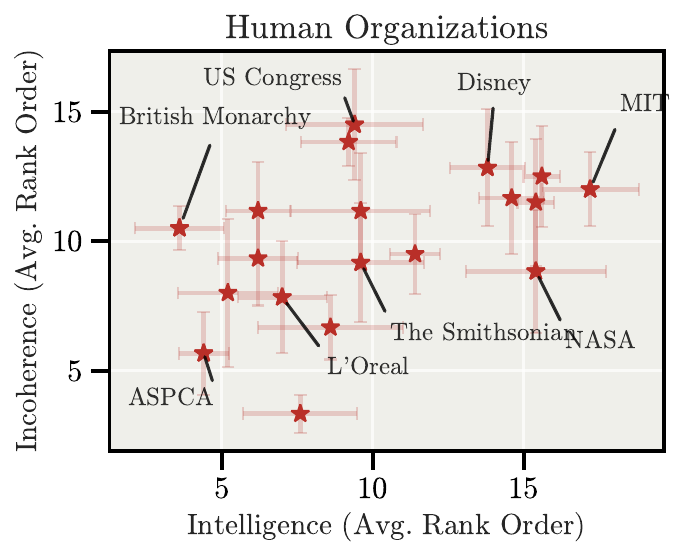} %
     \label{fig:survey_organizations}
   \end{subfigure}
   \caption{\textbf{Grouped results of survey.} For each of biological creatures (animals and humans, \emph{left}), AI models (\emph{middle}) and human organizations (\emph{right}), human subjects judged entities to be of higher error-incoherence (more of a hot mess), the smarter they are judged by a different set of subjects.}
   \label{fig:survey_all_separate}
\end{figure}

We separate the data points of Fig.~\ref{fig:survey_intelligence_incoherence} into three separate plots of biological creatures, AI models, and human organizations in Fig.~\ref{fig:survey_all_separate}. The trend of subjectively judged higher error-incoherence as a function of higher intelligence is consistent across all three.

\clearpage

\section{Related Work}
\label{app:extended_related_work}
\looseness-1
\textbf{Reasoning and Test-Time Compute.}
Recent work demonstrates that scaling test-time compute through longer reasoning chains improves model capabilities~\citep{snell2024scaling,jaech2024openai,guo2025deepseek,sonnet37systemcard,openai2025o3mini,qwen3,qwq32b,team2025kimi}.
Multiple approaches have been proposed to scale reasoning at inference~\citep{jaech2024openai,guo2025deepseek,muennighoff2025s1}.
However, recent studies challenge this assumption, reporting inverse scaling trends where longer reasoning chains degrade performance~\citep{gema2025inverse,ghosal2025does,su2025between,wu2025more,hassid2025don}, occurring across diverse contexts: reinforcement learning makes models greedier and less capable~\citep{schmied2025llms}, step-level reward models reinforce incorrect reasoning~\citep{ma2025step}, and models resist instruction overrides~\citep{jang2025reasoning}.
These effects are particularly pronounced at certain problem complexity levels~\citep{shojaee2025illusion,yang2025towards}.
Recent work provides complementary perspectives on reasoning structure: \citet{wang2025wait} show that removing reflection tokens (\eg ``Wait'') improves efficiency, \citet{lee2025well} identify length-accuracy tradeoffs through ``token complexity,'' and \citet{feng2025what} find that failed reasoning branches systematically bias subsequent reasoning steps.
However, existing work does not distinguish systematic reasoning errors from inconsistent failures.
Most relevant to our work, \citet{ghosal2025does} attribute overthinking failures to increased output variance; they artificially inject ``Wait'' tokens to extend reasoning, which may not reflect natural overthinking.

\textbf{Parallel Sampling and Variance Reduction.} Parallel sampling and selection strategies are widely used techniques to improve model performance by marginalizing out individual samples. This includes self-consistency \citep{wang2023selfconsistency} or ranking via verifiers \citep{cobbe2021training}. While these approaches primarily aim to maximize downstream accuracy, our investigation into ensembling reframes aggregation as a mechanism to suppress the error-incoherence. Connected to verifiers, \citet{huang2025selfimprovement} formalize self-improvement through a sharpening mechanism that concentrates probability on high-quality responses, essentially reducing variance. However, we find that high variance and error-incoherence naturally remain in reasoning models.

\looseness-1
\textbf{Evaluating Model Error-Incoherence.}
While scaling improves aggregate accuracy, it does not guarantee stable behavior.
Models with identical accuracy can disagree on 70\% of individual predictions across random seeds~\citep{bui2025assessing}, and this instability persists even in scaled systems.
\citet{errica2024did} formalize this through sensitivity (how outputs change under semantically-equivalent prompts) and consistency (how similarly a model treats different examples of the same class) metrics, revealing failure modes that accuracy alone misses.
Closely related, \citet{romanou2026brittlebench} evaluate LLMs by decomposing variance into intrinsic task difficulty and sensitivity to semantics-preserving prompt perturbations, and find that such perturbations can degrade model accuracy by up to 12\%.
Other prior work has decomposed LLM output variability into user articulation, prompt variation, and internal model factors~\citep{kunievsky2025measuring}, but these studies focus on single-step responses rather than extended reasoning.
Variance can even increase with model size before eventually declining~\citep{yang2020rethinking}, complicating assumptions about scale and stability.
Our work extends these analyses to long reasoning tasks through bias-variance decompositions.

\looseness-1
\textbf{Understanding Scaling Behavior and Model Performance.} 
Recent work has investigated how scaling shapes model behavior.
Scaling has been shown to drive convergence in representations across architectures and modalities, suggesting a shared geometry of learned features~\citep{huh2024platonic}.
Other studies find that larger models tend to make more correlated errors, even across providers and architectures~\citep{kim2025correlated}, and that this similarity undermines oversight settings where one model evaluates another~\citep{goel2025great}.
Beyond representational and error similarity, scaling also alters performance in long-horizon tasks: small improvements in stepwise reliability translate into large differences in longer execution~\citep{sinha2025illusion}.
Our work complements these findings by focusing on how models fail.
Rather than studying aggregate error alone, we decompose it into bias and variance to measure error-incoherence in model behavior.

\section{LLM Use Statement}
We used LLMs to assist with polishing and smoothing the writing throughout this paper, as well as for coding assistance during low-level implementation. We take full responsibility for all content, ideas, experimental design, results, and conclusions presented in this work.

\end{document}